\documentclass[lettersize,journal]{IEEEtran}
\usepackage{amsmath,amsfonts}
\usepackage{algorithmic}
\usepackage{algorithm}
\usepackage{array}
\usepackage{textcomp}
\usepackage{stfloats}
\usepackage{url}
\usepackage{verbatim}
\usepackage{graphicx}
\usepackage{cite}
\usepackage{multicol}
\usepackage{multirow}
\usepackage{tabularx}
\usepackage{booktabs}
\usepackage{bbding}
\usepackage{subfigure}
\usepackage{colortbl}
\usepackage{graphicx}
\usepackage{svg}
\usepackage{hyperref}
\usepackage{footnote}
\usepackage{orcidlink}
\hypersetup{
    colorlinks=true,
    linkcolor=red,
    urlcolor=red
}

\begin{document}

\title{A Survey on Autonomous Driving Datasets: Statistics, Annotation Quality, and a Future Outlook}

\author{Mingyu Liu*\orcidlink{0000-0002-8752-7950}, \and
Ekim Yurtsever\orcidlink{0000-0002-3103-6052}, \IEEEmembership{Member, IEEE}, \and
Jonathan Fossaert\orcidlink{0009-0005-1041-6338}, \and 
Xingcheng Zhou\orcidlink{0000-0003-1178-5221}, \and
\\
Walter Zimmer\orcidlink{0000-0003-4565-1272}, \and 
Yuning Cui\orcidlink{0000-0002-1279-5539}, \IEEEmembership{Student Member, IEEE}, \and
Bare Luka Zagar\orcidlink{0000-0001-5026-3368}, \and
Alois C. Knoll\orcidlink{0000-0003-4840-076X}, \IEEEmembership{Fellow, IEEE} 
        
\thanks{M. Liu, X. Zhou, W. Zimmer, Y. Cui, BL. Zagar, and AC. Knoll are with the Chair of Robotics, Artiﬁcial Intelligence and Real-Time Systems, Technical University of Munich, 85748 Garching bei München, Germany E-mail: \{mingyu.liu, xingcheng.zhou, walter.zimmer, bare.luka.zagar\}@tum.de, \{yuning.cui, knoll\}@in.tum.de}
\thanks{J. Fossaert is with the School of Engineering and Design, Technical University of Munich, 85748 Garching bei München, Germany (E-mail: jonathan.fossaert@tum.de)}
\thanks{E. Yurtsever is with the College of Engineering, Center for Automotive Research, The Ohio State University, Columbus, OH 43212, USA (E-mail: yurtsever.2@osu.edu)}
\thanks{* Corresponding author}}



\maketitle

\begin{abstract}
Autonomous driving has rapidly developed and shown promising performance due to recent advances in hardware and deep learning techniques. High-quality datasets are fundamental for developing reliable autonomous driving algorithms. Previous dataset surveys either focused on a limited number or lacked detailed investigation of dataset characteristics. To this end, we present an exhaustive study of 265 autonomous driving datasets from multiple perspectives, including sensor modalities, data size, tasks, and contextual conditions. We introduce a novel metric to evaluate the impact of datasets, which can also be a guide for creating new datasets. Besides, we analyze the annotation processes, existing labeling tools, and the annotation quality of datasets, showing the importance of establishing a standard annotation pipeline.
On the other hand, we thoroughly analyze the impact of geographical and adversarial environmental conditions on the performance of autonomous driving systems. Moreover, we exhibit the data distribution of several vital datasets and discuss their pros and cons accordingly. Finally, we discuss the current challenges and the development trend of the future autonomous driving datasets.
\href{https://github.com/MingyuLiu1/autonomous_driving_datasets.git}{https://github.com/MingyuLiu1/autonomous\_driving\_datasets.git}
\end{abstract}


\section{Introduction}
\IEEEPARstart{A}{utonomous} driving (AD) aims to revolutionize the transportation system by creating vehicles that can accurately perceive their environment, make intelligent decisions, and drive safely without human intervention. Due to thrilling technical development, various autonomous driving products have been implemented in several fields, such as robotaxis~\cite{mao20233d}. These rapid advancements in autonomous driving rely heavily on extensive datasets, which help autonomous driving systems be robust and reliable in complex driving environments.

\begin{figure}[t!]
    \centering
    \includegraphics[width=0.48\textwidth]{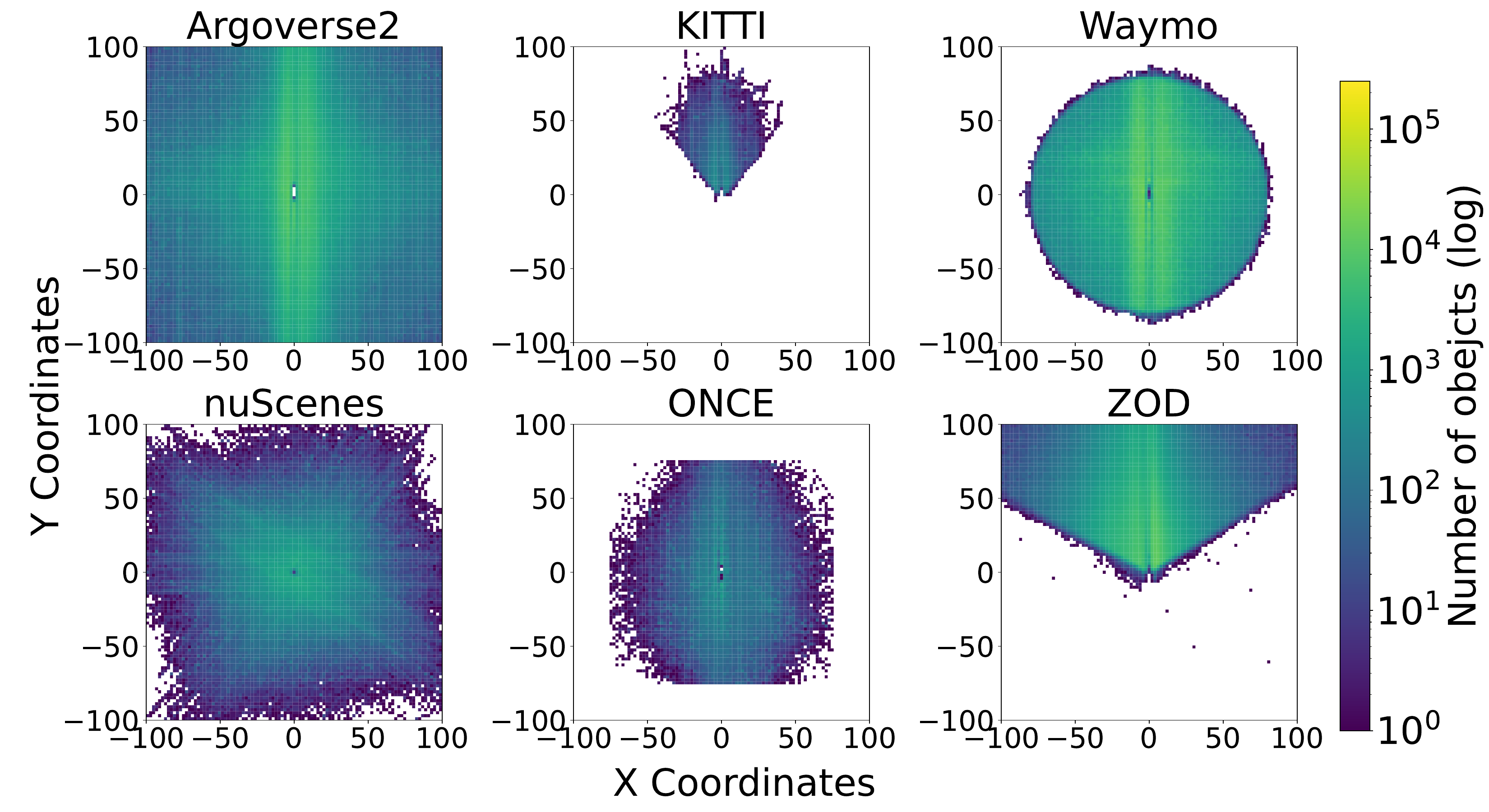}
    \caption{Bird's-Eye View object distribution of datasets. Each heatmap represents a dataset and is plotted using X and Y coordinates. Y is the driving direction of the ego-vehicle. The unique annotation characters of each dataset are reflected in the distribution range, density, and number of bounding boxes.}
    \label{fig:figure1}
\end{figure}

In recent years, there has been a significant increase in the quality and variety of autonomous driving datasets. The first apparent phenomenon in the development of datasets is the various data collection strategies, including synthetic datasets~\cite{ros2016synthia, gaidon2016virtual, richter2016playing, richter2017playing, behley2019semantickitti, hendrycks2019scaling, hurl2019precise, cabon2020virtual, sun2022shift, xu2022v2x, wang2023deepaccident} generated by simulators and recorded from the real world~\cite{geiger2012we, cordts2016cityscapes, zhu2016traffic, rasouli2017they, huang2018apolloscape, krajewski2018highd, chang2019argoverse, zhan2019interaction, rasouli2019pie, caesar2020nuscenes, sun2020scalability, yu2020bdd100k, geyer2020a2d2, sheeny2021radiate, yu2022dair, wilson2023argoverse, li2022automine}, to name just a few. Secondly, the datasets vary in composition, including but not limited to multiple sensor modalities like camera images and LiDAR point clouds, different annotation types for various tasks, and data distribution. Fig.~\ref{fig:figure1} depicts the distribution of 3D object bounding box in six well-known real-world datasets (\textit{Argoverse~2}\cite{wilson2023argoverse}, \textit{KITTI}~\cite{geiger2012we}, \textit{nuScenes}~\cite{caesar2020nuscenes}, \textit{ONCE}~\cite{mao2021one}, \textit{Waymo}~\cite{sun2020scalability}, and \textit{ZOD}~\cite{alibeigi2023zenseact}) under a Bird's-Eye View  (BEV), highlighting each dataset's unique annotation characteristics. The sensor mounting positions also reflect datasets' various sensing domains, including onboard, Vehicle-to-Everything (V2X), or drone domain. The datasets' geometric diversity and varying weather conditions also enhance the generalizability of autonomous driving datasets.


\subsection{Research Gap \& Motivation}
\begin{figure}
    \centering
    \includegraphics[width=0.48\textwidth, trim={0, 5, 0, 0}]{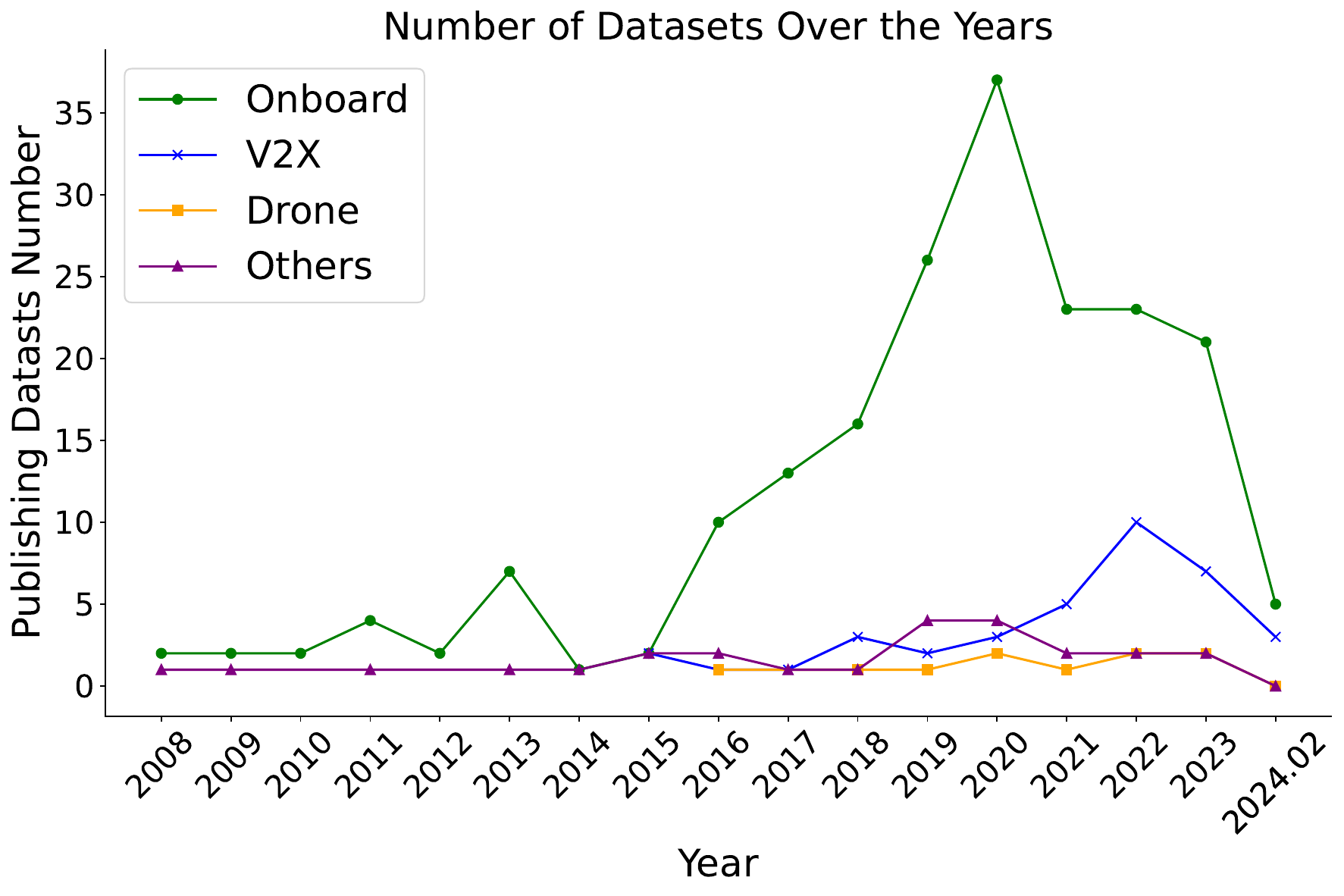}
    \caption{Overview of dataset publication trends from 2008 to 2024. The diagram demonstrates a significant increase in the publication of onboard datasets between 2015 and 2020, followed by a gradual decline thereafter. In contrast, there has been a rising trend in the publication of V2X datasets, indicating growing research interest in cooperative perception systems.}
    \label{fig:pub_trend}
\end{figure}

We demonstrate the yearly published number of perception datasets in Fig.~\ref{fig:pub_trend} to illustrate the growth trends in autonomous driving datasets. Given the vast and growing number of publicly available datasets, a comprehensive survey of these resources is valuable for advancing both academic and industrial research in autonomous driving. In prior work, Yin et al.~\cite{yin2017use} summarized 27 publicly available datasets containing data collected on public roads. As a sequential work of ~\cite{yin2017use}, \cite{kang2019test} extended the number of datasets. Guo et al.~\cite{guo2019safe} and Janai et al.~\cite{janai2020computer} proposed systematical introductions to the existing datasets from an application perspective. Beyond describing existing datasets, Liu et al.~\cite{liu2021survey} discussed the domain adaptation between synthetic and real data and automatic labeling methods. Li et al.~\cite{li2023open} summarized existing datasets and undertook an exhaustive analysis of the characters of the next-generation datasets. However, these surveys only summarized a small number of datasets, causing a non-wide scope. AD-Dataset~\cite{bogdoll2022ad} collected a large number of datasets while lacking detailed analysis for the attributes of these datasets. Compared to studies on versatile datasets, some researchers presented surveys on a particular type of autonomous driving dataset, such as anomaly detection~\cite{bogdoll2023perception}, synthetic datasets~\cite{song2023synthetic}, 3D semantic segmentation~\cite{gao2021we}, or decision-making~\cite{wang2023survey}. Additionally, some task-specific surveys~\cite{teng2023motion, chen2022milestones} also organized the related AD datasets. 

In this work, we present a comprehensive and systematic survey of a large number of datasets in autonomous driving. We compare our survey to others in Table~\ref{tab:survey_comparison}. Our survey covers all tasks from perception to control, considers real-world as well as synthetic data, and provides insights into the data modality and quality of several crucial datasets.

 \begin{table*}[htb]
  \centering
  \caption{We compare our survey paper with other AD dataset surveys in the following perspectives: collected dataset number (\#Dataset), relevant tasks, sensing domain (S. domain), sensor modality (S. moda.), geometric conditions (Geo.), environmental conditions (Env.), analyzing data distribution, introducing annotation quality and process. In the context of environmental conditions, we refer to the variability in weather conditions and illumination. Geometric conditions include scenario types and geographical scope. We describe the task types in a coarse granularity, including perception (Perc.), prediction (Pred.), planning (Pl.), control (C.), 0and end-to-end (E2E).}
  \resizebox{\textwidth}{!}{ 
    \begin{tabular}{l ccc cccccc cc}
    \toprule[1pt]
    \multirow{2}{*}{\textbf{Survey}} & \multicolumn{3}{c}{\textbf{General}} & \multicolumn{5}{c}{\textbf{Data}} & \multicolumn{2}{c}{\textbf{Annotation}} \\
    \cmidrule(lr){2-4} \cmidrule(lr){5-9} \cmidrule(lr){10-11}
    & \multicolumn{1}{c}{\textbf{Year}} & \textbf{\#Datasets} & \multicolumn{1}{c}{\textbf{Tasks}} & \textbf{S. domain} & \textbf{S. moda.} & \textbf{Geo.} & \textbf{Env.} & \multicolumn{1}{c}{\textbf{Data analysis}} & \textbf{Quality} & \textbf{Process}
    \\ \midrule
    When to use what dataset~\cite{yin2017use} & 2017 & 27 & Perc & ~ & $\checkmark$ & $\checkmark$ & $\checkmark$ & ~ & ~ & ~
    \\
    Self-driving Algorithm~\cite{kang2019test} & 2019 & 37 & Perc & ~ & $\checkmark$ & $\checkmark$ & $\checkmark$ & ~ & ~ & ~
    \\
    Is it safe to drive~\cite{guo2019safe} & 2019 & 54 & Perc, Pred, E2E & & $\checkmark$ & $\checkmark$ & $\checkmark$ & ~ & ~ & ~
    \\
    CV for AVs~\cite{janai2020computer} & 2020 & 33 & Perc & ~ & $\checkmark$ & $\checkmark$ & $\checkmark$ & ~ & ~ & $\checkmark$ 
    \\
    A Survey on AD Datasets~\cite{liu2021survey} & 2021 & 30 & Perc & ~ & $\checkmark$ & $\checkmark$ & $\checkmark$ & ~ & ~ & $\checkmark$
    \\
    3D Semantic Segmentation~\cite{gao2021we} & 2021 & 29 & Perc & ~ & $\checkmark$ & $\checkmark$ & $\checkmark$ & $\checkmark$ & $\checkmark$ & $\checkmark$  \\
    AD-Dataset~\cite{bogdoll2022ad} & 2022 & 204 & Perc, Pred, Pl, C & $\checkmark$ & $\checkmark$ & $\checkmark$ & ~ & ~ & ~ & ~  
    \\
    Anomaly Detection~\cite{bogdoll2023perception} & 2023 & 16 & Perc & ~ & $\checkmark$ & $\checkmark$ & $\checkmark$ & $\checkmark$ & ~ & ~
    \\
    Synthetic Datasets for AD~\cite{song2023synthetic} & 2023 & 17 & Perc, Pred & ~ & $\checkmark$ & $\checkmark$ & $\checkmark$ & ~ & ~ & ~
    \\
    Decision-making~\cite{wang2023survey} & 2023 & 25 & Pl, C & ~ & $\checkmark$ & $\checkmark$ & $\checkmark$ & ~ & ~ &  ~
    \\
    Open-sourced Data Ecosystem~\cite{li2023open} & 2023 & 70 & Perc, Pred, Pl & $\checkmark$ & $\checkmark$ & $\checkmark$ & ~ & ~ & ~ & $\checkmark$
    \\
    \rowcolor{gray!10}
    \textbf{Ours} & 2024 & 265 & Perc, Pred, Pl, C, E2E & $\checkmark$ & $\checkmark$ & $\checkmark$ & $\checkmark$ & $\checkmark$ & $\checkmark$ & $\checkmark$ 
    \\ 
    \bottomrule[1pt]
    \end{tabular}
}
  \label{tab:survey_comparison}
\end{table*}

\subsection{Main Contributions}
\textbf{The main contributions of this paper can be summarized as follows}:
\begin{itemize}
    \item We present an overview of the most exhaustive survey on autonomous driving datasets recorded to date. We show publicly available datasets as comprehensively as possible, recording their fundamental characteristics, such as published year, data size, sensor modalities, sensing domains, geometrical and environmental conditions, and support tasks. 
    \item We systematically illustrate the sensors and sensing domains for collecting AD data. Furthermore, we describe the main tasks in autonomous driving, including task goals, required data modalities, and evaluation metrics.
    \item We categorize datasets based on their sensing domains and relevant tasks, enabling researchers to efficiently identify and compile information for their target datasets. This approach facilitates more focused and productive research and development efforts.
    \item Additionally, we introduce an impact score metric to evaluate the influence of published perception datasets. This metric can also be a guide for developing future datasets. We deeply analyze datasets with high impact scores, highlighting their advantages and utility.
    \item We investigate the annotation quality of datasets and the existing labeling procedures for various autonomous driving tasks.
    \item Our detailed data statistics demonstrate the data distribution of various datasets from different perspectives, exhibiting their inherent limitations and suitable use cases.
    \item We analyze recent technology trends and the development direction of next-generation datasets, such as integrating language into AD data, generating AD data using Vision Language Models, standardizing data creation, and promoting an open data ecosystem.

\end{itemize}

\subsection{Scope \& Limitations}
We aim to conduct an exhaustive survey on the existing autonomous driving datasets to facilitate the development of future algorithms and datasets in this field. We collected datasets relevant to the five fundamental tasks of autonomous driving: perception, prediction, planning, control, and end-to-end (E2E) driving. To maintain clarity and prevent redundancy, we describe versatile datasets only within the main scope they support. Additionally, we collected a large number of datasets and exhibited them with their primary characters in tables. However, to ensure this survey aids researchers effectively, we focus our detailed discussions on the most impactful datasets rather than providing extensive descriptions of all datasets.

\subsection{Survey Structure}
\begin{figure}
    \centering
    \includegraphics[width=0.48\textwidth, trim={30 100 30 80}]{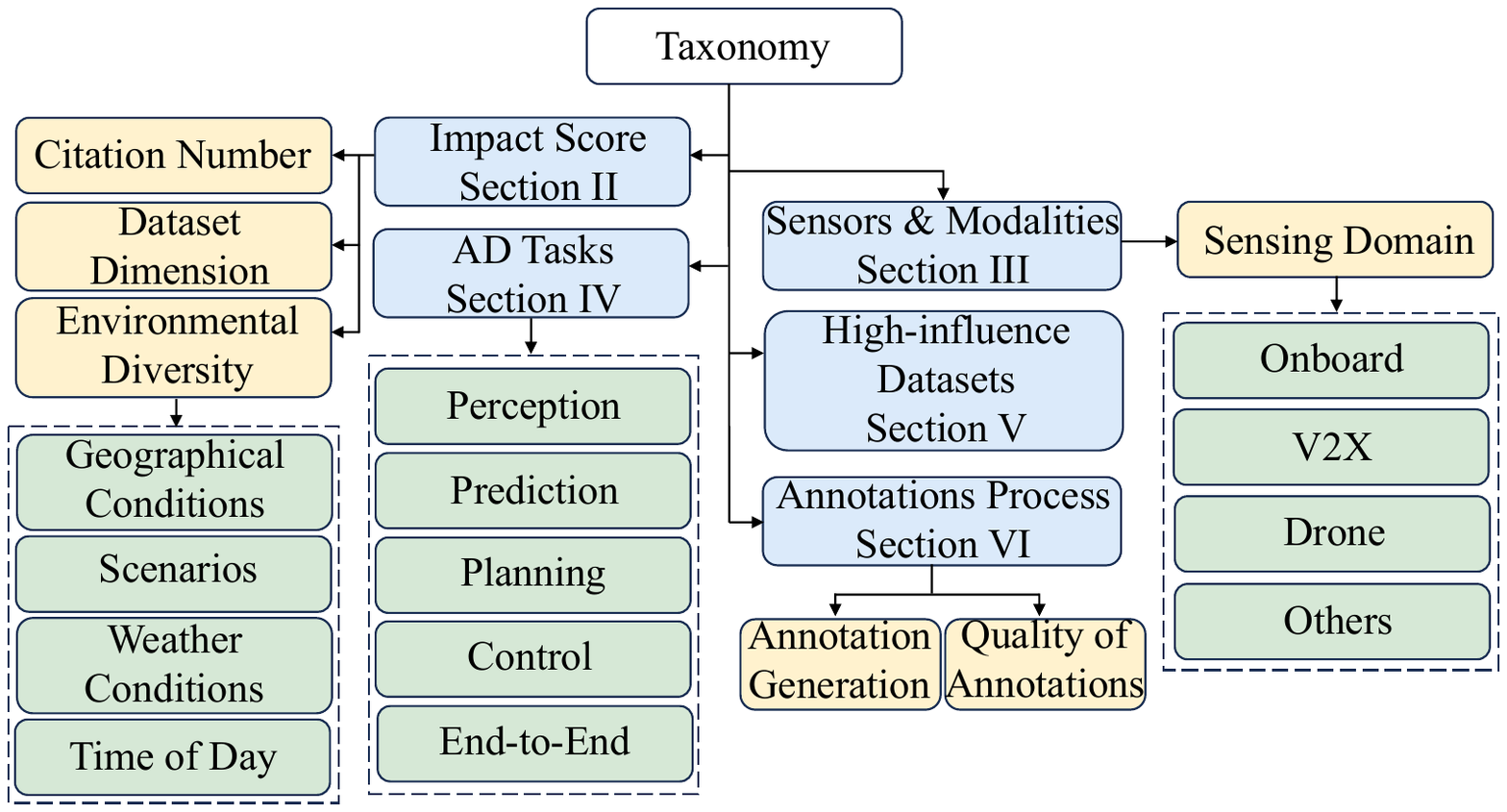}
    \caption{This survey's primary taxonomy includes impact score, sensors and modalities, autonomous driving tasks, high-influence datasets, and annotation process.}
    \label{fig:taxonomy}
\end{figure}

The rest of the survey is structured as follows: Section~\ref{methodology} introduces the approach leveraged to source public datasets and the evaluation metrics for them. Section~\ref{data_source_cp} demonstrates the primary sensors used in autonomous driving and their modalities. Section~\ref{tasks_ad} discusses autonomous driving tasks, related challenges, and required data. In-depth discussions of several important datasets are presented in Section~\ref{influence_impact}. The process of annotations and factors affecting annotation quality are addressed in Section~\ref{annotation_process}. Section~\ref{data_analysis} provides statistical analysis of data distribution across various datasets. Future trends and potential research directions in autonomous driving datasets are explored in Section~\ref{discussion_future_works}. The paper concludes with Section~\ref{conclusion}. The survey's taxonomy is shown in Fig.~\ref{fig:taxonomy}.

\section{Methodology}
\label{methodology}
This section comprises two parts: 1) the collection and filtering of datasets, as detailed in Section \ref{collec_dataset}, and 2) the evaluation metric of a dataset's impact on the autonomous driving domain, described in Section \ref{evaluation_metrics}.

\subsection{Datasets Collection}
\label{collec_dataset}
Following \cite{kitchenham2004procedures}, we conducted a systematic review to exhaustively collect published autonomous driving datasets.

To ensure source diversity, we utilized well-known search engines such as \textit{Google},
\textit{Google Scholar}
and \textit{Baidu}
to search datasets. To ensure a thorough dataset collection from various countries and regions, we conducted searches in English, Chinese, and German using keywords such as ``autonomous driving dataset/benchmark'', ``intelligent vehicle dataset/benchmark'' and terms related to object detection, classification, tracking, segmentation, prediction, planning, control, and end-to-end driving.

Furthermore, we explored \textit{IEEE Xplore}, \textit{Paperswithcode}, and pertinent conferences in autonomous driving and intelligent transportation systems to collate datasets from journals and conference proceedings. We verified datasets from these sources through keyword searches and manual title reviews.

Finally, to ensure the inclusion of specialized or lesser-known datasets, we searched through \textit{Github} repositories. Similar to databases, we performed both manual and keyword-based searches for datasets.


\subsection{Dataset Evaluation Metrics}
\label{evaluation_metrics}
We introduce a novel metric, impact score, to assess the significance of a published dataset, which can also be a guide to preparing a new dataset. In this section, we explain in detail the approach to calculate the impact score of autonomous driving datasets.

For a fair and compatible comparison, we only consider datasets related to the perception domain, which takes up a large portion of autonomous driving datasets. Additionally, to ensure the objectivity and comprehensibility of our scoring system, we take into account various factors, including citation score, data dimension, and environmental diversity. All the values are gathered from official papers or open-source dataset websites.

\noindent \textbf{Citation Score.} 
First, we calculate the citation scores from the total citation number and average annual citation. To gain fair citation counts, we choose the time of the earliest version of a dataset as its publication time. Moreover, all citation counts were collected as of March 05, 2024, to ensure the comparison is based on a consistent timeframe. The total citation number $c^{t}$ reflects the overall influence of a dataset. A higher count of this metric means that the dataset has been widely recognized and utilized by researchers. However, datasets published in earlier years can accumulate more citations. To address this unfairness, we leverage average annual citation $c^{a}$, which describes a dataset's yearly citation increase speed. The function is shown in Eq.~\ref{aa_citation}.
\begin{equation} \label{aa_citation}
    c^{a} = 
    \begin{cases} 
    c^{t} / \left(y_{curr} - y_{pub}\right) & \text{if } y_{curr} \neq y_{pub} \\
    c^{t} & \text{if } y_{curr} = y_{pub}
    \end{cases}
\end{equation}
where $y_{curr}$ and $y_{pub}$ represent the current year and the dataset published year, respectively. On the other hand, the citation number of distastes has a wide distribution range from single digits to tens of thousands. To alleviate the extreme imbalance and highlight the differences of each dataset, we apply a logarithmic transformation followed by Min-Max normalization to both $c^{t}$ and $c^{a}$, described in Eq.~\ref{citation_score}.
\begin{equation} \label{citation_score}
    c_{norm} = \min \text{-} \max \left(\log(c)\right)  
\end{equation}
Finally, the citation score $c_{score}$ is the summation of $c^{t}_{norm}$ and $c^{a}_{norm}$:
\begin{equation}
    c_{score} = 0.5 \* c^{t}_{norm} + 0.5 \* c^{a}_{norm}
\end{equation}
\noindent \textbf{Data Dimension Score.}
We measure the data dimension across four perspectives: dataset size, temporal information, task number, and labeled categories. Dataset size $f$ is represented by the frame number of a dataset, reflecting its volume and comprehensiveness. To get the dataset size score $f_{norm}$, we leverage the same method as the citation score to process the frame number to overcome the extreme imbalance between different datasets.

Temporal information is essential for autonomous driving as it enables the vehicle to understand how the surrounding environment changes over time. We use $t\in\left\{0,1\right\}$ to indicate whether a dataset includes temporal information. Regarding the task number, we only consider datasets related to the six fundamental tasks in the autonomous driving perception domain, such as 2D object detection, 3D object detection, 2D semantic segmentation, 3D semantic segmentation, tracking, and lane detection. Therefore, the task number score is recorded as ${t^{n}}\in\left\{1,2,3,4,5,6\right\}$. A large number of categories is critical for the robustness and versatility of a dataset. During the statistic, if a dataset supports multiple tasks and includes various types of annotation, we choose the largest number of categories. Afterward, the categories are divided into five levels, $cat=\left\{1,2,3,4,5\right\}$, based on quintiles. We normalize $t_{n}$ and $l$ before the following process to simplify the calculation.

In order to reflect the data dimension score $d_{score}$ as objectively as possible, we give different weights to the four components, as shown in Eq.~\ref{dimension_score}.
\begin{equation} \label{dimension_score}
    d_{score} = 0.5 \* f_{norm} + 0.1 \* t + 0.2 \* t^{n}_{norm} + 0.2 \* cat_{norm}
\end{equation}
\noindent \textbf{Environmental Diversity Score.}
We evaluate the environmental diversity of a dataset according to the following factors: 1) weather conditions (such as rain or snow), 2) time of day (e.g., morning or dusk), 3) types of driving scenarios (like urban or rural), and 4) geometric scope refers to the number of countries or cities where the data is recorded. It is worth noting that we treat the geographical scope for synthetic datasets as undefined. We follow the granularity with which a paper categorizes its data to quantify the diversity. Moreover, for the missing value, if a dataset announces that the data is recorded under diversity conditions, we assign the median value; otherwise, the missing value is set to one. We apply quintiles to quantify each factor into five distinct levels. After that, the environmental diversity score $e_{score}$ is the sum of these four factors.

In the end, we leverage Eq.~\ref{impact_score} to calculate the impact score $i_{score}$.
\begin{equation} \label{impact_score}
    i_{score} = 60 \* c_{score} + 20 \* d_{score} + 20 \* e_{score}
\end{equation}
The total impact score is 100, and 60 percent belongs to the citation score $c_{score}$. Data dimension score $d_{score}$ and environmental diversity score $e_{score}$ takes 40 percent.

\section{Sensors and Perception Technology in Autonomous Driving}
\label{data_source_cp}
In this section, we introduce the sensors and their modalities mainly used in autonomous driving (\ref{sensors_modality}). Subsequently, in \ref{sensing_domain_coperception_system}, we analyze the data acquisition methods and cooperative perception technologies.

\subsection{Sensor Modalities}
\label{sensors_modality}
\begin{figure}[htb]
    \centering
    \subfigure[Camera]{
        \begin{minipage}{0.135\textwidth}
            \includegraphics[width=\textwidth, trim={0, -10, 0, 0}]{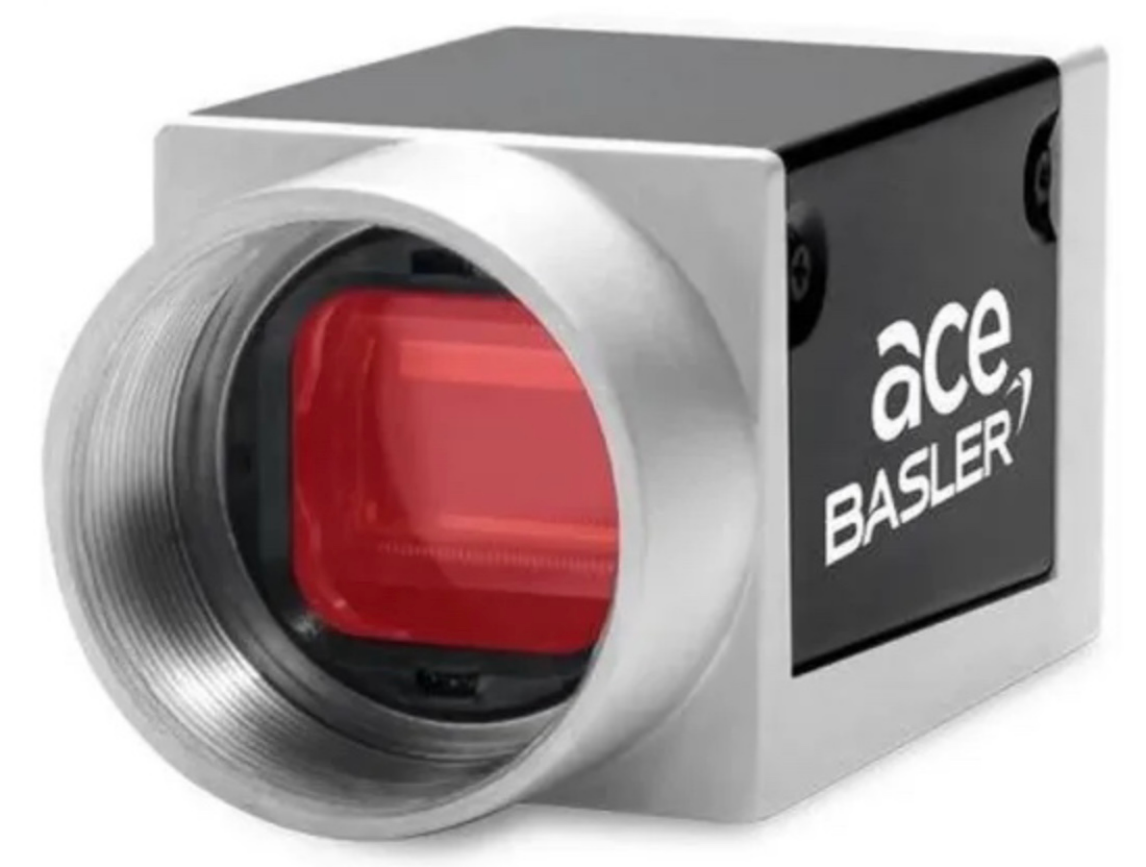}
        \end{minipage}
    }
    \subfigure[LiDAR]{
        \begin{minipage}{0.135\textwidth}
            \includegraphics[width=\textwidth, trim={0, -10, 0, 0}]{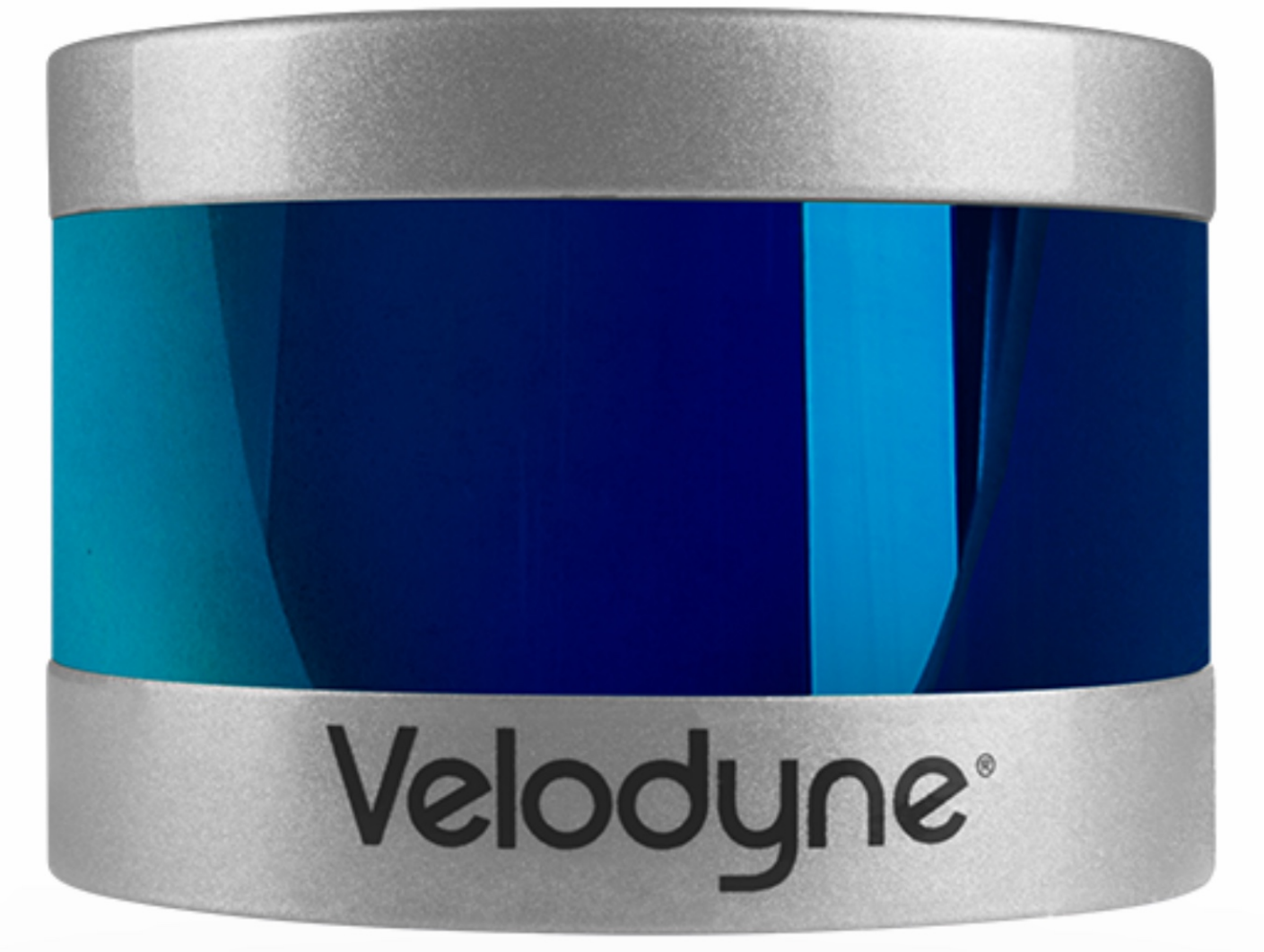}
        \end{minipage}
    }
    \subfigure[Radar]{
        \begin{minipage}{0.135\textwidth}
            \includegraphics[width=\textwidth, trim={0, -5, 0, 0}]{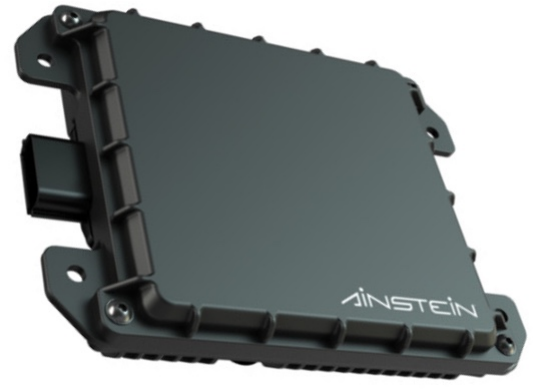}
        \end{minipage}
    }
    \\
    \subfigure[Event camera]{
        \begin{minipage}{0.135\textwidth}
            \includegraphics[width=\textwidth, trim={20, -25, 30, 0}]{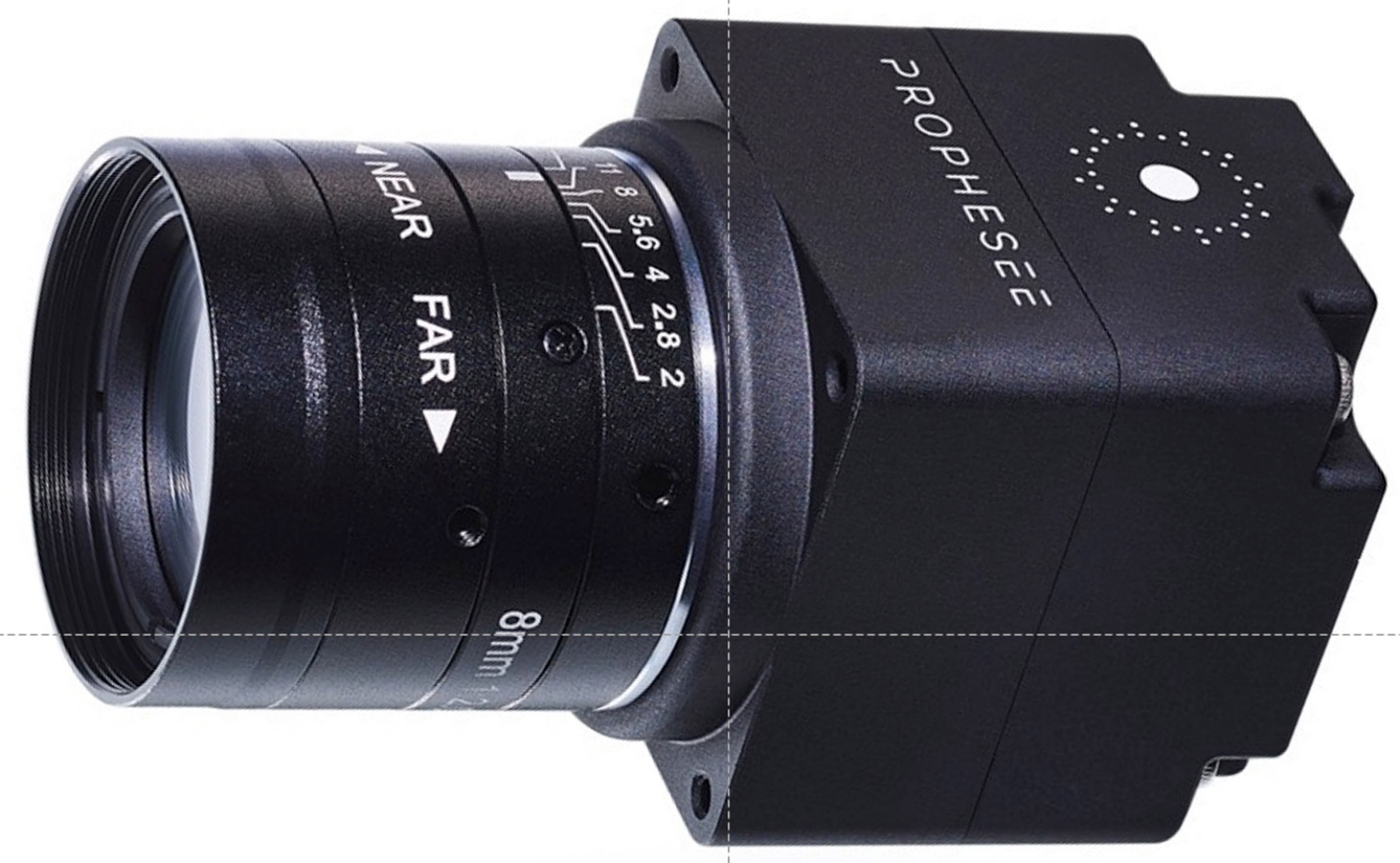}
        \end{minipage}
    }
    \subfigure[IMU]{
        \begin{minipage}{0.135\textwidth}
            \includegraphics[width=\textwidth, trim={0, -25, 30, 20}]{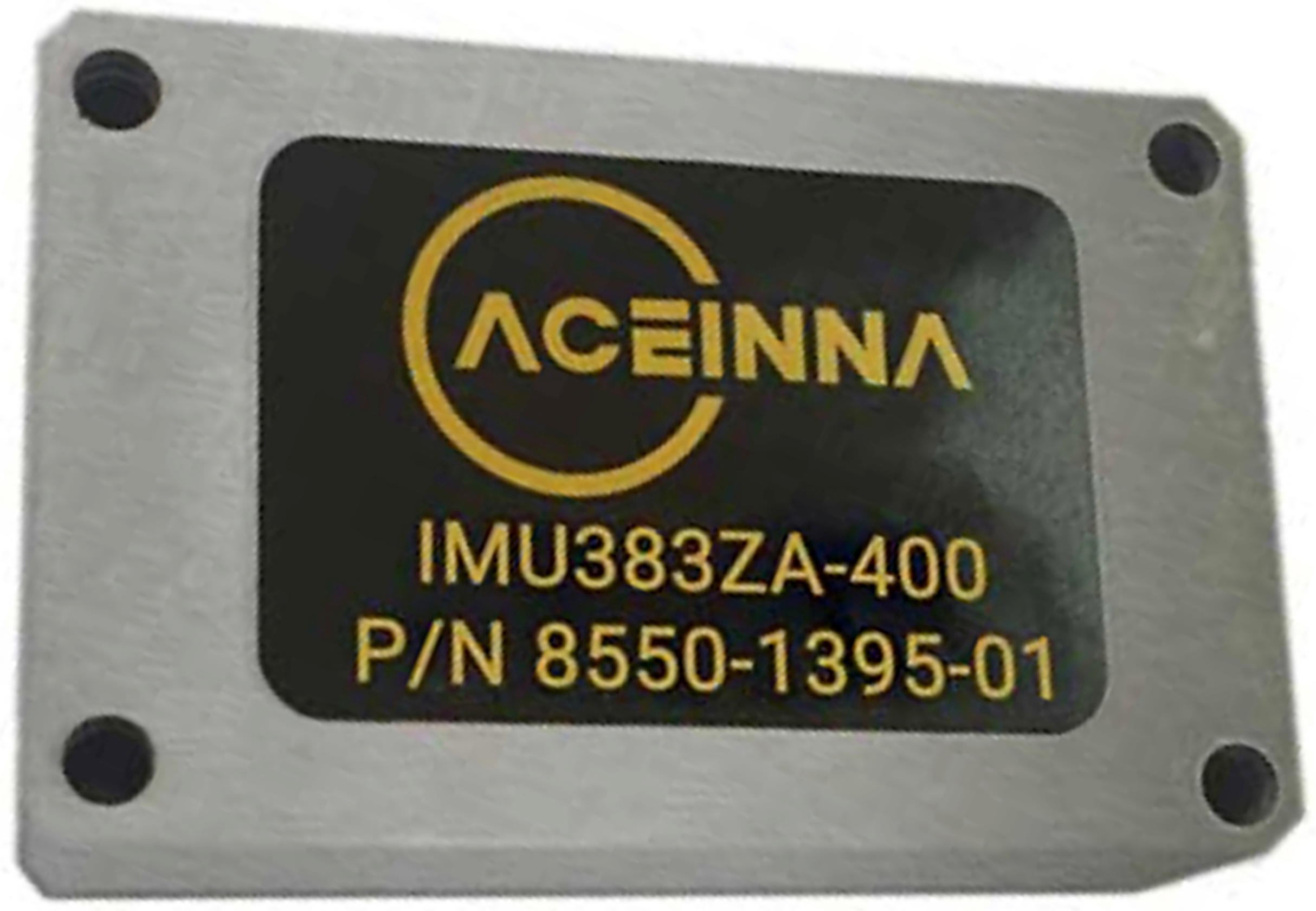}
        \end{minipage}
    }
    \subfigure[Thermal camera]{
        \begin{minipage}{0.125\textwidth}
            \includegraphics[width=\textwidth, trim={0, 0, 0, 10}]{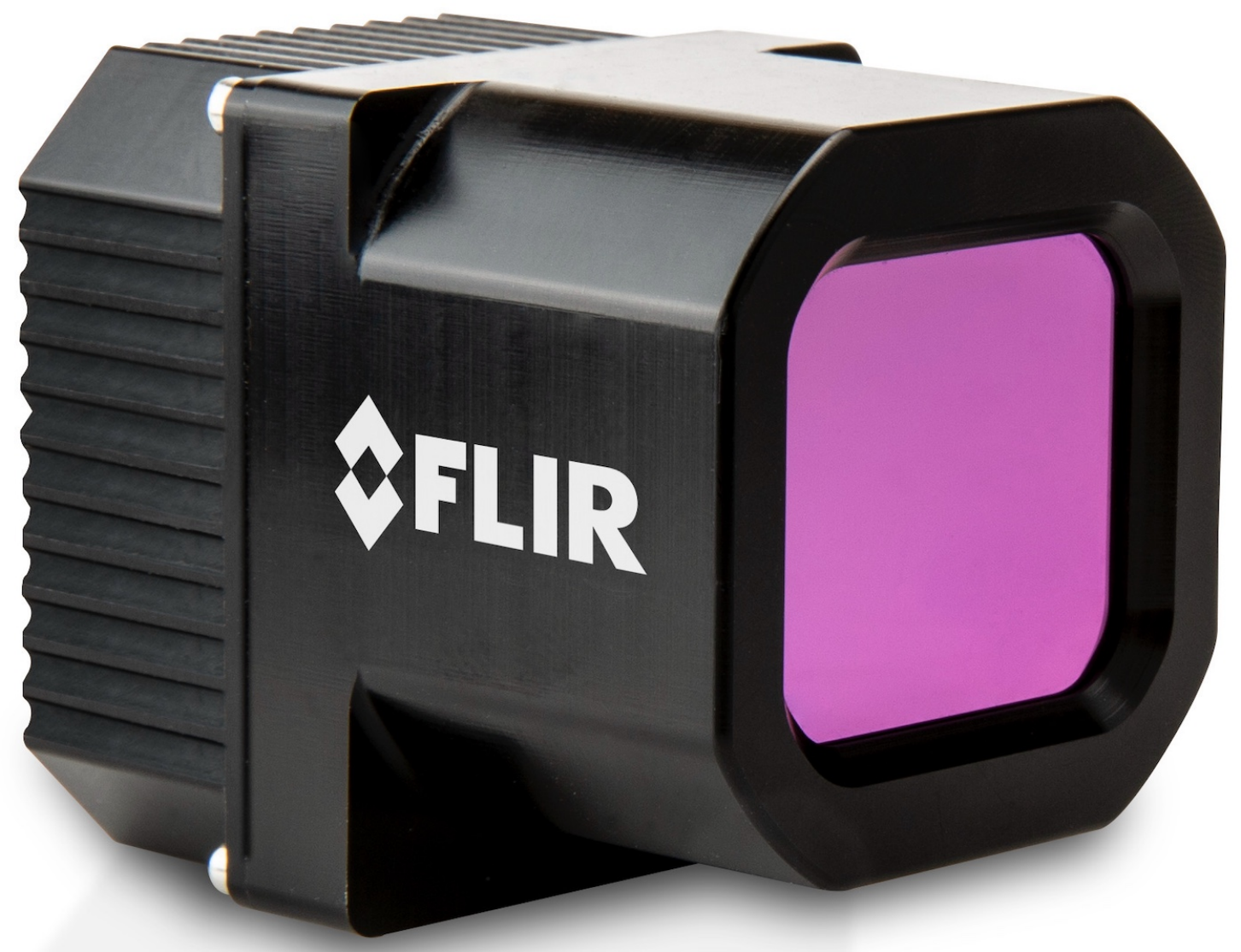}
        \end{minipage}
    }
    \caption{Sensors on autonomous driving vehicles. The type of each sensor is (a) Camera: \textit{Basler ace acA1600-20uc}, (b) LiDAR: \textit{Velodyne Puck LITE}, (c) Radar: \textit{Ainstein Launches K-79}, (d) Event-based camera: \textit{Evaluation Kit 4 HD}, (e) IMU: \textit{IMU383\_Aceinna-W} and (f) Thermal camera: \textit{FLIR\_2nd\_Gen\_ADK}. All figures are extracted from the websites hosting the sensors.}
    \label{fig:sensors_example}
\end{figure}

\begin{figure}[htb]
    \centering
    \includegraphics[width=0.48\textwidth, trim={10, 20, 10, 10}]{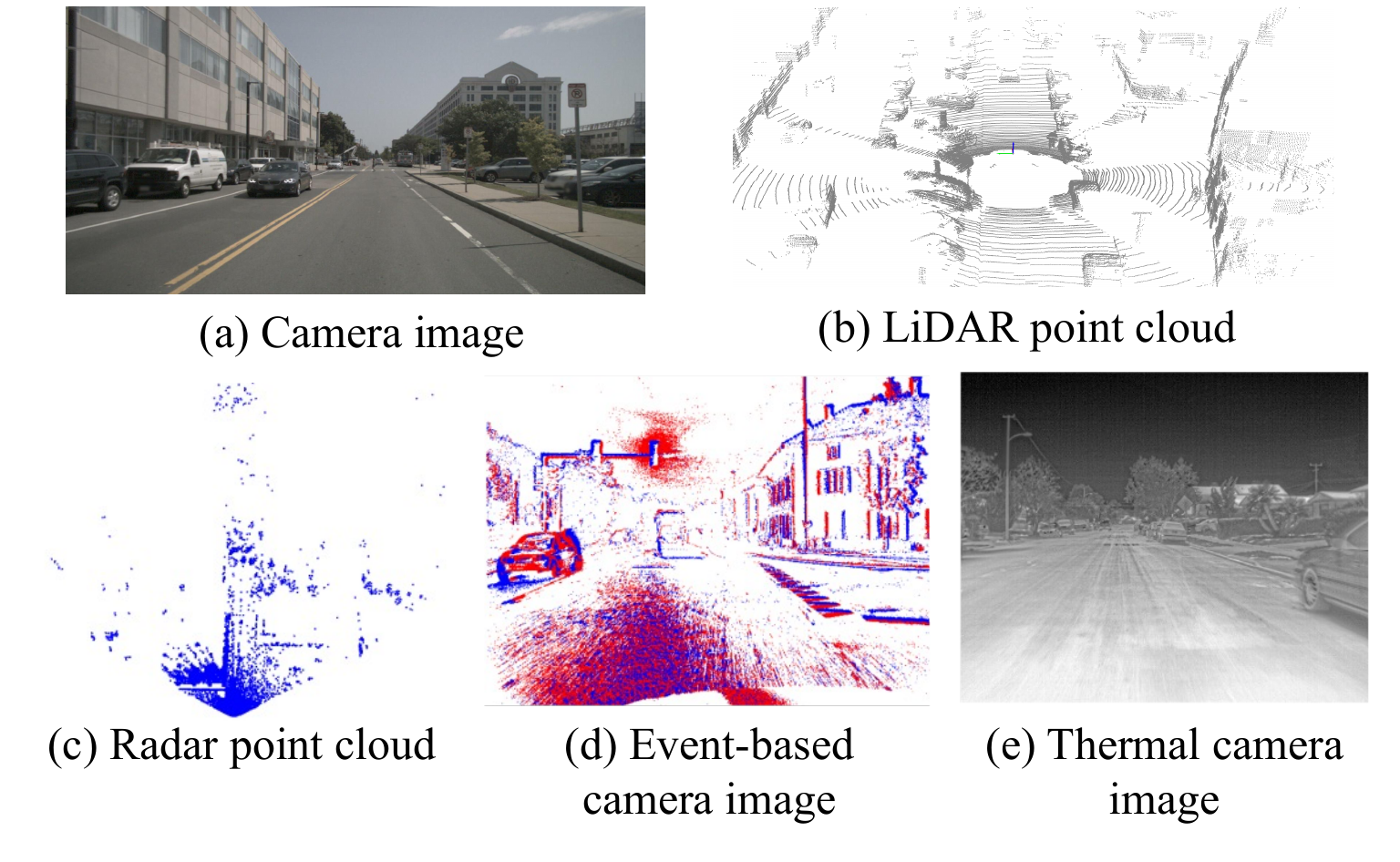}
    \caption{We present the sensor modalities to provide an intuitive understanding of each sensor's characteristics. (a) is from nuScenes~\cite{caesar2020nuscenes}, (b) is from KITTI~\cite{geiger2012we}, (c) is from \cite{weng2023all}, (d) is from \cite{gehrig2021dsec}, (e) is from \cite{flir}. All figures are collected from the open-source data of datasets.}
    \label{fig:sensor_modalities}
\end{figure}
Efficiently and precisely collecting data from the surrounding environment is the key to a reliable perception system in autonomous driving. To achieve this goal, various types of sensors are utilized on self-driving vehicles and infrastructures. The sensor examples are shown in Fig.~\ref{fig:sensors_example}. The most used sensors are cameras, LiDARs, and radars. Event-based and thermal cameras are also installed on vehicles or roadside infrastructure to improve the perception capability further.

\noindent \textbf{RGB Images.}
RGB images are usually recorded by monocular, stereo, or fisheye cameras. Monocular cameras offer a 2D view without depth; stereo cameras, with their dual lenses, provide depth perception; fisheye cameras use wide-angle lenses to capture a broad view. 
As shown in Fig.~\ref{fig:sensor_modalities} (a), the 2D images capture color information, rich textures, patterns, and visual details of the environment. Due to these characters, RGB images are mainly used to detect vehicles and pedestrians and recognize road signs. However, the RGB images are vulnerable to conditions like low illumination, rain, fog, or glare~\cite{grigorescu2020survey}.

\noindent \textbf{LiDAR Point Clouds.}
LiDARs use laser beams to measure the distance between the sensor and an object, creating a 3D environment representation~\cite{li2020lidar}. LiDAR point clouds (Fig.~\ref{fig:sensor_modalities} (b)) provide precise spatial information with high resolution and can detect objects over long distances. However, the density of these points can decrease with increasing distance, leading to sparser representations for distant objects. Weather conditions, e.g., fog, also limit the performance of LiDAR. In general, LiDARs are suitable in cases that require 3D concise information.

\noindent \textbf{Radar Point Clouds.}
Radars detect objects, distance, and relative speed by emitting radio waves and analyzing their reflection. Moreover, radars are robust under various weather conditions~\cite{zhou2020mmw}. Nevertheless, radar point clouds are generally coarser than LiDAR data, lacking the detailed shape or texture information of objects. Therefore, radars are generally used to assist other sensors. Fig.~\ref{fig:sensor_modalities} (c) exhibits the radar point clouds.

\noindent \textbf{Event of Event-based Camera.}
Event-based cameras asynchronously capture data, activating only when a pixel detects a change in brightness. The captured data is called events (Fig.~\ref{fig:sensor_modalities} (d)). Thanks to the specific data generation method, the recorded data has extremely high temporal resolution and captures fast motion without blur~\cite{chen2020event}.

\noindent \textbf{Infrared Images of Thermal Camera.}
Thermal cameras (see Fig.~\ref{fig:sensor_modalities} (e)) detect heat signatures by capturing infrared radiation~\cite{gade2014thermal}. Due to producing images based on temperature differences, thermal cameras can work in total darkness and are unaffected by fog or smoke. However, thermal cameras cannot discern colors or detailed visual patterns evident. Furthermore, the resolution of infrared images is lower compared to optical cameras. 

\noindent \textbf{Inertial Measurement Unit (IMU).}
An IMU is an electronic device that measures and reports an object's specific force, angular rate, and sometimes magnetic field surrounding the object~\cite{yurtsever2020survey}. In autonomous driving, it is used to track the movement and orientation of the vehicle. 

\begin{figure}[htb]
    \centering
    \includegraphics[width=0.48\textwidth, trim={0, 5, 0, 5}]{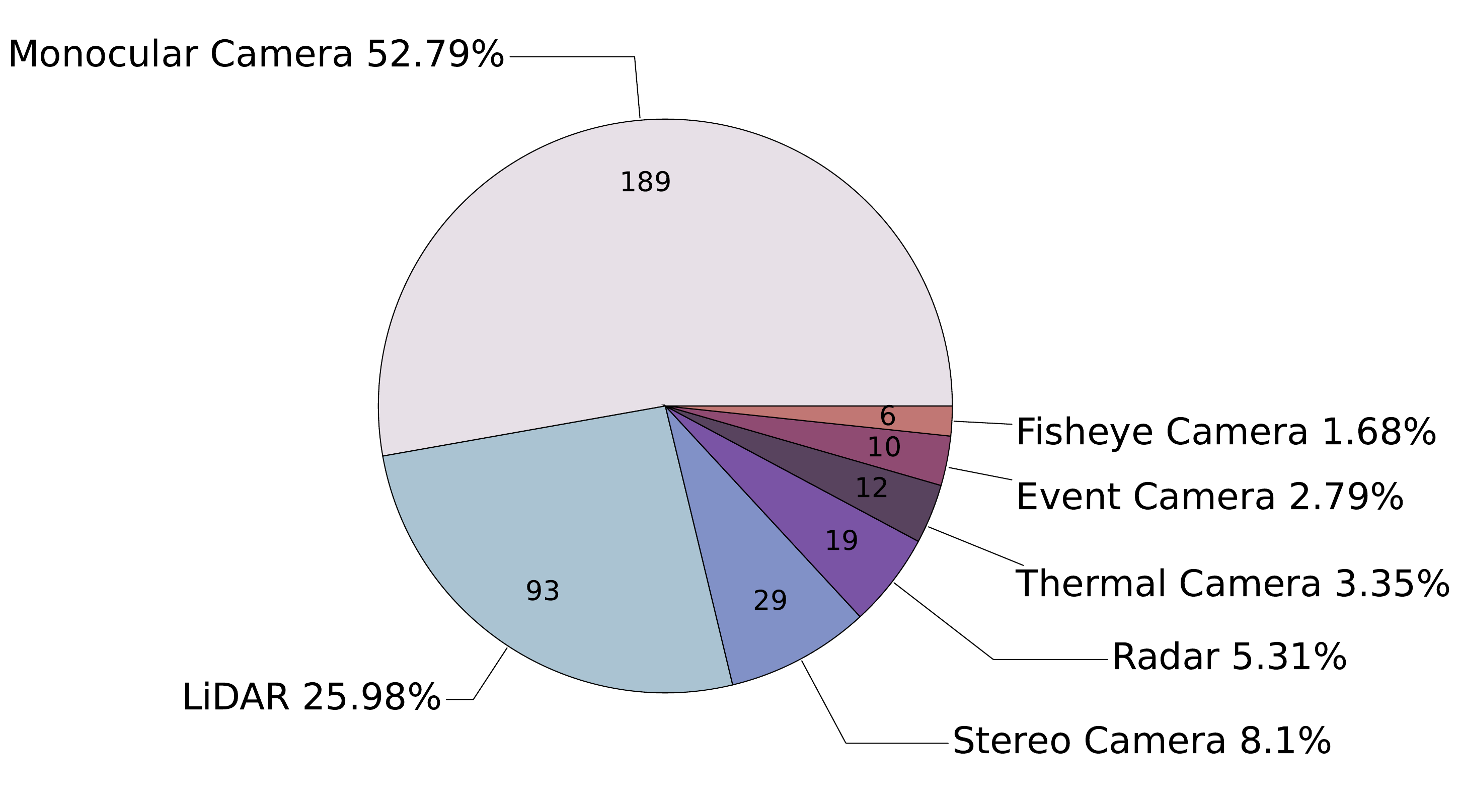}
    \caption{Sensor type distribution. We show the distribution of different sensors. Overall, RGB cameras and LiDARs are the most used sensors in autonomous driving datasets.}
    \label{fig:sensor_dist}
\end{figure}

We analyze the sensor distribution from the collected datasets, shown in Fig. \ref{fig:sensor_dist}. More than half of the sensors are monocular cameras (52.79\%) due to their low price and reliable performance. Additionally, 93 datasets (25.98\%) include LiDAR data, valued for its high resolution and precise spatial information. However, its high-cost limits LiDAR's widespread use. Beyond LiDAR point clouds, 29 datasets leverage stereo cameras to capture depth information. Furthermore, 5.31\%, 3.35\%, and 1.68\% datasets include radar, thermal camera, and fisheye camera. Given the temporal efficiency of capturing dynamic scenes, ten datasets generate data based on event-based cameras (2.79\%).

\subsection{Sensing Domains and Cooperative Perception Systems}
\label{sensing_domain_coperception_system}
Sensory data and cooperation between the ego vehicle and surrounding entities are pivotal for ensuring autonomous driving systems' safety, efficiency, and overall functionality. Therefore, the positioning of sensors is crucial as it determines the quality, angle, and scope of data that can be collected. Generally, sensing domains in autonomous driving can be categorized as onboard, Vehicle-to-Everything (V2X), drone-based, and others.

\begin{figure}[htb]
    \centering
    \includegraphics[width=0.48\textwidth, trim={10 10 10 10}]{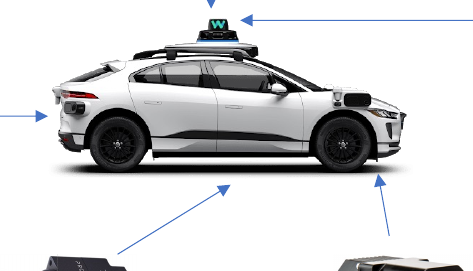}
    \caption{Overview of cooperative perception systems in autonomous driving. A complete autonomous driving perception system consists of the ego vehicle along with collaborative interactions between vehicles, infrastructure, and networks.}
    \label{fig:co_perception}
\end{figure}

\noindent \textbf{Onboard.}
Onboard sensors are installed directly on the autonomous vehicle and usually consist of cameras, LiDARs, radars, and IMUs. These sensors provide a direct perspective from the vehicle's standpoint, offering immediate feedback on the surroundings. Nevertheless, due to the limited detection scope, onboard sensors may have limitations in providing advanced warnings about obstacles in blind spots or detecting hazards around sharp bends.

\noindent \textbf{V2X.}
Vehicle-to-Everything encompasses communications between a vehicle and any other components in the transport system~\cite{huang2023v2x}, including Vehicle-to-Vehicle, Vehicle-to-Infrastructure, and Vehicle-to-Network (as shown in Fig.~\ref{fig:co_perception}). Beyond the immediate sensory input, the cooperative systems ensure multiple entities work harmoniously.

\begin{itemize}
    \item \textit{Vehicle-to-Vehicle (V2V)} \\
    V2V  enables nearby vehicles to share essential data, such as position, velocity, and captured sensory data (e.g., camera images or LiDAR scans). This shared information contributes to a more comprehensive understanding of driving scenarios.
    \item \textit{Vehicle-to-Infrastructure (V2I)} \\
    V2I facilitates communications between the autonomous vehicle and infrastructure components such as traffic lights, signs, or roadside sensors. The sensors implemented in road infrastructure collaborate to enhance the perception range and situational awareness of autonomous vehicles. In this survey, we categorize interactions involving single or multiple vehicles with single or multiple infrastructure components, as well as collaborations among multiple infrastructure elements, under V2I.
    \item \textit{Vehicle-to-Network (V2N)} \\
    V2N refers to exchanging information between a vehicle and a broader network infrastructure, often leveraging cellular networks to provide vehicles with access to cloud data. V2N aids the cooperation perception of V2V and V2I by sharing cross-area data or offering real-time updates about traffic congestion or road closures. 
    
\end{itemize}

\noindent \textbf{Drone.}
Drones, or Unmanned Aerial Vehicles (UAVs), offer an aerial perspective, providing data essential for trajectory prediction and route planning~\cite{krajewski2018highd}. For example, the real-time data from drones can be integrated into traffic management systems to optimize the traffic flow and alert autonomous vehicles of accidents ahead.

\noindent \textbf{Others.}
Data not collected by the previous three types is defined as others, such as other devices installed on non-vehicle objects or multiple domains.
\section{Tasks in Autonomous Driving}
\label{tasks_ad}
\begin{figure}
    \centering
    \includegraphics[width=0.48\textwidth, trim={20, 25, 20, 25}]{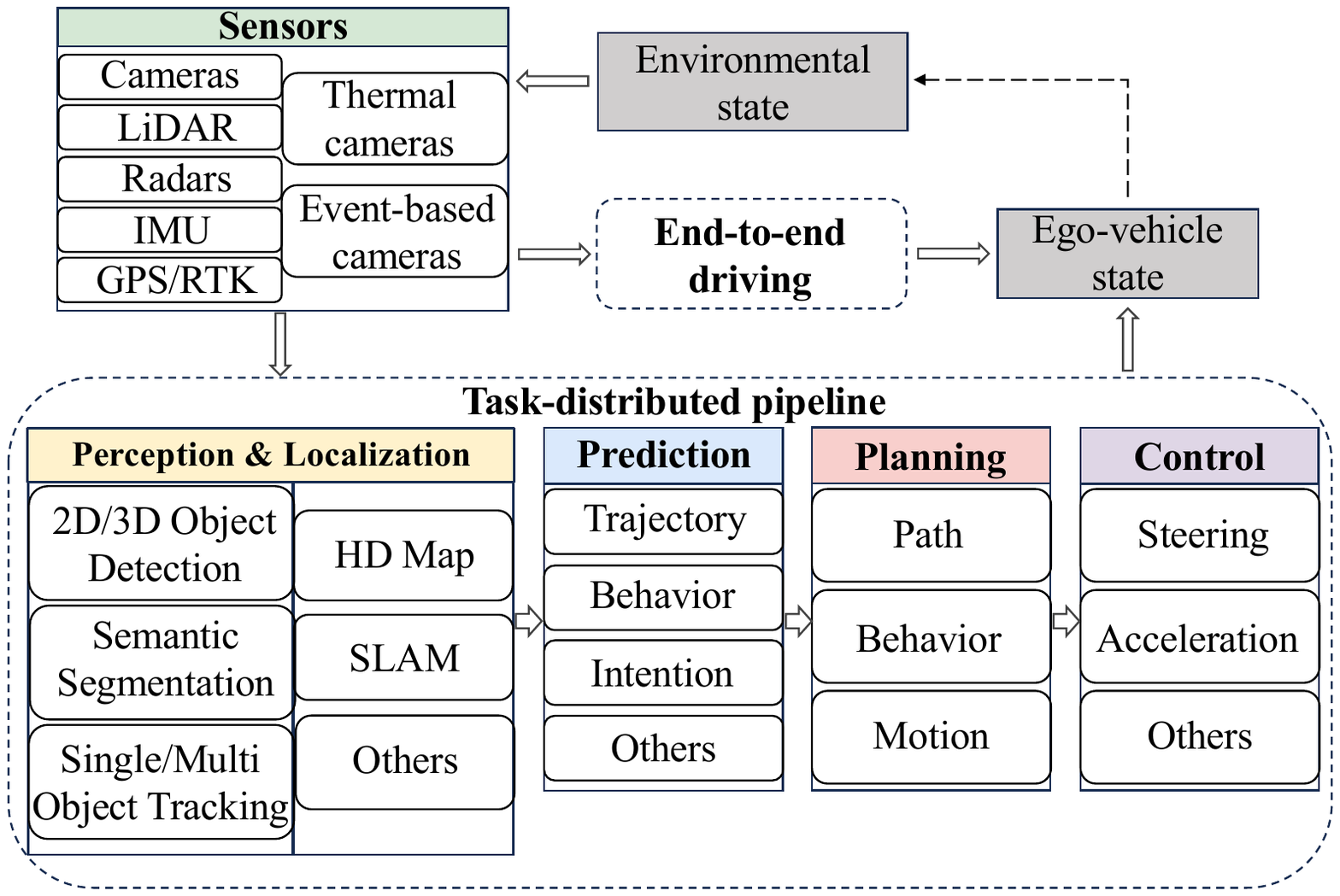}
    \caption{The overview of autonomous driving pipeline. Autonomous driving systems can be categorized into two types: modular-based and end-to-end. Both types rely on data collected by various sensors installed on vehicles or infrastructures. These systems interact with and respond to the surrounding environment during driving scenarios.}
    \label{fig:ad_pipeline}
\end{figure}

\begin{figure*}[htb]
    \centering

    \subfigure[3D Object detection]{
        \begin{minipage}[b]{0.231\textwidth}
        \includegraphics[height=2.2cm, width=\linewidth]{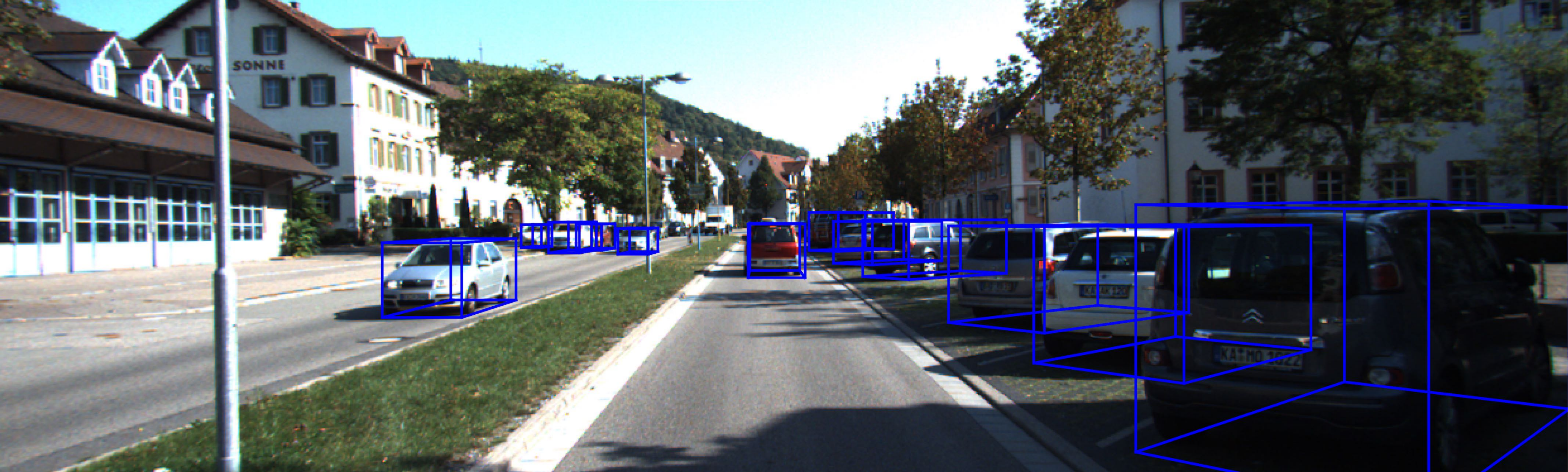}
        \end{minipage}
    }
    \subfigure[Semantic segmentation]{
        \begin{minipage}[b]{0.231\textwidth}
        \includegraphics[height=2.2cm, width=\linewidth]{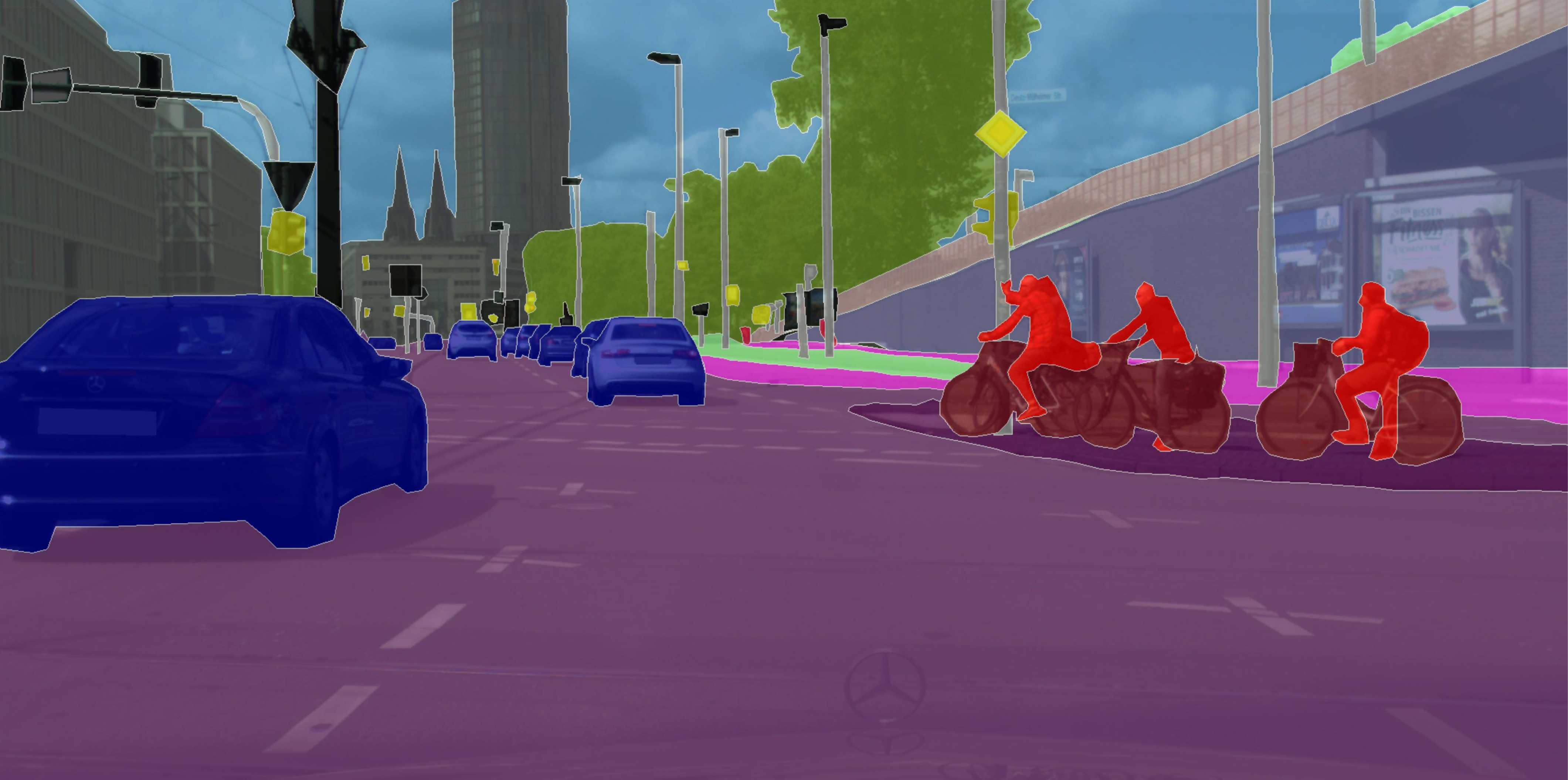}
        \end{minipage}
    }
    \subfigure[Trajectory prediction]{
        \begin{minipage}[b]{0.231\textwidth}
        \includegraphics[height=2.2cm, width=\linewidth]{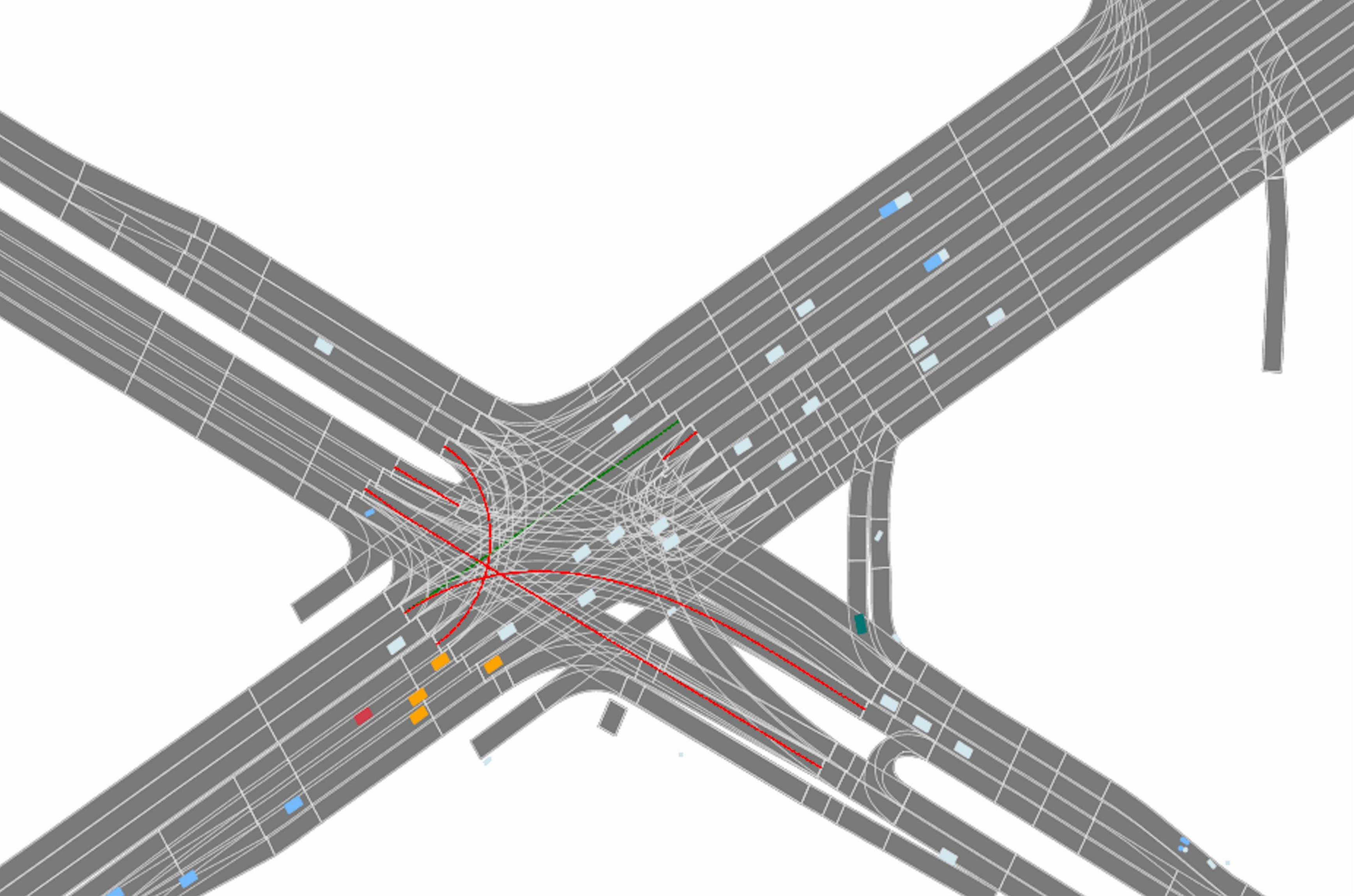}
        \end{minipage}
    }
    \subfigure[Lane detection]{
        \begin{minipage}[b]{0.231\textwidth}
        \includegraphics[height=2.2cm, width=\linewidth]{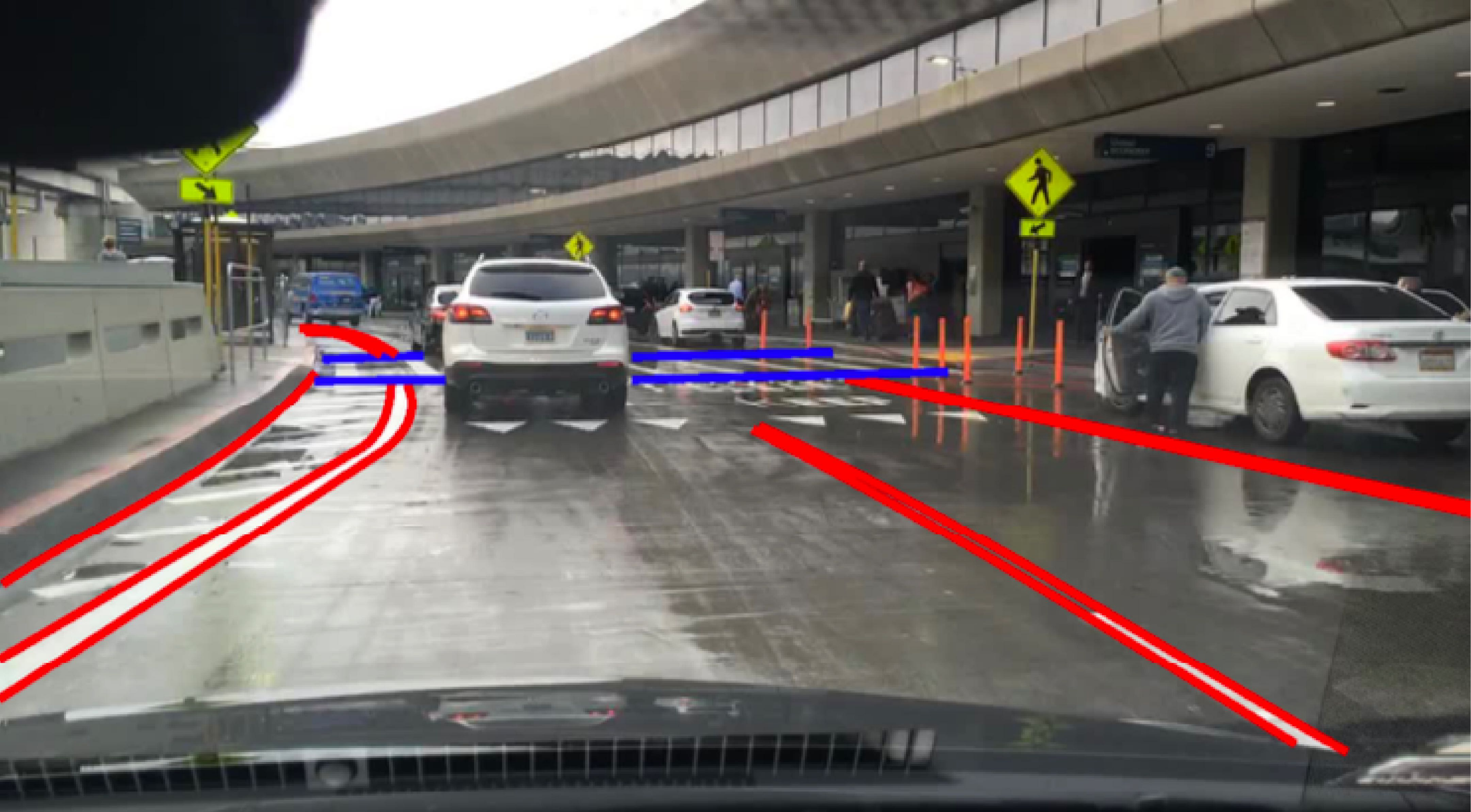}
        \end{minipage}
    }
        
    \subfigure[Visual referring]{
        \begin{minipage}[b]{0.231\textwidth}
        \includegraphics[height=2.2cm, width=\linewidth]{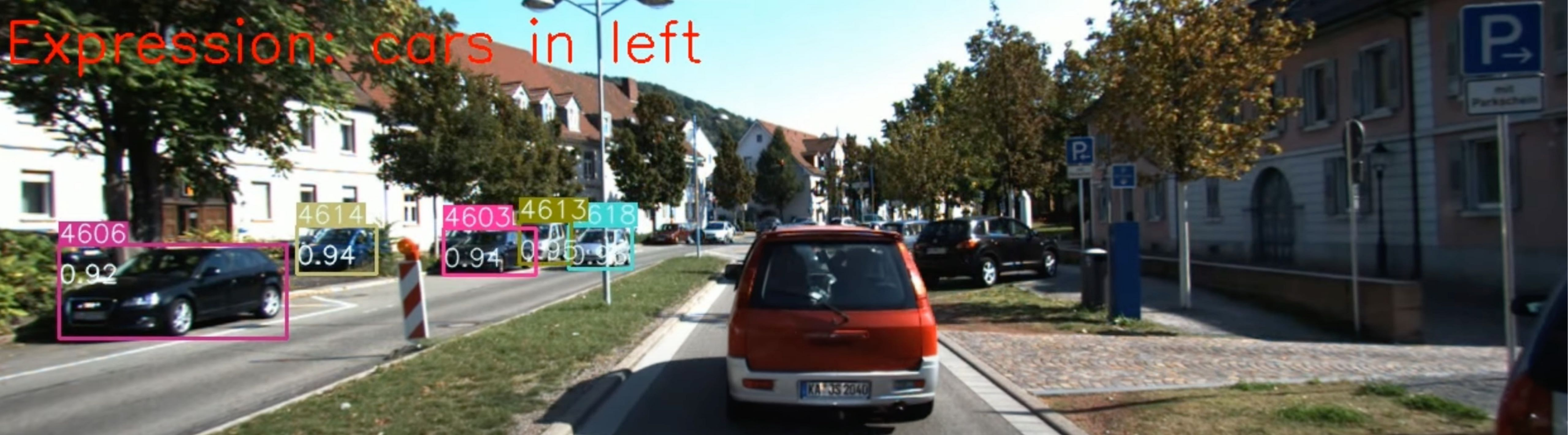}
        \end{minipage}
    }
    \subfigure[Point cloud segmentation]{
        \begin{minipage}[b]{0.231\textwidth}
        \includegraphics[height=2.2cm, width=\linewidth]{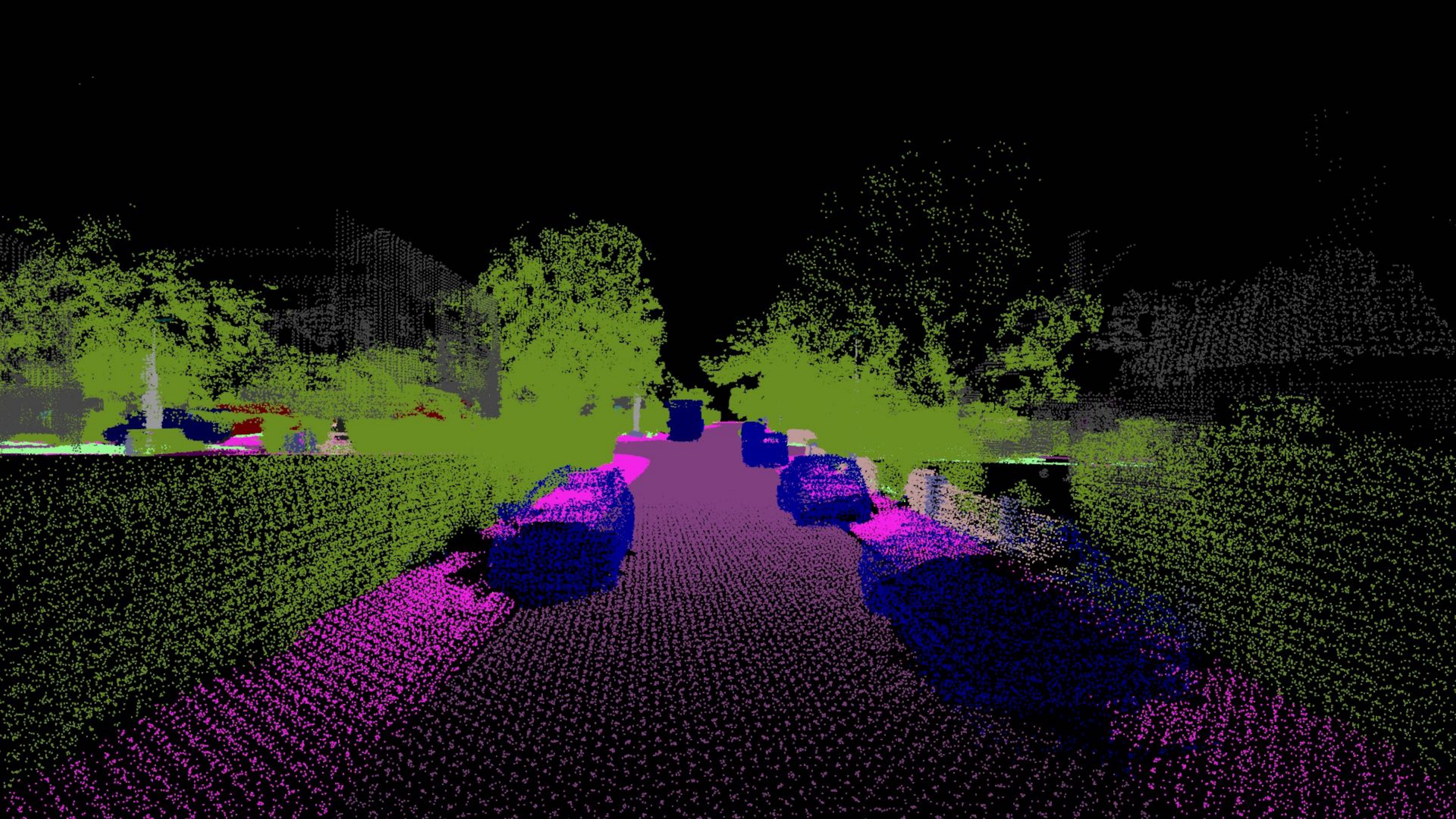}
        \end{minipage}
    }
    \subfigure[Driver attention prediction]{
        \begin{minipage}[b]{0.231\textwidth}
        \includegraphics[height=2.2cm, width=\linewidth]{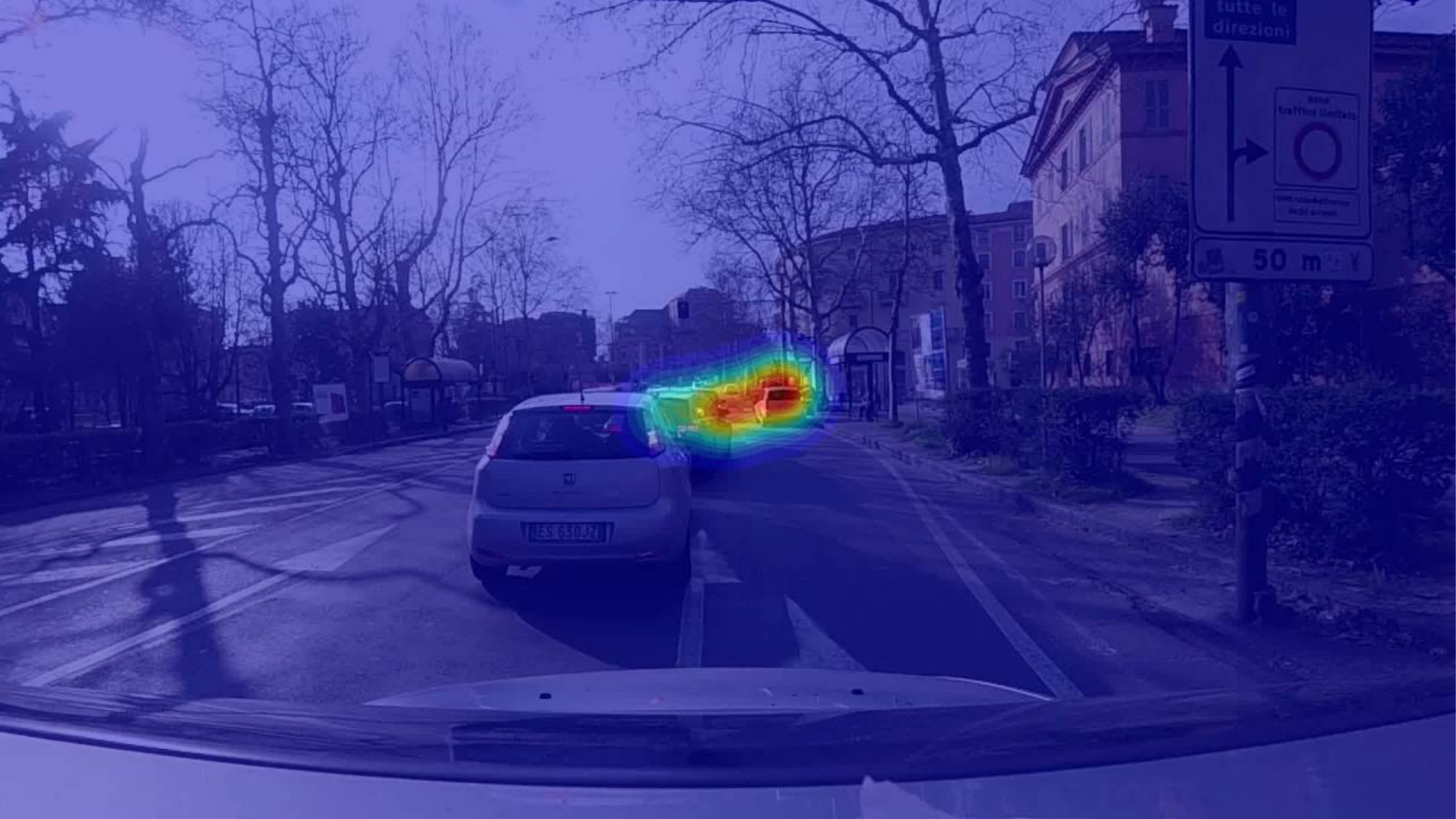}
        \end{minipage}
    }
    \subfigure[HD map]{
        \begin{minipage}[b]{0.231\textwidth}
        \includegraphics[height=2.2cm, width=\linewidth]{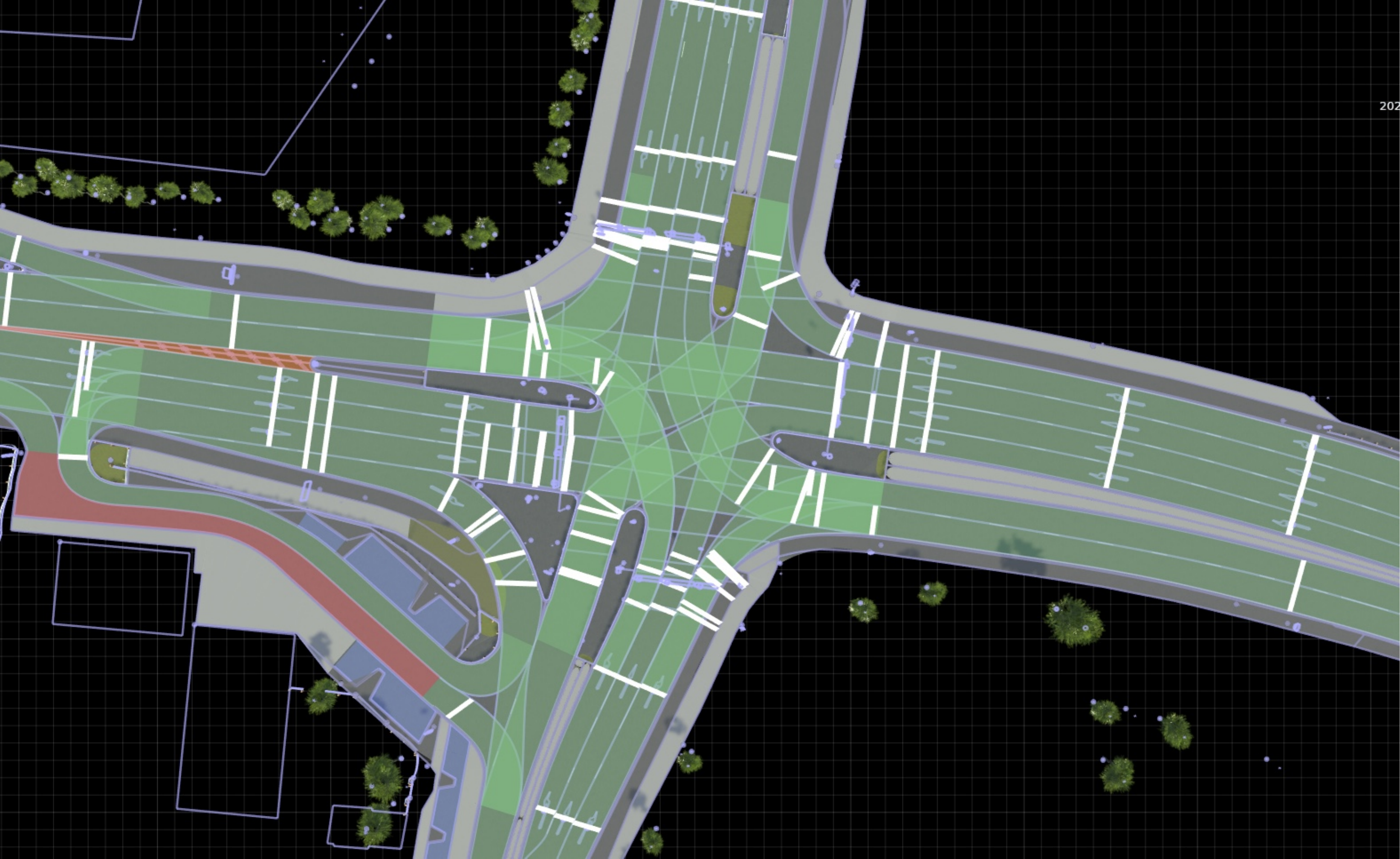}
        \end{minipage}
    }
    \caption{Examples of various autonomous driving tasks. (a) is from KITTI~\cite{geiger2012we}, (b) is from Cityscapes~\cite{cordts2016cityscapes}, (c) is from V2X-Seq~\cite{yu2023v2x}, (d) is from BDD100K~\cite{yu2020bdd100k}, (e) is from Refer-KITTI~\cite{wu2023referring}, (f) is from KITTI-360~\cite{liao2022kitti}, (g) is from Dr(eye)ve~\cite{palazzi2018predicting}, (h) is from TUMTraf~\cite{zimmer2024tumtrafv2x}. All figures are collected from the open-source data of datasets or the websites hosting the datasets.}
    \label{fig:ad_task_example}
\end{figure*}

This section offers insight into the pivotal tasks in autonomous driving, such as perception and localization (\ref{perception_localization}), prediction (\ref{prediction}), and planning and control (\ref{planning_control}). The overview of the autonomous driving pipeline is demonstrated in Fig.~\ref{fig:ad_pipeline}. We detail their objectives, the nature of data they rely upon, and inherent challenges. Fig.~\ref{fig:ad_task_example} highlights several main tasks in autonomous driving.
\subsection{Perception and Localization}
\label{perception_localization}
Perception focuses on understanding the environment based on the sensory data, while localization determines the autonomous vehicle's position within that environment.  

\noindent \textbf{2D/3D Object Detection.}
2D or 3D object detection aims to identify the locations and classification of other entities within the driving environment.  
Although the detection technologies have significantly advanced, several challenges remain, such as object occlusions, varying light conditions, and diverse object appearances.

Usually, the Average Precision (AP) metric~\cite{lin2014microsoft} is applied to evaluate the object detection performance. According to~\cite{mao20233d}, the AP metric can be formulated as
\begin{equation}
    AP = \int_{0}^{1} \max\left\{p\left(r'|r' \geq r\right)\right\}dr
    \label{ap_metric}
\end{equation}
where $p(r)$ is the precision-recall curve.

\noindent \textbf{2D/3D Semantic Segmentation.}
Semantic segmentation involves classifying each pixel of an image or point of a point cloud to its semantic category. From a dataset perspective, maintaining fine-grained object boundaries while managing extensive labeling requirements presents significant challenges for this task.

As mentioned in~\cite{garcia2018survey}, the main metrics used for segmentation are mean Pixel Accuracy (mPA):
\begin{equation}
   mPA = \frac{1}{k+1}\sum_{i=0}^{k}\frac{p_{ii}}{\sum_{j=0}^{k}p_{ij}}
\end{equation}
And the mean Intersection over Union (mIoU):
\begin{equation}
    mIoU = \frac{1}{k+1}\sum_{i=0}^{k}\frac{p_{ii}}{\sum_{j=0}^{k}p_{ij} + \sum_{j=0}^{k}p_{ji} - p_{ii}}
\end{equation}
Where $k\in \mathbb{N}$ is the number of classes, and $p_{ii}$, $p_{ij}$, and $p_{ji}$ represent true positives, false positives, and false negatives, respectively.

\noindent \textbf{Object Tracking.}
Object Tracking monitors the trajectories of a single or multiple objects over time. This task necessitates time-series RGB data, LiDAR, or radar sequences. Usually, object tracking includes single-object tracking or multi-object tracking (MOT).

Multi-Object-Tracking Accuracy (MOTA) is a widely utilized metric for multiple object tracking, which combines false negatives, false positives, and mismatch rate~\cite{luo2021multiple} (see Eq.~\ref{mota_metric}).
\begin{equation}
    MOTA = 1- \frac{\sum_{t}\left(fp_{t} + fn_{t} + e_{t}\right)}{\sum_{t} gt_{t}}
    \label{mota_metric}
\end{equation}
where $fp$, $fn$, and $e$ are the number of false positives, false negatives, and mismatch errors over time $t$. $gt$ is the ground truth.

Furthermore, instead of considering a single threshold, Average MOTA (AMOTA) is calculated based on all object confidence thresholds~\cite{guo2022review}.

\noindent \textbf{HD Map.}
HD mapping aims to construct detailed, highly accurate representations that include information about road structures, traffic signs, and landmarks. A dataset should provide LiDAR data for precise spatial information and camera data for visual details to ensure established map accuracy. According to~\cite{wilson2023argoverse}, HD map automation~\cite{li2022hdmapnet} and HD map change detection~\cite{lambert2022trust} have received more and more attention. Usually, the HD map quality is estimated using the accuracy metric.

\noindent \textbf{SLAM.}
Simultaneous Localization And Mapping (SLAM) entails building a concurrent map of the surrounding environment and localizing the vehicle within this map. Hence, data from cameras, IMUs for position tracking, and real-time LiDAR point clouds are vital. Sturm et al.~\cite{sturm2012benchmark} introduces two evaluation metrics, relative pose error (RPE) and absolute trajectory error (ATE), for evaluating the quality of the estimated trajectory from the input RGB-D images.

\subsection{Prediction}
\label{prediction}
Prediction refers to forecasting the future states or actions of surrounding agents. This capacity ensures safer navigation in dynamic environments.
Several evaluation metrics are used for prediction~\cite{huang2022survey,  mozaffari2020deep}, such as Root Mean Squared Error (RMSE):
\begin{equation}
    RMSE = \sqrt{\frac{1}{N}\sum_{n=1}^{N}\left(T_{pred}^{n} - T_{gt}^{n}\right)^{2}}
\end{equation}
where $N$ is the total number of samples, $T_{pred}$ and $T_{gt}$ represent the predicted trajectory and ground truth.

Negative Log Likelihood (NLL) (see Eq.~\ref{nll_metric}) is another metric focusing on determining the correctness of the trajectory, which can be used to compare the uncertainty of different models~\cite{ding2019predicting}.
\begin{equation}
    NLL = -\sum_{c=1}^{C}n_{c}log\left(\hat{n}_{c}\right)
    \label{nll_metric}
\end{equation}
Where $C$ is the total classes, $n_{c}$ is the binary indicator of the correctness of prediction, and $\hat{n}_{c}$ is the corresponding prediction probability. 

\noindent \textbf{Trajectory Prediction.}
Based on time-series data from sensors like cameras and LiDARs, trajectory prediction pertains to anticipating the future paths or movement patterns of other entities~\cite{huang2022survey}, such as pedestrians, cyclists, or other vehicles.

\noindent \textbf{Behavior Prediction.}
Behavior prediction anticipates the potential actions of other road users~\cite{mozaffari2020deep}, e.g., whether a vehicle will change the lane. Training behavior prediction models rely on extensive annotated data due to entities' vast range of potential actions within various scenarios. 

\noindent \textbf{Intention Prediction.}
Intention prediction focuses on inferring the higher-level goals behind the actions of objects, involving a deeper semantic comprehension of the physical or mental activities of humans~\cite {sharma2022pedestrian}.
Because of the task's complexity, it requires data from perception sensors like cameras, traffic signals, and hand gestures.

\subsection{Planning and Control}
\label{planning_control}
\subsubsection{Planning}
Planning represents the decision-making process in reaction to the perceived environment and predictions. A classic three-level hierarchical planning framework comprises path, behavioral, and motion planning~\cite{palazzi2018predicting}.

\noindent \textbf{Path Planning}
Path planning, also known as route planning, involves setting long-term objectives. It is a high-level process of determining the best path to the destination. 

\noindent \textbf{Behavior Planning.}
Behavior planning sits at the mid-level of the framework and is related to decision-making, including lane changes, overtaking, merging, and intersection crossing. This process relies on the correct understanding and interaction with the behavior of other agents. 

\noindent \textbf{Motion Planning.}
Motion planning deals with the actual trajectory the vehicle should follow in real time, considering obstacles, road conditions, and the predicted behavior of other road agents. In contrast to path planning, motion planning generates appropriate paths to achieve local objectives~\cite{palazzi2018predicting}.

\subsubsection{Control}
Control mechanisms in autonomous driving govern how the self-driving car executes the decided path or behavior from the motion planning system and corrects tracking errors~\cite{paden2016survey}. It translates high-level commands into actionable throttle, brake, and steering commands.

\subsection{End-to-end Autonomous Driving}
End-to-end approaches in autonomous driving are those where a single deep learning model handles everything from perception to control, bypassing the traditional modular pipeline. Such models can often be more adaptive as they rely on adjusting the whole model through learning. Their inherent promise is in their simplicity and efficiency by reducing the need for hand-crafted components~\cite{tampuu2020survey}. However, implementing end-to-end models faces limitations such as extensive training data requirements, low interpretability, and inflexible modular tuning.

Large-scale benchmarking for end-to-end AD can be categorized into closed-loop and open-loop evaluation~\cite{chen2023end}. The closed-loop evaluation is based on a simulated environment, while open-loop evaluation involves assessing a system's performance against expert driving behavior from real-world datasets.

\section{High-Influence Datasets}
\label{influence_impact}
This section describes the milestone autonomous driving datasets in the field of perception (\ref{perception_datasets}), prediction, planning, and control (\ref{prediction_planning_control_datasets}). We also exhibit datasets for end-to-end autonomous driving (\ref{e2e_datasets}).

\begin{figure*}
    \centering
    \includegraphics[width=\textwidth]{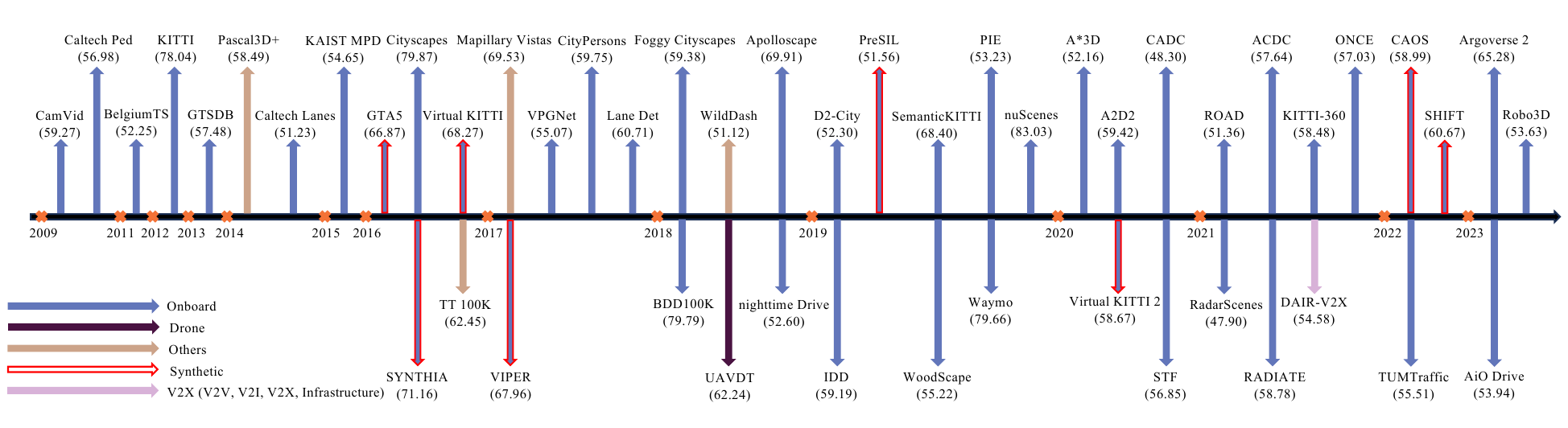}
    \vspace{-8mm}
    \caption{Chronological overview. We show the top 50 datasets ranked by their impact scores. This diagram covers datasets released from 2009 through March 17, 2024.}
    \label{fig:chronological_overview}
\end{figure*}

\begin{table*}[htb]
    \centering
    \caption{High-impact perception datasets. For a more comprehensive demonstration, we exhibit 50 perception datasets from different sensing domains instead of those with the highest scores. The citation number of each dataset is not shown in the table due to the fast change for this point.}
    \resizebox{\textwidth}{!}{
    \begin{tabular}{lcccccccccccccccccc}
        \toprule[1pt]
        \multirow{2}{*}{\textbf{Dataset}} & \multirow{2}{*}{\textbf{Year}} & \multirow{2}{*}{\textbf{Size}}  & \multirow{2}{*}{\textbf{Temp}} & \multicolumn{6}{c}{\textbf{Tasks}} & \textbf{Categories} & \textbf{Weather} & \multirow{2}{*}{\textbf{Time of day}} & \textbf{Scenario} & \textbf{Geographical} & \textbf{Impact} \\
        ~ & ~ & ~ & ~ & \textbf{2D Det} & \textbf{3D Det} & \textbf{2D Seg} & \textbf{3D Seg} & \textbf{Tracking} & \textbf{Lane Det} & \textbf{number} & \textbf{conditions} & ~ & \textbf{type} & \textbf{scope} & \textbf{score} \\
        \midrule[0.5pt]
        \multicolumn{16}{c}{Onboard} \\ 
        \midrule[0.5pt]
        nuScenes~\cite{caesar2020nuscenes} & 2019 & 40K & $\checkmark$ & $\checkmark$ & $\checkmark$ & $\checkmark$ & $\checkmark$ & $\checkmark$ & $\checkmark$ & 23 & 3 & 2 & 4 & 2 & 83.03 \\ 
        Cityscapes~\cite{cordts2016cityscapes} & 2016 & 25K & $\times$ & $\checkmark$ & $\checkmark$ & $\checkmark$ & & & & 30 & 1 & 1 & 3 & 1 & 79.87 \\ 
        BDD100K~\cite{yu2020bdd100k} & 2020 & 12M & $\checkmark$ & $\checkmark$ & & $\checkmark$ & & $\checkmark$ & $\checkmark$ & 40 & 5 & 2 & 4 & 1 & 79.79 \\ 
        Waymo~\cite{sun2020scalability} & 2019 & 230K & $\checkmark$ & $\checkmark$ & $\checkmark$ & $\checkmark$ & & $\checkmark$ & & 23 & 2 & 3 & 5 & 1 & 79.66 \\ 
        KITTI~\cite{geiger2012we} & 2012 & 41K & $\checkmark$ & $\checkmark$ & $\checkmark$ & & & & & 8 & 1 & 1 & 2 & 1 & 78.04 \\ 
        SYNTHIA~\cite{ros2016synthia} & 2016 & 13.4K & $\times$ & & & $\checkmark$ & & & & 13 & 2 & 3 & 5 & - & 71.16 \\ 
        Apolloscape~\cite{huang2018apolloscape} & 2018 & 143,906 & $\checkmark$ & $\checkmark$ & $\checkmark$ & $\checkmark$ & $\checkmark$ & $\checkmark$ & & 28 & 4 & 3 & 3 & 1 & 69.91 \\ 
        SemanticKITTI~\cite{behley2019semantickitti} & 2019 & 43,552 & $\checkmark$ & & & & $\checkmark$ & & & 28 & 1 & 1 & 4 & 1 & 68.40 \\ 
        Virtual KITTI~\cite{gaidon2016virtual} & 2016 & 21,260 & $\checkmark$ & $\checkmark$ & & $\checkmark$ & & $\checkmark$ & & 8 & 3 & 2 & 5 & - & 68.27 \\ 
        VIPER~\cite{richter2017playing} & 2017 & 254,064 & $\checkmark$ & $\checkmark$ & $\checkmark$ & $\checkmark$ & & $\checkmark$ & & 11 & 4 & 3 & 4 & - & 67.96 \\ 
        GTA5~\cite{richter2016playing} & 2016 & 24,966 & $\times$ & & & $\checkmark$ & & & & 19 & 2 & 2 & 2 & - & 66.87 \\ 
        Argoverse 2~\cite{wilson2023argoverse} & 2023 & 6M & $\checkmark$ & $\checkmark$ & $\checkmark$ & & & $\checkmark$ & & 30 & 2 & 1 & 1 & 6 & 65.28 \\ 
        Lane Det~\cite{pan2018spatial} & 2017 & 133,235 & $\times$ & & & & & & $\checkmark$ & 1 & 1 & 3 & 3 & 1 & 60.71 \\ 
        SHIFT~\cite{sun2022shift} & 2022 & 2,5M & $\checkmark$ & $\checkmark$ & $\checkmark$ & $\checkmark$ & & $\checkmark$ & & 23 & 5 & 5 & 3 & - & 60.67 \\ 
        CityPersons~\cite{zhang2017citypersons} & 2017 & 25K & $\times$ & $\checkmark$ & & $\checkmark$  & & & & 30 & 2 & 1 & - & 27 & 59.75 \\ 
        A2D2~\cite{geyer2020a2d2} & 2020 & 41,277 & $\times$ & $\checkmark$ & $\checkmark$ & $\checkmark$ & & & & 38 & 2 & 1 & 3 & 3 & 59.42 \\ 
        Foggy Cityscapes~\cite{sakaridis2018semantic} & 2018 & 20,550 & $\times$ & $\checkmark$ & & $\checkmark$ & & & & 19 & 1 & 1 & 2 & 1 & 59.38 \\ 
        CamVid~\cite{brostow2009semantic} & 2009 & 701 & $\times$ & & & $\checkmark$ & & & & 32 & 2 & 1 & 2 & 1 & 59.27 \\ 
        IDD~\cite{varma2019idd} & 2019 & 10,004 & $\times$ & & & $\checkmark$ & & & & 34 & 3 & 3 & 5 & 1 & 59.18 \\ 
        CAOS~\cite{hendrycks2019scaling} & 2022 & 13K & $\times$ & & & $\checkmark$ & & & & 13 & 3 & 2 & 3 & - & 58.98 \\ 
        RADIATE~\cite{sheeny2021radiate} & 2021 & 44,140 & $\checkmark$ & $\checkmark$ & & & & $\checkmark$ & & 8 & 5 & 2 & 4 & 1 & 58.78 \\ 
        Virtual KITTI 2~\cite{cabon2020virtual} & 2020 & 20,992 & $\checkmark$ & $\checkmark$ & $\checkmark$ & $\checkmark$ & & $\checkmark$ & & 8 & 4 & 2 & 3 & 1 & 58.67 \\ 
        KITTI-360~\cite{liao2022kitti} & 2021 & 150K & $\checkmark$ & $\checkmark$ & $\checkmark$ & $\checkmark$ & $\checkmark$ & $\checkmark$ & & 37 & 1 & 1 & 1 & 1 & 58.48 \\ 
        Dr(eye)ve~\cite{palazzi2018predicting} & 2018 & 555000 & $\times$ & & & $\checkmark$ & & & & 10 & 3 & 3 & 3 & - & 58.13 \\ 
        ACDC~\cite{sakaridis2021acdc} & 2021 & 4,006 & $\times$ & & & $\checkmark$ & & & & 19 & 4 & 2 & 3 & 1 & 57.64 \\
        GTSDB~\cite{houben2013detection} & 2013 & 900 & $\times$ & $\checkmark$ & & & & & & 4 & 2 & 2 & 3 & 1 & 57.48 \\ 
        ONCE~\cite{mao2021one} & 2021 & 1M & $\times$ & $\checkmark$ & $\checkmark$ & & & & &  5 & 3 & 2 & 5 & 1 & 57.03 \\
        Caltech Ped~\cite{dollar2009pedestrian} & 2009 & 250K & $\times$ & $\checkmark$ & & & & & & 1 & 1 & 1 & 1 & 2 & 56.98 \\ 
        STF~\cite{bijelic2020seeing} & 2020 & 13,500 & $\times$ & $\checkmark$ & $\checkmark$ & & & & & 1 & 4 & 2 & 3 & 4 & 56.85 \\ 
        \midrule[0.5pt]
        \multicolumn{16}{c}{V2X} \\ 
        \midrule[0.5pt]
        TUMTraf~\cite{cress2022a9,zimmer2023tumtrafintersection,cress2024tumtrafevent,zimmer2024tumtrafv2x} & 2022 & 50,253 & $\checkmark$ & $\checkmark$ & $\checkmark$ & $\checkmark$ & & $\checkmark$ & $\checkmark$ & 10 & 6 & 5 & 10 & 1 & 55.51 \\ 
        DAIR-V2X~\cite{yu2022dair} & 2021 & 71,254 & $\checkmark$ & $\checkmark$ & $\checkmark$ & & & & & 10 & 2 & 2 & 2 & 1 & 54.58 \\
        V2XSet~\cite{xu2022v2x} & 2022 & 11,447 & $\checkmark$ & $\checkmark$ & $\checkmark$ & & & & & 1 & 1 & 1 & 1 & - & 49.84 \\
        V2V4Real~\cite{xu2023v2v4real} & 2023 & 40K & $\checkmark$ & $\checkmark$ & $\checkmark$ & & & $\checkmark$ & & 5 & 2 & 1 & 2 & 1 & 40.28 \\ 
        Rope3D~\cite{ye2022rope3d} & 2022 & 50K & $\times$ & $\checkmark$ & $\checkmark$ & & & & & 12 & 3 & 3 & 2 & 1 & 48.96 \\ 
        V2X-Sim~\cite{li2022v2x} & 2022 & 10K & $\checkmark$ & $\checkmark$ & $\checkmark$ & $\checkmark$ & $\checkmark$ & $\checkmark$ & & 23 & 1 & 1 & 1 & - & 48.50 \\ 
        V2VNet~\cite{wang2020v2vnet} & 2020 & 51,2K & $\checkmark$ & $\checkmark$ & & & & & & 1 & 1 & 1 & 1 & - & 48.20 \\ 
        T\&J~\cite{chen2019cooper} & 2019 & 100 & $\checkmark$ & $\checkmark$ & $\checkmark$ & & & & & 1 & 1 & 1 & 2 & 1 & 46.63 \\ 
        Co-Percep~\cite{arnold2020cooperative} & 2020 & 10K & $\times$ & $\checkmark$ & $\checkmark$ & & & & & 1 & 1 & 1 & 1 & 1 & 42.99 \\
        DeepAccident~\cite{wang2023deepaccident} & 2023 & 285K & $\checkmark$ & & & & & & & 6 & 4 & 3 & 2 & - & 41.79 \\
        LUMPI~\cite{busch2022lumpi} & 2022 & 200K & $\checkmark$ & $\checkmark$ & $\checkmark$ & $\checkmark$ & & $\checkmark$ & & 3 & 6 & 3 & 1 & 1 & 40.42 \\
        \midrule[0.5pt]
        \multicolumn{16}{c}{Drone} \\ 
        \midrule[0.5pt]
        UAVDT~\cite{du2018unmanned} & 2018 & 80K & $\checkmark$ & $\checkmark$ & & & & $\checkmark$ & & 3 & 2 & 2 & 6 & 1 & 61.63 \\
        DroneVehicle~\cite{sun2022drone} & 2021 & 28,439 & $\times$ & $\checkmark$ & & & & & & 5 & 1 & 3 & 4 & 1 & 44.42 \\ 
        \midrule[0.5pt]
        \multicolumn{16}{c}{Others} \\ 
        \midrule[0.5pt]
        Mapillary Vistas~\cite{neuhold2017mapillary} & 2017 & 25K & $\times$ & & & $\checkmark$ & & & & 66 & 5 & 3 & 3 & 2 & 68.63 \\ 
        TT 100K~\cite{zhu2016traffic} & 2016 & 100K & $\times$ & $\checkmark$ & & & & & & 45 & 2 & 2 & 2 & 10 & 61.99 \\ 
        Pascal3D+~\cite{xiang2014beyond} & 2014 & 30,899 & $\times$ & $\checkmark$ & $\checkmark$ & $\checkmark$ & & & & 12 & 2 & 2 & 1 & - & 58.07 \\ 
        WildDash~\cite{zendel2018wilddash} & 2018 & 1,800 & $\times$ & & & $\checkmark$ & & & & 28 & 2 & 2 & 7 & 1 & 50.02 \\ 
        TorontoCity~\cite{wang2016torontocity} & 2016 & 56K & $\times$ & & & $\checkmark$ & & & & 4 & 2 & 2 & 1 & 1 & 45.82 \\ 
        DAWN~\cite{kenk2020dawn} & 2020 & 4,543 & $\times$ & $\checkmark$ & & & & & & 5 & 6 & 3 & 3 & - & 43.99 \\
        RAD~\cite{lis2019detecting} & 2019 & 60 & $\times$ & & & $\checkmark$ & & & & 19 & 1 & 1 & 1 & 1 & 37.86 \\
        STCrowd~\cite{cong2022stcrowd} & 2022 & 10,891 & $\checkmark$ & $\checkmark$ & $\checkmark$ & & & $\checkmark$ & & 1 & 3 & 1 & 1 & 1 & 35.19 \\
        \bottomrule[1pt]
    \end{tabular}}
    \label{tab:50_datasets}
\end{table*}

\subsection{Perception Datasets}
\label{perception_datasets}
Perception datasets are critical for developing and optimizing autonomous driving systems. They enhance vehicle reliability and robustness by providing rich, multimodal sensory data, ensuring effective perception and understanding of surroundings.

We leverage the proposed dataset evaluation metrics (see \ref{evaluation_metrics}) to calculate the impact scores of the collected perception datasets, subsequently selecting the top 50 datasets based on these scores to create a chronological overview, shown in Fig.~\ref{fig:chronological_overview}. Concurrently, as described in \ref{sensing_domain_coperception_system}, we categorize datasets into onboard, V2X, drone, and others, choosing a subset from each to compile a comprehensive table of 50 datasets (Tab.~\ref{tab:50_datasets}). It is worth noting that the datasets in the table are sorted by impact score within their respective categories and do not represent the overall top 50. In the following section, we choose several datasets with the highest impact scores within each sensing source and exhibit them, considering their published year.

\subsubsection{Onboard}
    \textbf{\textit{KITTI}.} 
    KITTI~\cite{geiger2012we} has been instrumental in advancing the autonomous driving domain since its release in 2012. KITTI data is
    recorded by various sensors, including cameras, LiDAR, and GPS/IMU, facilitating the development of algorithms for object detection, tracking, optical flow, depth estimation, and visual odometry. However, KITTI data was mainly recorded under ideal weather conditions in German urban areas. The relatively small geographic and environmental conditions scope limit its real-world applications.
    
    \textbf{\textit{Cityscapes}.}
    Cityscapes~\cite{cordts2016cityscapes} comprises a vast collection of images captured explicitly in intricate urban environments and has become a standard benchmark semantic segmentation. Cityscapes provides pixel-level segmentation through meticulous labeling for 30 object classes, including various vehicle types, pedestrians, roads, and traffic sign information. However, Cityscapes primarily focuses on German cities and lacks diverse weather conditions. These limitations hinder its usefulness for developing robust and generalized autonomous driving solutions.
    
    \textbf{\textit{VIPER}.}
    VIPER~\cite{richter2017playing} is a synthetic dataset collected based on driving, riding, and walking perspectives in a realistic virtual world. VIPER contains over 250K video frames annotated with ground-truth data for low- and high-level vision tasks. It encapsulates various weather conditions, lighting scenarios, and complex urban landscapes, making it an invaluable resource for testing the robustness of autonomous driving algorithms. Nonetheless, the transition from synthetic data to real-world application remains challenging, as algorithms must bridge the domain gap to maintain their effectiveness in real environments.
    
    \textbf{\textit{SemanticKITTI}.}
    SemanticKITTI~\cite{behley2019semantickitti} consists of over 43,000 LiDAR point cloud frames, making it one of the most extensive datasets for 3D semantic segmentation in outdoor environments. SemanticKITTI provides precise labels for 28 categories, such as car, road, building, etc., achieving a robust benchmark for semantic segmentation. However, similar to the previous datasets, SemanticKITTI faces limitations in real-world applicability due to its restricted environmental diversity and geographical scope.
    
    \textbf{\textit{nuScenes}.}
    nuScenes~\cite{caesar2020nuscenes} stands as an essential contribution to the field of autonomous driving, providing multimodal sensor setup, including LiDAR, radars, and cameras. It addresses the diversity in urban scenes and environmental conditions. The data recorded from Boston and Singapore cover varied driving behaviors and urban layouts, enhancing generalizability. Its six cameras provide a comprehensive perspective of the surrounding environment, making them widely utilized in multi-view object detection tasks. Yet, the dataset's coverage of rare driving cases, such as accidents, could be expanded to better assess algorithm performance under challenging conditions.
    
    \textbf{\textit{Waymo}.}
    The Waymo Open Dataset~\cite{sun2020scalability}, introduced in 2019, significantly influences research and advancement in autonomous driving. Waymo provides an extensive size of multimodal sensory data with high-quality annotations compared to others. Key contributions of the Waymo dataset include its comprehensive coverage of driving conditions and geographics, which are pivotal for the robustness and generability of different tasks. The real-world applicability of Waymo's data is enhanced by its diversity, although exploring its limitations in specific adverse conditions could provide deeper insights into areas needing improvement.
    
    \textbf{\textit{BDD100K}.}
    BDD100K~\cite{yu2020bdd100k} dataset is renowned for its size and diversity. It comprises 100,000 videos, each about 40 seconds in duration. Furthermore, it includes different times of day and weather conditions, offering a solid foundation for testing and improving the robustness of algorithms. Meanwhile, it provides various annotated labels for object detection, tracking, semantic segmentation, and lane detection. However, the varying quality of video annotations challenges the dataset's real-world applicability, highlighting the need for consistent, high-quality annotations to ensure algorithm effectiveness across all conditions.
    
    \textbf{\textit{Argoverse~2}.}
    As a sequel to Argoverse~1~\cite{chang2019argoverse}, Argoverse~2~\cite{wilson2023argoverse} introduces more diversified and complex driving scenarios, presenting the largest autonomous driving taxonomy to date. It captures various real-world driving scenarios across six cities and varying conditions. Argoverse~2 supports a wide range of essential tasks, including but not limited to 3D object detection, semantic segmentation, and tracking. The limitation of Argoverse~2 in the real world is similar to \cite{caesar2020nuscenes} and \cite{sun2020scalability}, considering more adversarial conditions and edge cases can extend its application field.

\subsubsection{V2X}

    \textbf{\textit{TUMTraf}.}
    The TUM Traffic Dataset family (\textit{TUMTraf}) is a cutting-edge real-world dataset comprising 50,253 labeled frames (9,545 point clouds and 40,708 images) across five releases, capturing diverse traffic scenarios in Munich, Germany \cite{cress2022a9,zimmer2023tumtrafintersection,cress2024tumtrafevent,zimmer2024tumtrafv2x}. It integrates multiple data modalities like RGB camera, event-based camera, LiDAR, GPS, and IMU. TUMTraf also provides perspectives from the infrastructure and vehicle, enabling research in cooperative perception. TUMTraf stands out for including edge cases like accidents, near-miss events, and traffic violations, providing a great resource to improve perception systems.
    
    \textbf{\textit{DAIR-V2X}.}
    DAIR-V2X~\cite{yu2022dair} is a pioneering resource in the Vehicle-Infrastructure Cooperative Autonomous Driving, providing large-scale, multi-modality, multi-view real-world data. The dataset is designed to tackle challenges such as the temporal asynchrony between vehicle and infrastructure sensors and the data transmission costs involved in such cooperative systems. DAIR-V2X has been instrumental in advancing vehicle-infrastructure cooperation, setting benchmarks for addressing V2X perception tasks.

\subsubsection{Drone}

    \textbf{\textit{UAVDT}.}
    The UAVDT~\cite{du2018unmanned} dataset consists of 80,000 accurately annotated frames with up to 14 kinds of attributes, such as weather conditions, flying attitude, camera view, vehicle category, and occlusion levels. The dataset focuses on UAV-based object detection and tracking in urban environments. Moreover, the UAVDT benchmark includes high-density scenes with small objects and significant camera motion, all challenging for the current state-of-the-art methods.
    
    \textbf{\textit{DroneVehicle}.}
    DroneVehicle~\cite{sun2022drone} proposes a large-scale drone-based dataset, which provides 28,439 RGB-Infrared image pairs to address object detection, especially under low-illumination conditions. Furthermore, it covers a variety of scenarios, such as urban roads, residential areas, and parking lots. This dataset is a significant step forward in developing autonomous driving technologies due to its unique drone perspective across a broad range of conditions.

\subsubsection{Others}

    \textbf{\textit{Pascal3D+}.}
    Pasacal3D+~\cite{xiang2014beyond} is an extension of the \textit{PASCAL VOC 2022}~\cite{everingham2010pascal},  overcoming the limitations of previous datasets by providing a richer and more varied set of annotations for images. Pasacal3D+ augments 12 rigid object categories, such as cars, buses, and bicycles, with 3D pose annotations and adds more images from ImageNet~\cite{deng2009imagenet}, resulting in high variability. However, Pascal3D+ only focuses on rigid objects and may not fully address the dynamic driving environments where pedestrians and other non-rigid objects are present.
    
    \textbf{\textit{Mapillary Vistas}.}
    Mapillary Vistas dataset, proposed by \cite{neuhold2017mapillary} in 2017, particularly aims at semantic segmentation of street scenes. The 25,000 images in the dataset are labeled with 66 object categories and include instance-specific annotations for 37 classes. It contains images from diverse weather, time of day, and geometric locations, which helps mitigate the bias towards specific regions or conditions.

\subsection{Prediction, Planning, and Control Datasets}
\label{prediction_planning_control_datasets}
Prediction, planning, and control datasets serve as the foundation for facilitating the development of driving systems. These datasets are critical for forecasting traffic dynamics, pedestrian movements, and other essential factors that influence driving decisions. Hence, we demonstrate in detail several high-impact datasets related to these tasks according to the data size, modalities, and citation number. We summarize these datasets into task-specific and multi-task groups.

\begin{table*}[!ht]
    \centering
    \caption{Prediction, planning, and control datasets. We demonstrate several crucial datasets related to prediction, planning, and control. BP: behavior prediction, DAP: driver's attention prediction, IP: intention prediction, MP: motion prediction, TP: trajectory prediction, MPlan: motion planning, DM: decision-making, OT: object tracking, QA: question-answering, DBP: driver behavior recognition}
    \resizebox{\textwidth}{!}{
    \begin{tabular}{lccccccc}
        \toprule[1pt]
        \multirow{1}{*}{\textbf{Dataset}} & \multirow{1}{*}{\textbf{Year}} & \multirow{1}{*}{\textbf{Sensing domain}} & \multirow{1}{*}{\textbf{Size}} & \multicolumn{1}{c}{\textbf{Tasks}} & \textbf{Weather conditions} & \textbf{Time of day} & \textbf{Scenario conditions} \\
        \midrule[0.5pt]
        \multicolumn{8}{c}{Task-Specific} \\
        \midrule[0.5pt]
        
        Brain4Cars~\cite{jain2016brain4cars} & 2015 & others & 2M frames & Maneuver Anticipation & ~ & ~ & ~ \\
        
        \multirow{1}{*}{JAAD~\cite{rasouli2017they}} & \multirow{1}{*}{2017} & \multirow{1}{*}{onboard} & 75K frames & \multirow{1}{*}{pedestrian IT} & \multirow{1}{*}{sunny, rainy, cloudy, snowy} & day, afternoon, night & \multirow{1}{*}{urban} \\ 

        \multirow{1}{*}{Dr(eye)ve~\cite{palazzi2018predicting}} & \multirow{1}{*}{2018} & \multirow{1}{*}{onboard} & \multirow{1}{*}{500K frames} & \multirow{1}{*}{DAP} & \multirow{1}{*}{sunny, rainy, cloudy} & \multirow{1}{*}{day, night} & \multirow{1}{*}{urban, countryside, highway} \\
        
        \multirow{1}{*}{highD~\cite{krajewski2018highd}} & \multirow{1}{*}{2018} & \multirow{1}{*}{drone} & 45K km distance & \multirow{1}{*}{TP} & \multirow{1}{*}{sunny} & \multirow{1}{*}{8 am to 5 pm} & \multirow{1}{*}{highway} \\ 
        
        \multirow{1}{*}{PIE~\cite{rasouli2019pie}} & \multirow{1}{*}{2019} & \multirow{1}{*}{onboard} & 293K frames & \multirow{1}{*}{pedestrian IT} & \multirow{1}{*}{sunny, overcast} & \multirow{1}{*}{day} & \multirow{1}{*}{urban} \\

        \multirow{1}{*}{USyd~\cite{zyner2019naturalistic}} & \multirow{1}{*}{2019} & \multirow{1}{*}{onboard} & 24K trajectories & \multirow{1}{*}{driver IT} & \multirow{1}{*}{~} & \multirow{1}{*}{~} & \multirow{1}{*}{5 intersections} \\ 

        \multirow{1}{*}{Argoverse~\cite{chang2019argoverse}} & \multirow{1}{*}{2019} & \multirow{1}{*}{onboard} & \multirow{1}{*}{300K trajectories} & \multirow{1}{*}{TP} & \multirow{1}{*}{variety} & \multirow{1}{*}{variety} & \multirow{1}{*}{urban} \\

        Drive$\&$Act~\cite{martin2019drive} & 2019 & others & 9.6M images & DBR & ~ & ~ & ~ \\
        
        DbNet~\cite{chen2019dbnet} & 2019 & onboard & 100 km distance & DBR & various road types \\
        
        D$^{2}$CAV~\cite{toghi2020maneuver} & 2020 & onboard & ~ & behavioral strategy & ~ & ~ & ~ \\
        
        \multirow{1}{*}{inD~\cite{bock2020ind}} & \multirow{1}{*}{2020} & drone & 11.5K trajectories & \multirow{1}{*}{road user prediction} & \multirow{1}{*}{sunny} & \multirow{1}{*}{day} & \multirow{1}{*}{4 urban intersections} \\ 

        \multirow{1}{*}{PePscenes~\cite{rasouli2020pepscenes}} & \multirow{1}{*}{2020} & onboard & 719 frames & \multirow{1}{*}{pedestrian BP} & \multirow{1}{*}{~} & \multirow{1}{*}{~} & \multirow{1}{*}{~} \\ 

        \multirow{1}{*}{openDD~\cite{breuer2020opendd}} & \multirow{1}{*}{2020} & \multirow{1}{*}{drone} & \multirow{1}{*}{84,774 trajectories} & \multirow{1}{*}{pedestrian BP} & \multirow{1}{*}{~} & \multirow{1}{*}{~} & \multirow{1}{*}{7 roundabouts} \\

        \multirow{1}{*}{nuPlan~\cite{caesar2021nuplan}} & \multirow{1}{*}{2021} & \multirow{1}{*}{drone} & \multirow{1}{*}{1.5K hours data} & \multirow{1}{*}{MPlan} & \multirow{1}{*}{~} & \multirow{1}{*}{~} & \multirow{1}{*}{4 cities} \\

        DriPE~\cite{guesdon2021dripe} & 2021 & others & 10K pictures & DBR & ~ & daytime & ~ \\

        Speak2label~\cite{ghosh2021speak2label} & 2021 & others & 586 videos & DAP & sunny, cloudy & daytime & ~ \\
        CoCAtt~\cite{shen2022cocatt} & 2022 & onboard & 11.9 hours & DAP & ~ & ~ & countryside \\
        
        \multirow{1}{*}{exiD~\cite{moers2022exid}} & \multirow{1}{*}{2022} & \multirow{1}{*}{drone} & \multirow{1}{*}{16 hours data} & \multirow{1}{*}{TP} & \multirow{1}{*}{sunny} & \multirow{1}{*}{daytime} & \multirow{1}{*}{7 locations on highway} \\

        \multirow{1}{*}{MONA~\cite{gressenbuch2022mona}} & \multirow{1}{*}{2022} & \multirow{1}{*}{drone} & \multirow{1}{*}{702K trajectories} & \multirow{1}{*}{TP} & \multirow{1}{*}{sunny, overcast, rain} & \multirow{1}{*}{8 am to 5 pm} & \multirow{1}{*}{urban} \\
        \multirow{1}{*}{Occ3D-nuScenes~\cite{tian2024occ3d}} & 2024 & onboard & 40K frames & occupancy prediction \\
        
        \multirow{1}{*}{Occ3D-Waymo~\cite{tian2024occ3d}} & 2024 & onboard & 200K frames & occupancy prediction \\

        \midrule[0.5pt]
        \multicolumn{8}{c}{Multi-Task} \\
        \midrule[0.5pt]
        
        HDD~\cite{ramanishka2018toward} & 2018 & onboard & 104 hours data & driver behavior, causal reasoning & ~ & ~ & suburban, urban, highway \\ 
        
        \multirow{1}{*}{INTERACTION~\cite{zhan2019interaction}} & \multirow{1}{*}{2019} & \multirow{1}{*}{drone, V2X} & \multirow{1}{*}{110K trajectories} & \multirow{1}{*}{MPlan and MP, DM} & \multirow{3}{*}{~} & \multirow{2}{*}{~} & \multirow{1}{*}{(un)signalized intersection} \\

        \multirow{1}{*}{BLVD~\cite{xue2019blvd}} & \multirow{1}{*}{2019} & \multirow{1}{*}{onboard} & \multirow{1}{*}{120K frames} & \multirow{1}{*}{4D OT,
        5D event recognition} & \multirow{1}{*}{~} & \multirow{1}{*}{day, night} & \multirow{1}{*}{urban, highway} \\
        
        \multirow{1}{*}{rounD~\cite{krajewski2020round}} & \multirow{1}{*}{2019} & \multirow{1}{*}{drone} & \multirow{1}{*}{13,746 road users} & \multirow{1}{*}{scenario classification, BP} & \multirow{1}{*}{sunny} & \multirow{1}{*}{daytime} & \multirow{1}{*}{(sub-)urban} \\
        
        PREVENTION~\cite{izquierdo2019prevention} & 2019 & onboard & 356 mins video & IP, TP & ~ & ~ & highway, urban areas \\
        
        DrivingStereo~\cite{yang2019drivingstereo} & 2019 & onboard & 180K images & trajectory planning & sunny, rainy, cloudy & daytime & suburban, urban, highway \\
        & & & & & foggy, dusky & & elevated road, country road  \\
        
        \multirow{1}{*}{Lyft Level 5~\cite{houston2021one}} & \multirow{1}{*}{2021} & \multirow{1}{*}{drone} & \multirow{1}{*}{1.1K hours data} & \multirow{1}{*}{MPlan, MP} & \multirow{1}{*}{~} & \multirow{1}{*}{~} & \multirow{1}{*}{suburban} \\

        \multirow{1}{*}{LOKI~\cite{girase2021loki}} & \multirow{1}{*}{2021} & \multirow{1}{*}{onboard} & \multirow{1}{*}{644 scenarios} & \multirow{1}{*}{TP, BP} & \multirow{1}{*}{variety} & \multirow{1}{*}{variety} & \multirow{1}{*}{(sub-)urban} \\

        \multirow{1}{*}{SceNDD~\cite{prabu2022scendd}} & \multirow{1}{*}{2022} & \multirow{1}{*}{onboard} & \multirow{1}{*}{68 driving scenes} & \multirow{1}{*}{MPlan, MP} & \multirow{1}{*}{~} & \multirow{1}{*}{~} & \multirow{1}{*}{urban} \\

        \multirow{1}{*}{DeepAccident~\cite{wang2023deepaccident}} & \multirow{1}{*}{2023} & \multirow{1}{*}{V2X} & \multirow{1}{*}{57K frames} & \multirow{1}{*}{MP, accident prediction} & \multirow{1}{*}{sunny, rainy, cloudy, wet} & \multirow{1}{*}{noon, sunset, night} & \multirow{1}{*}{synthetic} \\

       \multirow{1}{*}{V2X-Seq (forecasting)~\cite{yu2023v2x}} & \multirow{1}{*}{2023} & \multirow{1}{*}{V2X} & \multirow{1}{*}{50K scenarios} & \multirow{1}{*}{online/offline VIC TP} & \multirow{1}{*}{~} & \multirow{1}{*}{~} & \multirow{1}{*}{28 urban intersections} \\

        \bottomrule[1pt]
    \end{tabular}
    }
    \label{tab:ppc_dataset}
\end{table*}
\subsubsection{Task-Specific Datasets}
    \textbf{\textit{highD}.}
    The drone-based highD~\cite{krajewski2018highd} dataset provides a large-scale collection of naturalistic vehicle trajectories on German highways, containing post-processed trajectories of 110K cars and trucks. It is designed to address the shortcomings of traditional measurement techniques in scenario-based safety validation, which often miss capturing authentic road user behaviors or lack comprehensive, high-quality data. But highD is recorded under ideal weather conditions, limiting its application under adversarial weather conditions.
    
    \textbf{\textit{PIE}.}
    The Pedestrian Intention Estimation (PIE) dataset proposed by \cite{rasouli2019pie} represents a significant advancement in understanding pedestrian behaviors in urban environments. It encompasses over 6 hours of driving footage recorded in downtown Toronto under various lighting conditions. PIE offers rich annotations for perception and visual reasoning, including bounding boxes with occlusion flags, crossing intention confidence, and text labels for pedestrian actions. 
    
    \textbf{\textit{Argoverse}.}
    Argoverse~\cite{chang2019argoverse} is a crucial dataset in 3D object tracking and motion forecasting. Argoverse provides 360$^{\circ}$ images from 7 cameras, forward-facing stereo imagery, and LiDAR point clouds. The recorded data covers over 300K extracted vehicle trajectories from 290km of mapped lanes. With the assistance of rich sensor data and semantic maps, Argoverse is pivotal in advancing research and development in prediction systems. Real-world implementation has verified the effectiveness of Argoverse in urban environments, though its performance in different geographical areas may be constrained.
    
    \textbf{\textit{nuPlan}.}
    nuPlan~\cite{caesar2021nuplan} is the world's first closed-loop machine learning-based planning benchmark in autonomous driving. This multimodal dataset comprises around 1,500 hours of human driving data from four cities across America and Asia. It features different traffic patterns, such as merges, lane changes, interactions with cyclists and pedestrians, and driving in construction zones. These characters of the nuPlan dataset take into account the dynamic and interactive nature of actual driving, allowing for a more realistic evaluation. Its real-world impact is seen in developing more adaptive and context-aware planning systems.
    
    \textbf{\textit{exiD}.}
    The exiD~\cite{moers2022exid} trajectory dataset, presented in 2022, is a pivotal contribution to the highly interactive highway scenarios. It takes advantage of drones to record traffic without occlusion, minimizing the influence on traffic and ensuring high data quality and efficiency. This drone-based dataset surpasses previous datasets regarding the diversity of interactions captured, especially those involving lane changes at highway entries and exits. Extending the data from weather conditions and nighttime could be an improvement direction for this dataset.
    
    \textbf{\textit{MONA}.}
    The Munich Motion Dataset of Natural Driving (MONA)~\cite{gressenbuch2022mona} is an extensive dataset, with 702K trajectories from 130 hours of videos, covering urban roads with multiple lanes, an inner-city highway stretch, and their transitions. This dataset boasts an average overall position accuracy of 0.51 meters, which exhibits the quality of the data collected using highly accurate localization and LiDAR sensors. However, only recording the data in a particular city may limit its generalizability to other geographic locations.

\subsubsection{Multi-Task Datasets}

    \textbf{\textit{INTERACTION}.}
    The INTERACTION~\cite{zhan2019interaction} dataset is a versatile platform that offers diverse, complex, and critical driving scenes, along with a comprehensive semantic map, making it suitable for a multitude of tasks, such as motion prediction, imitation learning, and validation of decision and planning. Its inclusion of different countries and continents further improves the robustness of analyzing the driving behavior across different cultures. A potential shortcoming of the dataset is that the impact of environmental conditions has not been explicitly addressed.
    
    \textbf{\textit{rounD}.}
    The rounD dataset presented by \cite{krajewski2020round} is pivotal for scenario classification, road user behavior prediction, and driver modeling. It provides a large number of road user trajectories at roundabouts. The dataset utilizes a drone equipped with a 4K resolution camera to collect over six hours of video, recording more than 13K road users. The broad recorded traffic situations and the high-quality recordings make rounD an essential dataset in autonomous driving, facilitating the study of naturalistic driving behaviors in public traffic. Yet, similar to all datasets only collected under clear weather conditions, developing models capable of performing under challenging situations could be restricted.
    
    \textbf{\textit{Lyft Level 5}.}
    Lyft Level 5~\cite{houston2021one} represents one of the most extensive autonomous driving datasets for motion prediction to date, with over 1,000 hours of data. It encompasses 17,000 25-second long scenes, a high-definition semantic map with over 15,000 human annotations, 8,500 lane segments, and a high-resolution aerial image of the area. It supports multiple tasks like motion forecasting, motion planning, and simulation. The numerous multimodal data with detailed annotations make it a vital benchmark for prediction and planning. The dataset shows limitations in accurately representing scenes with uncommon traffic conditions or rare pedestrian behaviors.
    
    \textbf{\textit{LOKI}.}
    LOKI~\cite{girase2021loki}, standing for Long Term and Key Intentions, is an essential dataset in multi-agent trajectory prediction and intention prediction. LOKI tackles a crucial gap in intelligent and safety-critical systems by proposing large-scale, diverse data for heterogeneous traffic agents, including pedestrians and vehicles. This dataset makes a multidimensional view of traffic scenarios available by utilizing camera images with corresponding LiDAR point clouds, making it a highly flexible resource for the community.

    \textbf{\textit{DeepAccident}.}
    The synthetic dataset, DeepAccident~\cite{wang2023deepaccident} is the first work that provides direct and explainable safety evaluation metrics for autonomous vehicles. The extensive dataset with 57K annotated frames and 285K annotated samples supports end-to-end motion and accident prediction, which is vital in avoiding collisions and ensuring safety. Moreover, this multimodal dataset is versatile for various V2X-based perception tasks, such as 3D object detection, tracking, and BEV semantic segmentation. The various environmental conditions also enhance the generalizability of this dataset. Due to the domain adaptation issue, the performance of models trained on DeepAccident in real driving situations needs further study.

\subsection{End-to-end Datasets}
\label{e2e_datasets}
    End-to-end has become a growing trend as an alternative to modular-based architecture in autonomous driving~\cite{tampuu2020survey}. Several versatile datasets (like nuScenes~\cite{caesar2020nuscenes} and Waymo~\cite{sun2020scalability}) or simulators like \textit{CARLA}~\cite{dosovitskiy2017carla} provide the opportunity to develop end-to-end autonomous driving. Meanwhile, some works present datasets that are especially for end-to-end driving.
    
    \textbf{\textit{DDD17}.} The DDD17~\cite{binas2017ddd17} dataset is notable for using event-based cameras. The dataset provides a concurrent stream of standard active pixel sensor (APS) images and dynamic vision sensors (DVS) temporal contrast events, offering a unique blend of visual data. Additionally, DDD17 captures diverse driving scenarios, including highway and city driving and various weather conditions, thus providing exhaustive and realistic data for training and testing end-to-end autonomous driving algorithms.

\section{Annotations process}
The success and reliability of AD algorithms rely not only on the numerous data but also on high-quality annotations. In this section, we first explain the methodology for annotating data~\ref{annotation_creation}. Additionally, we analyze the most important aspects for ensuring annotation quality~\ref{annotation_quality}.
\label{annotation_process}

\subsection{Annotation Generation}
\label{annotation_creation}

\begin{figure}[htb]
    \centering
    \includegraphics[width=0.4\textwidth, trim={10, 90, 10, 90}]{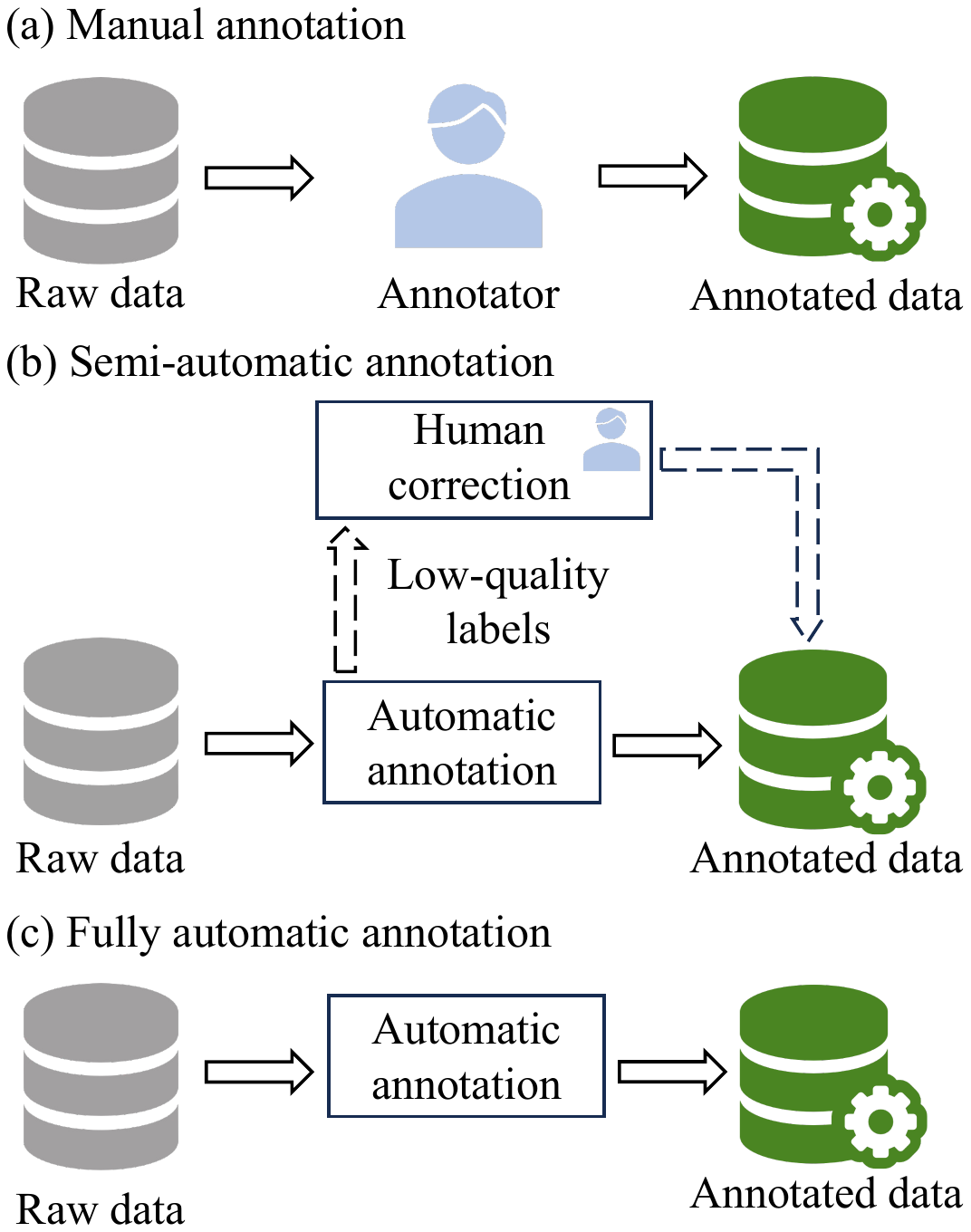}
    \caption{Annotation pipelines. We demonstrate (a) Manual annotation: The professional annotators label the raw data using annotation tools. (b) Semi-automatic annotation: After generating annotations using an automatic annotation algorithm, the low-quality labels are refined by annotators. (c) Fully automatic annotation: The framework annotates data without human correction.}
    \label{fig:annotation_pipeline}
\end{figure}

Different AD tasks require specific types of annotation. For example, object detection requires bounding box labels of instance, segmentation is based on pixel- or point-level annotations, and continually labeled trajectory is critical for trajectory prediction. On the other hand, as shown in Fig.~\ref{fig:annotation_pipeline}, the annotation pipeline can be categorized into three types: manual annotation, semi-automatic annotation, and fully automatic annotation. In this section, we detail the labeling approaches for different annotation types.

\noindent \textbf{Annotate 2D/3D Bounding Boxes.}
The quality of the bounding box annotations directly impacts the effectiveness and robustness of the perception system (like object detection) of autonomous vehicles in real-world scenarios. The annotation process generally involves labeling images with rectangular boxes or point clouds with cuboids to precisely encompass the objects of interest.

\textit{Labelme}~\cite{russell2008labelme} is a prior tool focusing on labeling images for object detection. However, generating bounding boxes by professional annotators faces the same issue as manual segmentation annotation. Wang et al.~\cite{wang2018semi} presented a semi-automatic video labeling tool based on the open-source video annotation system \textit{VATIC}\cite{vatic}. Manikandan et al.~\cite{manikandan2019deep} propose another automatic video annotation tool for AD scenes. Process bounding box annotations in the nighttime is more challenging than daytime annotation. Schorkhuber et al.~\cite{schorkhuber2021bounding} introduced a semi-automatic approach leveraging the trajectory to solve this problem.

In contrast to 2D annotations, 3D bounding boxes contain richer spatial information, such as accurate location, the object's width, length, height, and orientation in space. Hence, labeling high-quality 3D annotations requires a more sophisticated framework. Meng et al.~\cite{meng2021towards} applied a two-stage weakly supervised learning framework using human-in-the-loop to label LiDAR point clouds. \textit{ViT-WSS3D}~\cite{zhang2023simple} generated pseudo-bounding boxes by modeling global interactions between LiDAR points and corresponding weak labels. \textit{Apolloscape}~\cite{huang2018apolloscape} employed a labeling pipeline similar to~\cite{xie2016semantic}, which consists of a 3D labeling and a 2D labeling branch, to handle static background/objects and moving objects, respectively. \textit{3D~BAT}~\cite{zimmer20193d} developed an annotation toolbox to assist in obtaining 2D and 3D labels in semi-automatic labeling.

\noindent \textbf{Annotate Segmentation Data.}
The target of annotating segmentation data is to assign a label to each pixel in an image or each point in a LiDAR frame to indicate which object or region it belongs to. After labeling, all pixels belonging to an object are annotated with the same class. For the manual annotation process, the annotator first draws boundaries around an object and then fills in the area or paints over the pixels directly. However, generating pixel/point-level annotations in this way is costly and inefficient.

Many studies have proposed fully or semi-automatic annotation methods to improve annotation efficiency. Barnes et al.~\cite{barnes2017find} presented a fully automatic annotation approach based on weakly supervised learning to segment proposed drivable paths in images. The semi-automatic annotation method~\cite{saleh2016built} utilizes the objectness priors to generate segmentation masks. After that, \cite{petrovai2017semi} offered a semi-automatic method considering 20 classes. \textit{Polygon-RNN++}\cite{acuna2018efficient} presents an interactive segmentation annotation tool following the idea of~\cite{castrejon2017annotating}. Instead of using image information to generate pixel-level labels, \cite{xie2016semantic} explores transferring 3D information into 2D image domains to generate semantic segmentation annotations. For labeling 3D data, \cite{chen2019image} proposes an image-assisted annotation pipeline. Luo et al.~\cite{luo2018semantic} leverage active learning to select a few points and to form a minimal training set to avoid labeling the whole point cloud scenes. Liu et al.~\cite{liu2022less} introduce an efficient labeling framework with semi/weakly supervised learning to label outdoor point clouds.

\noindent \textbf{Annotate Trajectories.}
A trajectory is essentially a series of points that map the path of an object over time, reflecting the spatial and temporal information. Labeling trajectory data for AD is a process that entails annotating the path or movement patterns of various entities within a driving environment, such as vehicles, pedestrians, and cyclists. Usually, the annotating process relies on object detection and tracking results.

As one of the prior works in trajectory annotation, \cite{wang2011action} online generates actions for maneuvers and is annotated into the trajectory.
The annotation approach~\cite{moosavi2017annotation} consists of a crowd-sourcing step followed by a precise process of expert aggregation. Jarl et al.~\cite{jarl2022active} develop an active learning framework to annotate driving trajectory. Precisely anticipating movement patterns of pedestrians is critical for driving safety. Styles et al.~\cite{styles2019forecasting} introduce a scalable machine annotation scheme for pedestrian trajectory annotations without human effort.

\noindent \textbf{Annotate on Synthetic Data.}
Due to the expensive and time-consuming manual annotations on real-world data, synthetic data generated by computer graphics and simulators provide an alternative to address this issue. Since the data generation process is controllable and the attributes of each object in the scene (like position, size, and movement) are known, synthetic data can be automatically and accurately annotated.

The generated synthetic scenarios are designed to mimic real-world conditions, including multiple objects, various landscapes, weather conditions, and lighting variations. Selecting a suitable simulation tool is crucial to achieve this goal. \textit{Torcs}~\cite{wymann2000torcs} and \textit{DeepDriving}~\cite{chen2015deepdriving} are two prior works for simulating autonomous vehicles while lacking multi-modality information and other types of objects like pedestrians. Recently, the open source simulators, such as \textit{CARLA}~\cite{dosovitskiy2017carla}, \textit{SUMO}~\cite{krajzewicz2012recent}, and \textit{AirSim}~\cite{shah2018airsim}, have been widely used in data generation. They provide transparent platforms where researchers and developers can freely access and modify the source code, tailoring the simulator to specific requirements. In contrast, because of non-open sources, commercial simulation tools like \textit{NVIDIA's Drive Constellation}~\cite{nvidia} can make it difficult for users to create special driving environments. Rather than professional simulators, the game engines, including \textit{Grand Theft Auto 5 (GTA5)} and \textit{Unity}~\cite{juliani2018unity}, are also popular for creating synthetic autonomous driving data.

Specifically, some researchers utilize the GTA5 game engine to build datasets~\cite{richter2016playing} and \cite{richter2017playing}. Krahenbuhl et al.~\cite{krahenbuhl2018free} present a real-time system based on multiple games to generate annotations for various AD tasks. Instead of applying game videos, \textit{SHIFT}~\cite{sun2022shift}, \textit{CAOS}~\cite{hendrycks2019scaling}, \textit{FedBEVT}~\cite{song2023fedbevt}, and \textit{V2XSet}~\cite{xu2022v2x} are created based on the CARLA simulator. Compared to \cite{xu2022v2x}, \textit{V2X-Sim}~\cite{li2022v2x} studies employing multiple simulators~\cite{dosovitskiy2017carla, krajzewicz2012recent} to generate dataset for V2X perception tasks. \textit{CODD}~\cite{arnold2021fast} further exploits using \cite{dosovitskiy2017carla} to generate 3D LiDAR point clouds for cooperative driving. Other works~\cite{gaidon2016virtual, cabon2020virtual, ros2016synthia, saleh2018effective} leverage the Unity development platform to generate synthetic datasets.

\subsection{Quality of Annotations}
\label{annotation_quality}
Supervised learning-based autonomous driving (AD) algorithms rely heavily on extensive, well-labeled datasets. High-quality datasets ensure that these systems can accurately perceive and interpret complex driving environments, enhancing safety and reliability on the road. This in turn fosters user trust, a crucial factor for the widespread adoption of autonomous vehicles. Conversely, poor dataset quality can result in system errors and safety risks, undermining user confidence, hindering acceptance, and failing to meet the criteria for trustworthy AI as discussed by~\cite{li2022features}. Therefore, ensuring the quality of annotations is fundamental for improving accuracy while driving in complex real-world environments. It is even more important to fine-tune the data quality and use active learning methods for dataset curation to get robust performance results on a test set than fine-tuning the model architecture~\cite{monarch2021human}.
\begin{figure}[htb]
    \centering
    \includegraphics[width=0.48\textwidth, trim={50, 100, 50, 90}]{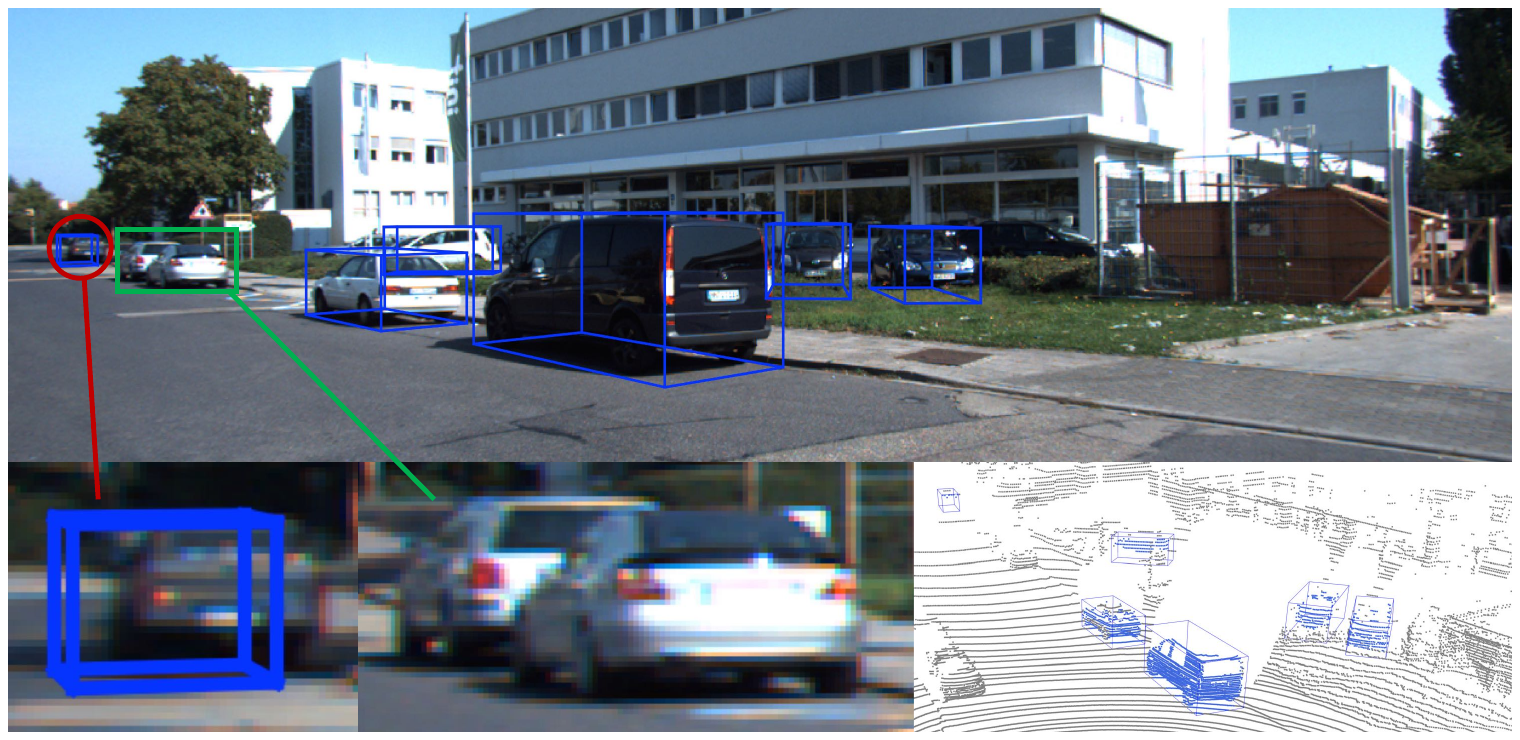}
    \caption{Mislabeling example of KITTI~\cite{geiger2012we} dataset. We show the ground truth in \textcolor{blue}{blue}. The bounding box of a car (in \textcolor{red}{red} circle) is not precise. Two cars (in \textcolor{green}{green} cube) are not annotated, although sensors obviously capture them.}
    \label{fig:low_quality_annotation}
\end{figure}

According to the study~\cite{heyn2023automotive}, the annotation quality is affected by several factors, such as consistency, correctness, precision, and validation. Consistency is the foremost criterion in evaluating annotation quality. It involves maintaining uniformity across the entire dataset and is crucial for avoiding confusion in models trained on this data. For example, if a particular type of vehicle is labeled as a car, it should be consistently annotated the same in all other instances. Annotation precision is another vital indicator, which refers to whether the labels match the actual state of the objects or scenarios. In contrast, correctness highlights that annotated data are appropriate and relevant for the purpose of the dataset and annotation guidelines. After annotation, it is essential to validate the annotated data to ensure its accuracy and completion. The process can be done through manual review by experts or algorithms. Validation helps effectively prevent issues in datasets before they infect the performance of autonomous vehicles, decreasing potential safety risks. Inel et al.~\cite{inel2019validation} presented a data-agnostic validation method for expert annotated datasets.

A failure case of annotation from KITTI~\cite{geiger2012we} is shown in Fig.~\ref{fig:low_quality_annotation}. We illustrate the ground truth bounding boxes (blue) in the corresponding image and LiDAR point cloud. On the left side of the image, the annotation of a car (circled in red) is inaccurate because it does not contain the whole object car. Additionally, two cars (highlighted by the green rectangle) are not annotated, even though the camera and LiDAR capture them clearly. 
Furthermore, datasets like the \textit{IPS300+} \cite{wang2021ips300+} contain many labeled objects within a frame (319.84 labels per frame), but on the other hand, the annotations are of bad quality. 
Many large datasets like the \textit{Pandaset}~\cite{xiao2021pandaset}, \textit{Oxford}~\cite{barnes2020oxford}, \textit{CADC}~\cite{pitropov2021canadian}, nuScenes~\cite{caesar2020nuscenes}, and \textit{Lyft Level 5}~\cite{caesar2020nuscenes} were labeled by specialized labeling companies like \textit{Scale AI} and provide high-quality annotations. Labeling the nuScenes dataset took about 7,937 hours and cost 100k USD.
Another way to label datasets is to use custom labeling tools like \textit{3D BAT}~\cite{zimmer20193d,zimmer2024tumtrafv2x}, which was used to create the \textit{TUMTraf} dataset. The Waymo and KITTI datasets were both labeled with custom labeling tools. \textit{V2V4Real}~\cite{xu2023v2v4real} has used the \textit{SUSTechPoints}~\cite{li2020sustech} labeling tool to generate the dataset.

\section{Data analysis}
\label{data_analysis}

In this section, we systematically analyze datasets from different perspectives in detail, such as the distribution of data around the world (\ref{worldwide_dist}), chronological trend~\ref{chronological_trend}, and the data distribution~\ref{data_distribution}.  

\subsection{Worldwide Distribution}
\label{worldwide_dist}
\begin{figure}[h!]
    \centering
    \includegraphics[width=0.45\textwidth]{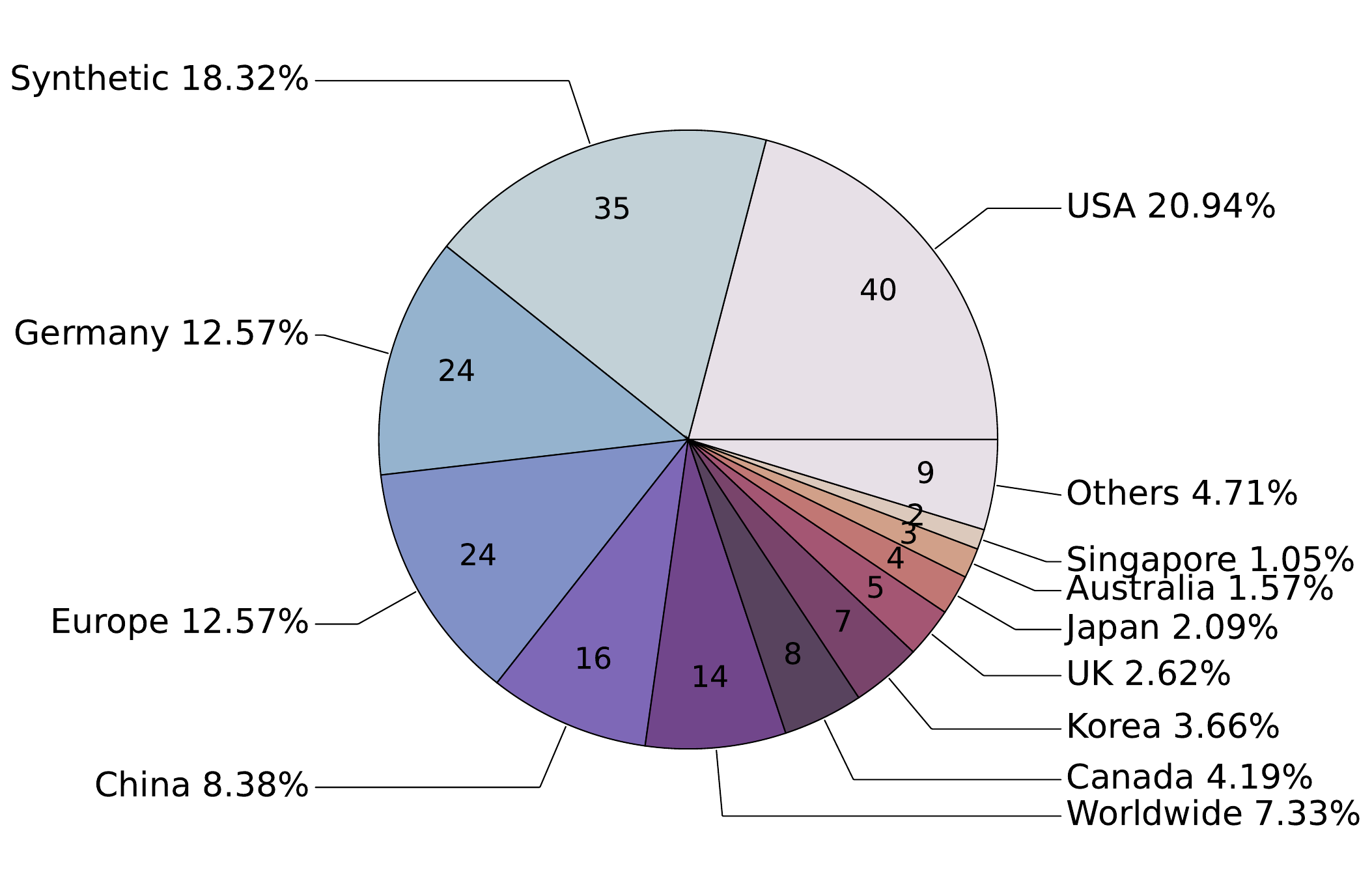}
    \caption{The distribution of datasets around the world. This figure illustrates the distribution of data collection locations of the datasets.}
    \label{fig:world_distribution}
\end{figure}

We demonstrate an overview of the global distribution of 191 autonomous driving datasets in Fig.~\ref{fig:world_distribution}. The chart indicates that the USA is at the forefront with 40 datasets (21\% share), underscoring its leadership in the autonomous driving domain. Germany accounts for 24 datasets, and China closely follows with 16 datasets. On the other hand, developed countries, including Canada, Korea, the UK, Japan, and Singapore, cover the rest smaller segments. Although 11 datasets are collected worldwide and 24 are from the European region (except Germany), all these countries or regions are considered high-income areas. The dominance of the USA, western Europe, and East Asia reflects an extremely unbalanced development of autonomous driving around the world.

Specifically, one of the most classic datasets, KITTI~\cite{geiger2012we}, is collected in the metropolitan area of Karlsruhe, Germany. In comparison, both Waymo~\cite{sun2020scalability} and Argoverse~2~\cite{wilson2023argoverse} datasets are collected from a wide variety of places, including six different cities in the USA individually. Apolloscapes~\cite{huang2018apolloscape} and DAIR-V2X~\cite{yu2022dair} are recorded in China. Instead of collecting data solely within a country, nuScenes~\cite{caesar2020nuscenes} is established based on the data from America (Boston) and Singapore, both known for complex and highly challenging traffic situations. Other well-known autonomous driving datasets~\cite{yu2020bdd100k, ye2022rope3d, sun2022drone, krajewski2018highd, rasouli2019pie, gressenbuch2022mona} are collected in those previously mentioned countries. It is noted that thanks to the geographic diversity, \cite{caesar2020nuscenes, sun2020scalability} have been widely used in transfer learning to verify the generalizability of autonomous driving algorithms. 

Moreover, autonomous driving faces unique challenges across different geographical regions. However, relying solely on data from a single source can introduce bias that could result in autonomous vehicles' failure to perform under varied or unseen regions and cases. For instance, the diversity and quantity of electric scooters in China far exceed those in Germany, meaning an algorithm trained exclusively on German data may struggle to accurately recognize objects in China. Hence, Recording data from different continents and countries can assist in addressing unique challenges caused by geographical locations. This diverse regional distribution enhances the robustness of the collected data and highlights the international efforts and collaborations in the research community and industry. 

Furthermore, 35 synthetic datasets generated by simulators like CARLA~\cite{dosovitskiy2017carla} take up 18.32\% percent. Due to the limitation of recording from real-world driving environments, these synthetic datasets overcome such drawbacks and are critical for exploiting more robust and reliable driving systems. However, the domain adaptation from synthetic data to real-world data is still a challenging research topic, which limits a broader application of synthetic data and relevant simulators.

\subsection{Chronological Trends in Perception Datasets}
\label{chronological_trend}
In Fig.~\ref{fig:chronological_overview}, we introduce a chronological overview of perception datasets with the top 50 impact scores from 2009 to 2024 (until the writing of this paper). The datasets are color-coded according to their sourcing domain, and synthetic datasets are marked with an external red outline, clearly illustrating the progress toward the diverse data collection strategy. A noticeable trend shows the increase in the number and variety of datasets over the years, indicating the requirement for high-quality datasets with the growing advancements in the field of autonomous driving. 

In general, most of the datasets provide a perception perspective from the sensors equipped on the ego vehicle (onboard) because of the importance of the capability of the autonomous vehicle to efficiently and precisely precept the surroundings. On the other hand, due to the high-cost real-world data, some researchers propose high-influence synthetic datasets like VirtualKITTI~\cite{gaidon2016virtual} (2016) to alleviate the reliance on real data. Facilitated by the effectiveness of simulators, there are many novel synthetic datasets~\cite{hendrycks2019scaling} \cite{sun2022shift} published in recent years. In the timeline, V2X datasets like DAIR-V2X~\cite{yu2022dair} and TUMTraf (\cite{cress2022a9,zimmer2023tumtrafintersection,cress2024tumtrafevent,zimmer2024tumtrafv2x}) also exhibit a trend toward cooperative driving systems. Furthermore, because of the non-occlusion perspective provided by UAV, drone-based datasets, such as UAVDT~\cite{du2018unmanned} published in 2018, take a crucial position in advancing perception systems.

\subsection{Data Distribution}
\label{data_distribution}
\begin{figure}[htb]
    \centering
    \includegraphics[width=0.48\textwidth]{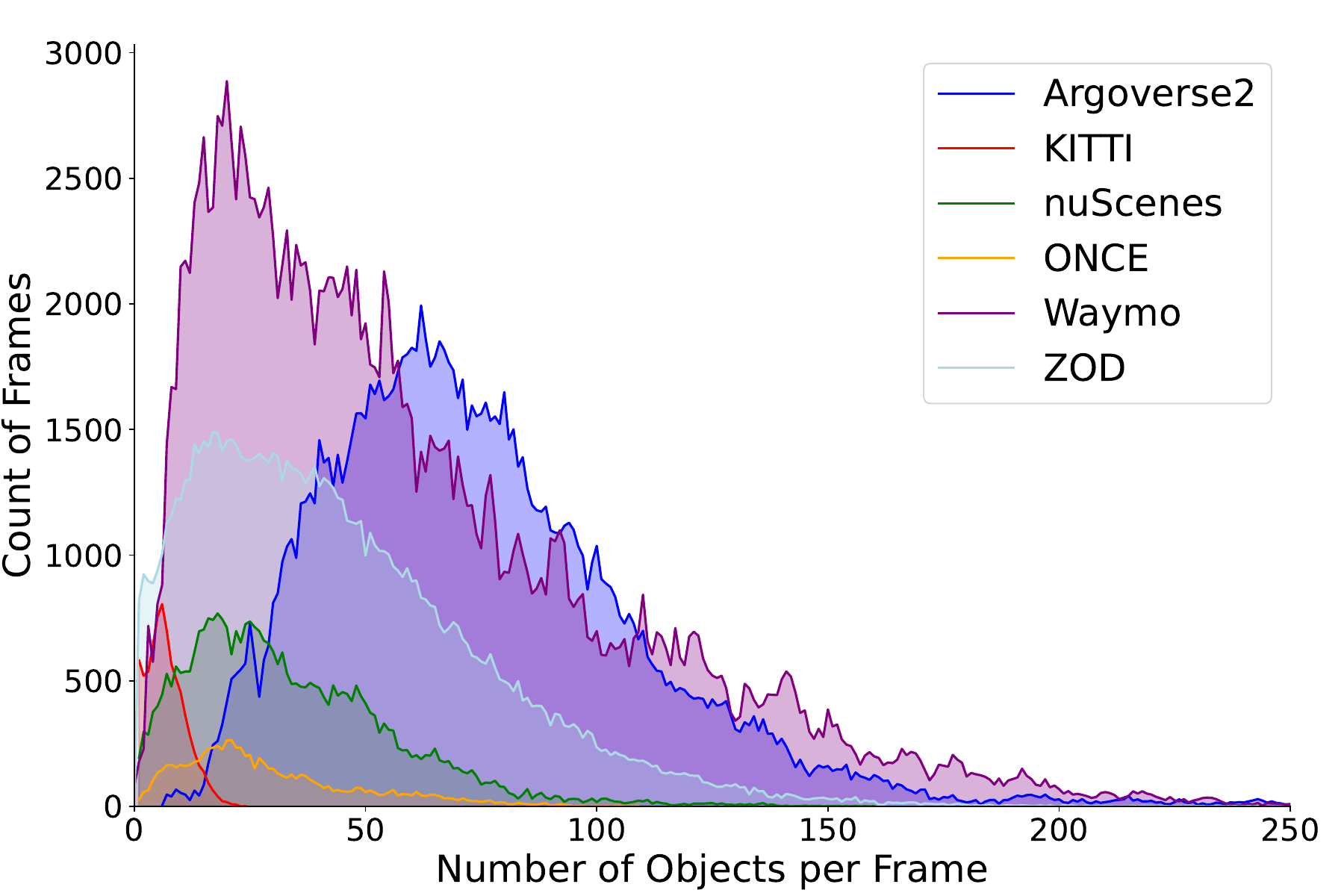}
    \caption{Comparison of the distribution of the number of objects per frame across several datasets: Argoverse~2~\cite{wilson2023argoverse}, KITTI~\cite{geiger2012we}, nuScenes~\cite{caesar2020nuscenes}, ONCE~\cite{mao2021one}, Waymo~\cite{sun2020scalability}, and ZOD~\cite{alibeigi2023zenseact}. The horizontal axis quantifies the number of objects detected in a single frame, while the vertical axis represents the count of frames containing that number of objects.}
    \label{fig:sample_dist}
\end{figure}
\begin{figure} [htb]
    \centering
    \includegraphics[width=0.48\textwidth]{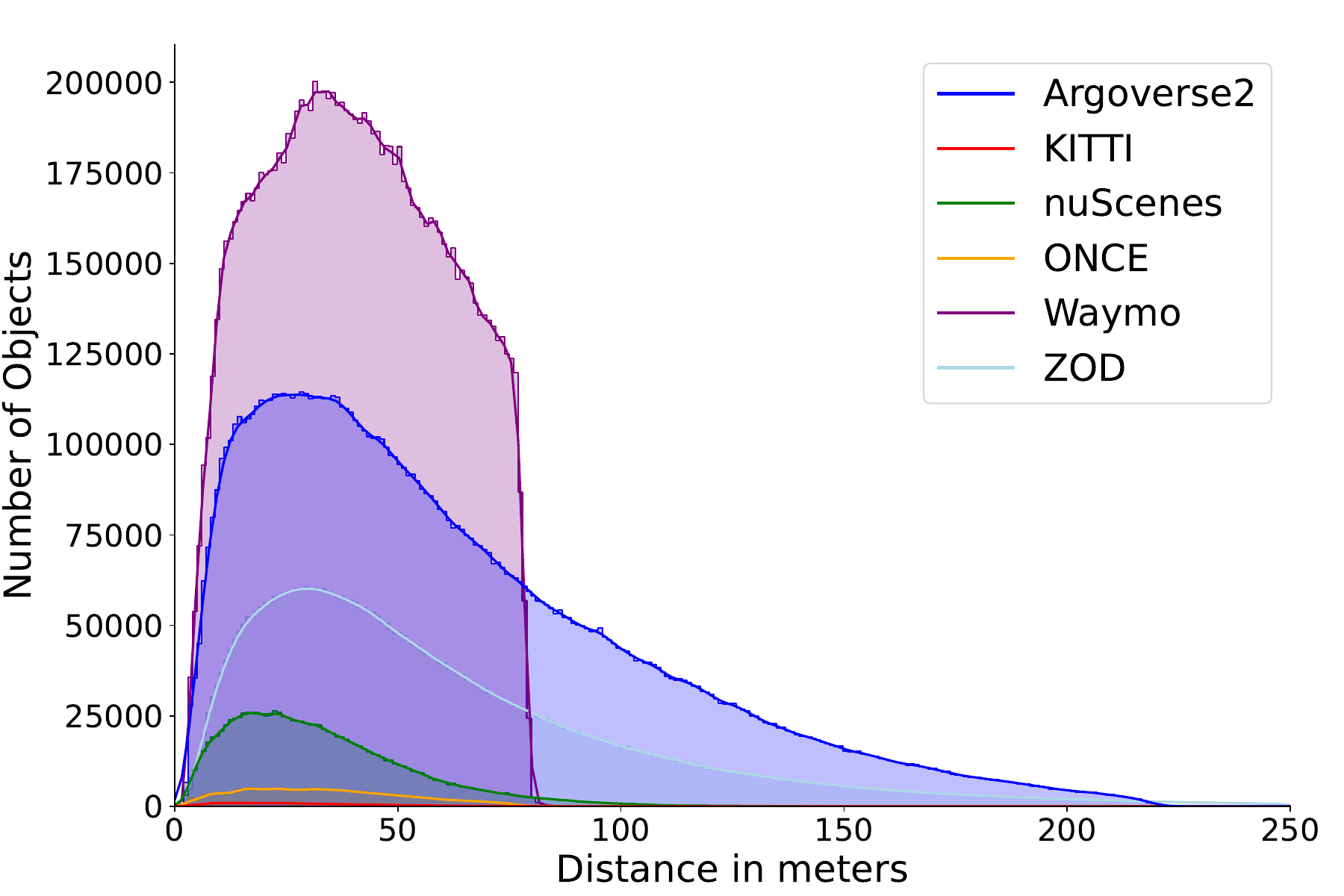}
    \caption{Comparison of the distribution of the number of objects detected at various distances across several datasets: Argoverse~2~\cite{wilson2023argoverse}, KITTI~\cite{geiger2012we}, nuScenes~\cite{caesar2020nuscenes}, ONCE~\cite{mao2021one}, Waymo~\cite{sun2020scalability}, and ZOD~\cite{alibeigi2023zenseact}. The horizontal axis measures the distance from the ego vehicle in meters, and the vertical axis quantifies the number of objects detected at that distance.}
    \label{fig:distance_dist}
\end{figure}

We introduce an insight into the number of objects per frame for these datasets in Fig.~\ref{fig:sample_dist}. Notably, Waymo~\cite{sun2020scalability} exhibits an extreme number of frames with less than 50 objects while maintaining a broad presence across the chart, illustrating a wide range of scenarios from low to high object density per frame. Contrastingly, KITTI~\cite{geiger2012we} shows a more constrained distribution and limited data size. Argoverse~2~\cite{wilson2023argoverse} features a substantial number of frames with a higher object count---its peak appears around 70, which indicates its complex environmental representations in general. For ONCE~\cite{mao2021one}, its density of objects evenly distributes in the supported perception range. Datasets like nuScenes~\cite{caesar2020nuscenes} and ZOD~\cite{alibeigi2023zenseact} demonstrate similar curves with a quick rise and slow decline, implying a moderate level of environmental complexity with a decent variability of object counts per frame. 

Beyond the number of objects in a scene, the object distribution based on the distance to the ego vehicle is another essential point for revealing a dataset's variety and significant differences, illustrated in Fig.~\ref{fig:distance_dist}. The Waymo dataset demonstrates numerous labeled objects in near-field to mid-field scenarios. In contrast, Argoverse~2 and ZOD show a wider detection range, with some frames even including bounding boxes out of 200 meters. The curve of the nuScenes means it is particularly rich in objects in a shorter range, which is typical for urban driving scenarios. Nevertheless, as the distance increases, the nuScenes dataset quickly tappers off for the number of objects with annotations. The ONCE dataset covers a more even distribution of objects across distances, while the KITTI dataset focuses more on close-range objects.

\subsection{Influence of Adversarial Environmental Conditions.}
\begin{table}[!htb]
    \centering
    \caption{3D object detection performance under adversarial environmental conditions on the nuScenes~\cite{caesar2020nuscenes} dataset. We report mAP (\%) as the evaluation metric. 'L' and 'C' refer to LiDAR and cameras, respectively. The term 'Normal'  indicates that the model was tested across the entire validation set, encompassing all environmental conditions.}
    \resizebox{\linewidth}{!}{
    \begin{tabular}{cccc}
    \toprule[1pt]
         \textbf{Method} & \textbf{Modality} & \textbf{Adversarial Conditions} & \textbf{mAP $\uparrow$} \\
         \midrule
         \multirow{3}{*}{VoxelNext~\cite{chen2023voxelnext}} & \multirow{3}{*}{L} & normal & \textbf{60.5} \\
         & & nighttime & 32.4 \\
         & & rainy & 58.7 \\
         \midrule
         \multirow{3}{*}{UVTR~\cite{li2022unifying}} & \multirow{3}{*}{L} & normal & \textbf{60.8} \\
         & & nighttime & 33.9 \\
         & & rainy & 56.5 \\
         \midrule
         \multirow{3}{*}{Transfusion~\cite{bai2022transfusion}} & \multirow{3}{*}{L+C} & normal & \textbf{68.9} \\
         & & nighttime & 37.5 \\
         & & rainy & 65.8 \\
    \bottomrule[1pt]
         
    \end{tabular}}

    \label{tab:det_res_adversarial_conditions}
\end{table}
We further study the influence of adversarial environmental conditions (low lightning and rain) on the performance of 3D object detectors in the autonomous driving system. The experiment results are reported in Tab.~\ref{tab:det_res_adversarial_conditions}. We utilize the nuScenes~\cite{caesar2020nuscenes} dataset and choose three state-of-the-art methods, \textit{VoxelNext}~\cite{chen2023voxelnext}, \textit{UTVR}~\cite{li2022unifying}, and \textit{Transfusion}~\cite{bai2022transfusion}. For a fair comparison, we directly utilize the pre-trained checkpoints provided in the repository of each approach. The rainy and nighttime data are also manually collected from the nuScenes validation set. All three methods demonstrate similar trends under different environmental conditions. Specifically, the detection accuracy declines significantly under low illumination conditions compared with evaluations of the whole validation set. Besides, all models exhibit slight decreases under the non-heavy rainy situations recorded by nuScenes. The reliability of detectors in the real world could be much worse once the rain is heavy. Therefore, taking image restoration into account is a promising way for camera or sensor fusion-based approaches to overcome these challenges~\cite{cui2023focal, cui2023image}. In conclusion, increasing the amount of data recorded on various types and intensities of weather conditions is critical for training a robust and reliable autonomous driving system.

\section{Discussion and Future Works}
\label{discussion_future_works}
\begin{table*}[!t]
    \centering
    \caption{VLM autonomous driving dataset. OT: object tracking, MOT: multi-object tracking, QA: question-answering, DM: decision-making, IP: intention prediction, VR: visual reasoning, SR: spatial reasoning}
    \resizebox{\textwidth}{!}{
    \begin{tabular}{lcccccc}
        \toprule[1pt]
        \textbf{Dataset} & \textbf{Year} & \textbf{Size}& \textbf{Temporal} & \textbf{Sensing domain} & \textbf{Tasks} & \textbf{Real/Synthetic} \\
        \midrule[0.5pt]
        BDD-X~\cite{kim2018textual} & 2018 & 8.4M & $\checkmark$ & onboard & reasoning, planning & real \\
        Cityscapes-Ref~\cite{vasudevan2018object} & 2018 & 5,000 stereo videos & $\checkmark$ & onboard & object referring & real \\
        TOUCHDOWN~\cite{chen2019touchdown} & 2019 & 9,326 examples & $\checkmark$ & onboard & reasoning, navigation & real \\
        Talk2Car~\cite{deruyttere2019talk2car} & 2019 & 11,959 commands, 850 videos & $\checkmark$ & onboard & object referring & real \\
        BDD-OIA~\cite{xu2020explainable} & 2020 & 11,303 scenarios & $\times$ & onboard & explainable decision-making & real \\
        CityFlow-NL~\cite{feng2021cityflow} & 2021 & 5,289 samples & $\checkmark$ & onboard & OT & real\\
        CARLA-NAV~\cite{jain2023ground} & 2022 & 83K & $\checkmark$ & onboard & navigation & real \\
        NuPrompt~\cite{wu2023nuprompt} & 2023 & 34K frames, 35K prompts & $\checkmark$ & onboard & MOT & real \\
        NuScenes-QA~\cite{qian2023nuscenes} & 2023 & 1K scenarios, 460K QA pairs & $\checkmark$ & onboard & visual QA & real \\
        Refer-KITTI~\cite{wu2023referring} & 2023 & 6,650 frames & $\checkmark$ & onboard & referring MOT & real \\
        Driving LLMs~\cite{chen2023driving} & 2023 & 10K driving situations, 160K QA pairs & $\times$ & drone & visual QA & synthetic \\
        DRAMA~\cite{malla2023drama} & 2023 & 17,785 scenarios & $\checkmark$ & onboard & reasoning, visual QA & real \\
        Rank2Tell~\cite{sachdeva2023rank2tell} & 2023 & 116 scenarios & $\checkmark$ & onboard & importance level ranking & real \\
        LamPilot~\cite{ma2023lampilot} & 2023 & 4,900 samples & $\checkmark$ & others & planning & synthetic \\
        LangAuto CARLAR~\cite{shao2023lmdrive} & 2023 & 64K data clips & $\checkmark$ & onboard & closed-loop driving & synthetic \\
        NuScenes-MQA~\cite{inoue2023nuscenes} & 2023 & 34K scenarios, 1.4M QA pairs & $\times$ & onboard & visual QA & real \\
        DriveMLM~\cite{wang2023drivemlm} & 2023 & 280 hours & $\checkmark$ & onboard & planning, control & synthetic \\
        DriveLM-nuScenes~\cite{sima2023drivelm} & 2023 & 4,871 frames & $\checkmark$ & onboard & end-to-end driving & real \\
        DriveCARLA~\cite{sima2023drivelm} & 2023 & 183,373 frames & $\checkmark$ & onboard & end-to-end driving & real \\
        LiDAR-text~\cite{yang2023lidar} & 2023 & 420K 3D captioning data, 280K 3D grounding data & $\times$ & onboard & 3D scene understanding & real \\
        Talk2BEV~\cite{dewangan2023talk2bev} & 2023 & 1K BEV scenarios, 20K QA pairs & $\checkmark$ & onboard & DM, IP, VR, SR & real \\
        NuScenes-MQA~\cite{inoue2024nuscenes} & 2024 & 1.4M QA pairs, 34,149 scenarios & $\checkmark$ & onboard & visual QA & real \\
        VLAAD~\cite{park2024vlaad} & 2024 & 10,379 scenarios &  $\checkmark$ & onboard & visual QA, reasoning & real \\
        \bottomrule[1pt]
    \end{tabular}}
    \label{vlm_dataset}
\end{table*}

With rapid technological development, powerful computation resources, and excellent artificial intelligent algorithms, many novel trends in next-generation autonomous driving datasets have occurred while proposing new challenges and requirements.

\noindent \textbf{End-to-End Driving Datasets.}
Compared to the modular-designed autonomous driving pipeline, the end-to-end architecture simplifies the overall design process and reduces integration complexities. The success of \textit{UniAD}~\cite{hu2023planning} verifies the potential ability of end-to-end models. However, the number of datasets for end-to-end AD is limited~\cite{binas2017ddd17}. Therefore, introducing datasets focusing on end-to-end driving is crucial for advancing autonomous vehicles. On the other hand, implementing an automatic labeling pipeline in a data engine can significantly facilitate the development of end-to-end driving frameworks and data~\cite{chen2023end}. 


\noindent \textbf{Potential Applications of AD Datasets}
Future autonomous driving datasets should provide extensive real-world environmental and traffic data, supporting a wide range of applications beyond the ego-vehicle and basic vehicle-infrastructure cooperation. For instance, the interaction data between autonomous vehicles and intelligent infrastructure components can guide the advancement of Internet-of-Thing (IoT)-enabled devices like smart traffic signals. Moreover, the detailed insights into traffic patterns, congestion, and vehicle behavior across different times and conditions can facilitate urban planning, optimize the traffic flow, and enhance overall traffic management strategies.

\noindent \textbf{Introduce Language into AD Datasets.}
Vision language models (VLMs) have recently achieved impressive advancement in many fields. Its inherent advantage in providing language information to vision tasks makes autonomous driving systems more explainable and reliable. The survey~\cite{cui2023survey} highlights the prominent role of Multimodal Large Language Models in various AD tasks, such as perception~\cite{vasudevan2018object, wu2023nuprompt}, motion planning~\cite{ma2023lampilot}, and motion control~\cite{dewangan2023talk2bev}. Autonomous driving datasets including language labels are shown in Tab.~\ref{vlm_dataset}. Overall, incorporating language into AD datasets is a trend of the future development of AD datasets.

\noindent \textbf{Data Generation via VLMs.}
The powerful capability of VLMs can be used to generate data~\cite{zhou2023vision, wang2024does, li2022novel, tian2023vistagpt}. For example, \textit{DriveGAN}~\cite{kim2021drivegan} generated high-quality AD data by disentangling different components without supervision. Additionally, due to the capability of world models for comprehending driving environments, some works~\cite{hu2023gaia, wang2023drivedreamer, jia2023adriver} explored world models to generate high-quality driving videos. \textit{DriveDreamer}~\cite{wang2023drivedreamer} as a pioneering work derived from real-world scenarios, addressing the limitation of gaming environments or simulated settings. The most recent advancements in text-video generation, exemplified by \textit{Sora}~\cite{videoworldsimulators2024}, have opened a new avenue for generating autonomous driving data. With a brief description, Sora can generate high-quality synthetic data that is very close to real scenes. This capability significantly enhances dataset augmentation, especially by extending the data available for rare events like traffic accidents. The increased data availability provided by VLMs is poised to enhance the training and evaluation of autonomous driving systems, potentially improving their safety and reliability.

\noindent \textbf{Domain Adaptation.}
Domain adaptation is a critical challenge in developing autonomous vehicles~\cite{schwonberg2023survey}, referring to the ability of a model trained on one dataset (the source domain) to perform stable on another dataset (the target domain). This challenge manifests in multiple aspects, such as diversity in driving conditions~\cite{erkent2020semantic}, sensor settings~\cite{wei2022lidar}, or synthetic-to-real transform~\cite{hu2022sim}. Therefore, a trend of developing the next-generation datasets is incorporating data from heterogeneous sources. First of all, datasets should not only cover a wide range of environmental conditions (various weather conditions and day-night illumination) but also include diverse geographic locations. Second, combining data from various sensor types is also pivotal to overcoming the domain adaptation issue. Another solution is taking into account blending high-quality synthetic data and real-world data in a balanced way to improve generalization.

\noindent \textbf{Uncertainty Issues in Autonomous Driving.}
Usually, uncertainty is modeled in a probabilistic way in the field of machine learning. The uncertainty can be referred to two types: a) aleatoric uncertainty, where the uncertainty like noise stems from the data, and b) epistemic uncertainty refers to uncertainty caused by unawareness about the best model~\cite{hullermeier2021aleatoric}. For autonomous driving, one of the major reasons resulting in uncertainty is the incompleteness of training data~\cite{shafaei2018uncertainty}. The insufficient training data can not wholly represent the driving environment, causing autonomous vehicles to problematically deal with rare cases during operation. Therefore, improving the diversity of the dataset for training reliable autonomous driving systems is crucial. Moreover, datasets include rare events and edge cases, allowing models to better understand and quantify the uncertainty associated with unexpected situations and safely handle them. 

\noindent \textbf{Standardization of Data Creation.} 
Addressing the standardization of data creation is critical aspect for developing new datasets, as it directly impacts the accuracy, efficiency, and reliability of autonomous vehicle models. In general, data standardization refers to the attributes of the data, terminology and structure of the dataset, and data storage~\cite{gal2019data}. For the data attributes, unifying data formats across different sensor types and source domains is valuable to facilitate integrated processing and analysis. Developing comprehensive guidelines for dataset labeling is crucial to guarantee consistent, high-quality annotations, enhancing model training across diverse datasets and boosting performance reliability. Moreover, establishing standardized data storage and access protocols is vital, enabling seamless dataset sharing and integration across multiple sources, and fostering collaboration within the autonomous driving research community.

\noindent \textbf{Data Privacy.}
The advancement of autonomous vehicles require a lot of data to guarantee the driving safety. However, the more the available data, the greater the concern that private data will be abused. For the earlier autonomous driving datasets like KITTI~\cite{geiger2012we}, there is no anonymization progress for the recorded data, especially for images, which leads to a potential risk for leak private information. With the introduction and refinement of relevant regulations in various countries, data anonymization has been widely adopted in recent datasets~\cite{cress2022a9, caesar2020nuscenes, sun2020scalability}. Nevertheless, the information provided by a large dataset is more than individual biographic features or vehicle plates. Professional institutes can indirectly extract external information by analyzing the existing information in the data. For instance, by analyzing the types of vehicles and the attire of pedestrians in the dataset, one can infer information about the infrastructure, basic construction, and other characteristics of the surrounding area in which the data were collected.

\noindent \textbf{Open Data Ecosystems.}
The primary objective of open data ecosystems (ODE) for autonomous driving is to foster innovation, enhance transparency, and facilitate collaboration across governments, companies, and research communities. This is achieved through the free exchange of datasets, breaking down the traditional barriers that have restricted access to corporations and research institutions. By doing so, ODE empowers a wider range of innovators to participate in autonomous driving development, thereby boosting a more diverse and inclusive innovation ecosystem. Additionally, ODE establishes dynamic feedback loops where users can report issues, propose improvements, and contribute to enhancing datasets. 
Nevertheless, unrestricted access to data raises strong concerns about data security and privacy. Addressing these concerns necessitates carefully crafting and continuously refining relevant legal frameworks to safeguard sensitive information while prompting the growth of ODE. The balance is pivotal for maintaining the integrity of open data ecosystems and ensuring their sustainable development in autonomous driving.

\section{Conclusion}
\label{conclusion}

In this paper, we exhaustively and systematically reviewed 265 existing autonomous driving datasets. We started with the sensor types and modalities, sensing domains, and tasks relevant to autonomous driving. We introduced a novel evaluation metric called impact score to validate the influence and importance of perception datasets. Our analysis covered the attributes and significance of high-impact datasets across perception, prediction, planning, control, and end-to-end driving tasks. Additionally, we investigated the annotation methodologies and the factors affecting annotation quality. We also analyzed datasets from chronological and geographical perspectives to understand current trends and conducted experiments demonstrating the crucial need for data diversity under adversarial conditions. With the study on data distribution, we offered a specific viewpoint for comprehending the variances across different datasets. Our findings underscore the essential role of diverse, high-quality datasets in shaping the future of autonomous driving. Looking forward, we highlighted key challenges and future directions for autonomous driving datasets, including the adoption of VLM, domain adaptation, uncertainty challenges, standardization, data privacy, and open data ecosystems. These areas represent promising avenues for future research and are pivotal for advancing autonomous driving technology, setting a clear roadmap for further innovation in this rapidly evolving field.

\appendix
\section*{Supplementary Tables of AD Datasets}
\begin{table*}[!h]
    \centering
    \caption{Part I of Autonomous Driving Datasets. SS: semantic segmentation, OD: object detection, OT: object tracking, MOT: multi-object tracking}
    \resizebox{\textwidth}{!}{
    \begin{tabular}{lcccccc}
        \toprule[1pt]
        \textbf{Dataset} & \textbf{Year} & \textbf{Size} & \textbf{Temporal} & \textbf{Sensing domain} & \textbf{Tasks} & \textbf{Real/Synthetic} \\
        \midrule [0.5pt]
        ETH Ped~\cite{ess2007depth} & 2007 & 2,293 frames & $\checkmark$ & onboard & pedestrian detection & real \\
        TUD-Brussels~\cite{wojek2009multi} & 2009 & 1,600 frames & $\times$ & onboard & (3D) OD & real \\
        Collective activity~\cite{choi2009they} & 2009 & 44 short videos & $\checkmark$ & others & human activity recognition & real \\
        San Francisco Landmark~\cite{chen2011city} & 2011 & 150K panoramic images & $\times$ & others & landmark identification & real \\
        Daimler Stereo Ped~\cite{li2016new} & 2011 & 21,790 frames & $\checkmark$ & onboard & pedestrian detection & real \\
        BelgiumTS~\cite{timofte2014multi} & 2011 & 13K traffic sign annotations & $\times$ & onboard & traffic sign detection & real \\
        Stanford Tack~\cite{teichman2011towards} & 2011 & 14K tracks & $\checkmark$ & onboard & classification & real \\
        TME Motorway~\cite{caraffi2012system} & 2012 & 30K frames & $\checkmark$ & onboard & OD, OT & real \\
        MSLU~\cite{blanco2014malaga} & 2013 & 36.8 km distances &  $\checkmark$ & onboard & SLAM & real \\
       SydneyUrbanObject~\cite{de2013unsupervised} & 2013 & 588 object scans & $\times$ & onboard & classification & real \\
        Ground Truth SitXel~\cite{pfeiffer2013exploiting} & 2013 & 78,500 frames & $\checkmark$ & onboard & stereo confidence & real \\
        NYC3DCars~\cite{matzen2013nyc3dcars} & 2013 & 2K images & $\times$ & onboard & OD & real \\
        AMUSE~\cite{koschorrek2013multi} & 2013 & 117,440 frames & $\checkmark$ & onboard & SLAM & real \\
        Daimler Ped~\cite{schneider2013pedestrian} & 2013 & 12,485 frames & $\checkmark$ & onboard & pedestrian path prediction & real \\
        LISA~\cite{mogelmose2014traffic} & 2014 & 6,610 frames & $\checkmark$ & onboard & traffic sign detection & real \\
        Paris-rue-Madame~\cite{serna2014paris} & 2014 & 643 objects & $\times$ & onboard & OD, SS & real \\
        TRANCOS~\cite{guerrero2015extremely} & 2015 & 1.2K images & $\times$ & V2X & onboard number estimation & real \\
        FlyingThings3D~\cite{mayer2016large} & 2015 & 26,066 frames & $\checkmark$ & others & scene flow estimation & synthetic \\
        Ua-detrac~\cite{wen2020ua} & 2015 & 140K frames & $\checkmark$ & V2X & OD, MOT & real \\
        NCLT~\cite{carlevaris2016university} & 2015 & 34.9 hours & $\checkmark$ & onboard & odometry & real \\
        CCSAD~\cite{guzman2015towards} & 2015 & 96K frames & $\checkmark$ & onboard & scene understanding & real \\
        KAIST MPD~\cite{hwang2015multispectral} & 2015 & 95K color-thermal pair frames & $\times$ & onboard & pedestrian detection & real \\
        SAP~\cite{sap} & 2016 & 19K frames & $\times$ & drone & OD, OT & real \\
        LostAndFound~\cite{pinggeralost} & 2016 & 2,104 frames & $\checkmark$ & onboard & obstacle detection & real \\
        UAH-Driveset~\cite{romera2016need} & 2016 & 500 mins & $\checkmark$ & onboard & lane detection, detection & real \\
        CURE-TSR~\cite{temel2017cure} & 2017 & 2.2M annotated images & $\times$ & onboard & traffic sign detection & real \\
        TuSimple~\cite{tusimple} & 2017 & 6,408 frames & $\times$ & onboard & lane detection, velocity estimation & real \\
        TRoM~\cite{liu2017benchmark} & 2017 & 712 frames & $\times$ & onboard & road marking detection & real \\
        NEXET~\cite{klein2017nexet} & 2017 & 91,190 frames & $\times$ & onboard & OD & real \\
        DIML~\cite{diml} & 2017 & 470 videos & $\checkmark$ & onboard & lane detection & real \\
        Bosch STL~\cite{behrendt2017deep} & 2017 & 13,334 images & $\checkmark$ & onboard & traffic light detection and classification & real \\
        Complex Urban~\cite{jeong2019complex} & 2017 & around 190km paths & \checkmark & onboard & SLAM & real \\
        DDD20~\cite{hu2020ddd20} & 2017 & 51 hours event frames & $\checkmark$ & onboard & end-to-end driving & real \\
        PedX~\cite{kim2019pedx} & 2018 & 5K images & $\checkmark$ & onboard & pedestrian detection and tracking & real \\
        LiVi-Set~\cite{chen2018lidar} & 2018 & 10K frames & $\checkmark$ & onboard & driving behavior prediction & real \\
        Syncapes~\cite{wrenninge2018synscapes} & 2018 & 25K images & $\times$ & onboard & OD, SS & synthetic \\
        NVSEC~\cite{zhu2018multivehicle} & 2018 & around 28 km distances & $\checkmark$ & others & SLAM & real \\
        Aachen Day-Night~\cite{sattler2018benchmarking} & 2018 & 4,328 images, 1.65M points & $\checkmark$ & onboard & visual localization & real \\
        CADP~\cite{shah2018cadp} & 2018 & 1,416 scenes & $\times$ & V2X & traffic accident analysis & real \\
        TAF-BW~\cite{fleck2019towards} & 2018 & 2 scenarios & $\checkmark$ & V2X & MOT, V2X communication & real \\
        comma2k19~\cite{schafer2018commute} & 2018 & 2M images & $\times$ & onboard & pose estimation, end-to-end driving & real \\
        nighttime drive~\cite{dai2018dark} & 2018 & 35K & $\times$ & onboard & SS & real \\
        VEIS~\cite{saleh2018effective} & 2018 & 61,305 frames & $\times$ & onboard & OD, SS & synthetic \\
        Paris-Lille-3D~\cite{roynard2018paris} & 2018 & 2,479 frames & $\times$ & onboard & 3D SS, classification & real \\
        NightOwis~\cite{neumann2019nightowls} & 2018 & 56K frames & $\checkmark$ & onboard & pedestrian detection, tracking & real \\
        EuroCity Persons~\cite{braun1805eurocity} & 2018 & 47,300 frames & $\times$ & onboard & OD & real \\
        RANUS~\cite{choe2018ranus} & 2018 & 4K frames & $\times$ & onboard & SS, scene understanding & real \\
        SynthCity~\cite{griffiths2019synthcity} & 2019 & 367.9M points in 30 scans & $\times$ & onboard & (3D) OD, (3D) SS & synthetic \\
        D$^{2}$-City~\cite{che2019d} & 2019 & 700K annotated frames & $\checkmark$ & onboard & OD, MOT & real \\
        Caltech Lanes~\cite{aly2008real} & 2019 & 1,224 frames & $\times$ & onboard & lane detection & real \\
        Mcity~\cite{dong2019mcity} & 2019 & 1,7500 frames & $\checkmark$ & onboard & SS & real \\
        DET~\cite{cheng2019det} & 2019 & 5,424 event-based camera images& $\times$ & onboard & lane detection & real \\
        PreSIL~\cite{hurl2019precise} & 2019 & 50K frames & $\checkmark$ & onboard & OD, 3D SS & synthetic \\
        H3D~\cite{patil2019h3d} & 2019 & 27,721 frames & $\checkmark$ & onboard & (3D) OD, MOT & real \\
        LLAMAS~\cite{behrendt2019unsupervised} & 2019 & 100K frames & $\checkmark$ & onboard & lane segmentation & real \\
        MIT DriveSeg~\cite{ding2021value} & 2019 & 5K frames & $\times$ & onboard & SS & real \\
        Astyx~\cite{meyer2019automotive} & 2019 & 500 radar frames & $\times$ & onboard & 3D OD & real \\
        UNDD~\cite{nag2019s} & 2019 & 7.2K frames & $\checkmark$ & onboard & SS & real \\
        Boxy~\cite{behrendt2019boxy} & 2019 & 200K frames & $\checkmark$ & onboard & OD & real \\
        RUGD~\cite{wigness2019rugd} & 2019 & 37K frames & $\checkmark$ & others & SS & real \\
        ApolloCar3D~\cite{song2019apollocar3d} & 2019 & 5,277 driving images & $\times$ & onboard & 3D instance understanding & real \\
        HEV~\cite{yao2019egocentric} & 2019 & 230 video clips & $\checkmark$ & onboard & object localization & real \\
        HAD~\cite{kim2019grounding} & 2019 & 5,675 video clips & $\checkmark$ & onboard & end-to-end driving & real \\
        CARLA-100~\cite{codevilla2019exploring} & 2019 & 100 hours driving & $\checkmark$ & onboard & path planning, behavior cloning & synthetic \\
        Brno Urban~\cite{ligocki2020brno} & 2019 & 375.7 km & $\checkmark$ & onboard & recognition & real \\
        VERI-Wild~\cite{lou2019veri} & 2019 & 416,314 images & $\checkmark$ & V2X & onboard re-identification & real \\
        CityFlow~\cite{tang2019cityflow} & 2019 & 200K bounding boxes & $\checkmark$ & V2X & OD, MOT, re-identification & real \\
        VLMV~\cite{cordes2021vehicle} & 2020 & 306K frames & $\checkmark$ & V2X & lane merge & real \\
        Small Obstacles~\cite{singh2020lidar} & 2020 & 3K frames & $\times$ & onboard & small obstacle segmentation & real \\
        Cirrus~\cite{wang2021cirrus} & 2020 & 6,285 frames & $\checkmark$ & onboard & (3D) OD & real \\
        KITTI InstanceMotSeg~\cite{mohamed2021monocular} & 2020 & 12,919 frames & $\checkmark$ & onboard & moving instance segmentation & real \\
        A*3D~\cite{pham20203d} & 2020 & 39,179 point cloud frames & $\times$ & onboard & (3D) OD & real \\
        Toronto-3D~\cite{tan2020toronto} & 2020 & 4 scenarios & $\times$ & onboard & 3D SS & real \\
        MIT-AVT~\cite{ding2020avt} & 2020 & 1.15M 10s video clips & $\checkmark$ & onboard & SS, anomaly detection & real \\
        CADC~\cite{pitropov2021canadian} & 2020 & 56K & $\times$ & onboard & (3D) OD, OT & real \\
        SemanticPOSS~\cite{pan2020semanticposs} & 2020 & 2,988 point cloud frames & $\checkmark$ & onboard & 3D SS & synthetic \\
        IDDA~\cite{alberti2020idda} & 2020 & 1M frames & $\times$ & onboard & segmentation & synthetic \\
        \bottomrule[1pt]
        
    \end{tabular}}
    \label{all_dataset}
\end{table*}
\begin{table*}[!h]
    \centering
    \caption{Part II of Autonomous Driving Datasets}
    \resizebox{\textwidth}{!}{
    \begin{tabular}{lcccccc}
        \toprule[1pt]
        \textbf{Dataset} & \textbf{Year} & \textbf{Size} & \textbf{Temporal} & \textbf{Sensing domain} & \textbf{Tasks} & \textbf{Real/Synthetic} \\
        \midrule [0.5pt]
        CARRADA~\cite{ouaknine2021carrada} & 2020 & 7,193 radar frames & $\checkmark$ & onboard & SS & real \\
        Titan~\cite{malla2020titan} & 2020 & 75,262 frames & $\checkmark$ & onboard & OD, action recognition & real \\
        NightCity~\cite{xie2023boosting} & 2020 & 4,297 frames & $\times$ & onboard & nighttime SS & real \\
        PePScenes~\cite{rasouli2020pepscenes} & 2020 & 40K frames & $\checkmark$ & onboard & (3D) OD, pedestrian action prediction & real \\
        DDAD~\cite{guizilini20203d} & 2020 & 21,200 frames & $\times$ & onboard & depth estimation & real \\
        MulRan~\cite{kim2020mulran} & 2020 & 41.2km paths & $\checkmark$ & onboard & place recognition & real \\
        Oxford Radar RobotCar~\cite{barnes2020oxford} & 2020 & 240K scans & $\checkmark$ & onboard & odometry & real \\
        OTOH~\cite{houston2021one} & 2020 & 170K scenes & $\checkmark$ & drone & trajectory prediction, planning & real \\
        DA4AD~\cite{zhou2020da4ad} & 2020 & 9 sequences & $\checkmark$ & onboard & visual localization & real \\
        CPIS~\cite{arnold2020cooperative} & 2020 & 10K frames & $\times$ & V2X & cooperative 3D OD & synthetic \\
        EU LTD~\cite{yan2020eu} & 2020 & around 37 hours & $\checkmark$ & onboard & odometry & real \\
        Newer College~\cite{ramezani2020newer} & 2020 & 290M points, 2300 seconds & $\checkmark$ & others & SLAM & real \\
        CCD~\cite{bao2020uncertainty} & 2020 & 4.5K videos & $\checkmark$ & onboard & accident prediction & real \\
        LIBRE~\cite{carballo2020libre} & 2020 & 40 frames & $\times$ & onboard & LiDAR performance benchmark & real \\
        Gated2Depth~\cite{gruber2019gated2depth} & 2020 & 17,686 frames & $\checkmark$ & onboard & depth estimation & real \\
        TCGR~\cite{wiederer2020traffic} & 2020 & 839,350 frames & $\checkmark$ & others & traffic control gesture recognition & real \\\
        DSEC~\cite{gehrig2021dsec} & 2021 & 53 sequences (3193 seconds in total) & $\checkmark$ & onboard & dynamic perception & real \\
        4Seasons~\cite{wenzel20214seasons} & 2021 & 350km recordings & $\checkmark$ & onboard & SLAM & real \\
        PVDN~\cite{saralajew2021dataset} & 2021 & 59,746 frames & $\checkmark$ & onboard & nighttime OD, OT & real \\
        ACDC~\cite{sakaridis2021acdc} & 2021 & 4,006 images & $\times$ & onboard & SS on adverse conditions & real \\
        DRIV100~\cite{sakashita2021driv100} & 2021 & 100K frames & $\times$ & onboard & domain adaptation SS & real \\
        NEOLIX~\cite{wang2020neolix} & 2021 & 30K frames & $\checkmark$ & onboard & 3D OD, OT & real \\
        IPS3000+~\cite{wang2021ips300+} & 2021 & 14,198 frames & $\checkmark$ & V2X & 3D OD & real \\
        AUTOMATUM~\cite{spannaus2021automatum} & 2021 & 30 hours & $\checkmark$ & drone & trajectory prediction & real \\
        DurLAR~\cite{li2021durlar} & 2021 & 100K frames & $\checkmark$ & onboard & depth estimation & real \\
        Reasonable-Crowd~\cite{helou2021reasonable} & 2021 & 92 scenarios & $\checkmark$ & onboard & driving behavior prediction & synthetic \\
        MOTSynth~\cite{fabbri2021motsynth} & 2021 & 768 driving sequences & $\checkmark$ & onboard & Pedestrian detection and tracking & synthetic \\
        MAVD~\cite{valverde2021there} & 2021 & 113,283 images & $\checkmark$ & onboard & OD and OT with sound & real \\
        Multifog KITTI~\cite{mai20213d} & 2021 & 15K frames & $\times$ & onboard & 3D OD & synthetic \\
        Comap~\cite{yuan2021comap} & 2021 & 4,391 frames & $\checkmark$ & V2X & 3D OD & synthetic \\
        R3~\cite{oh2022towards} & 2021 & 369 scenes & $\times$ & onboard & out-of-distribution detection & real \\
        WIBAM~\cite{howe2021weakly} & 2021 & 33,092 frames & $\checkmark$ & V2X & 3D OD & real \\
        CeyMo~\cite{jayasinghe2022ceymo} & 2021 & 2,887 frames & $\times$ & onboard & road marking detection & real \\
        RaidaR~\cite{jin2021raidar} & 2021 & 58,542 rainy street scenes & $\times$ & onboard & SS & real \\
        Fishyscapes~\cite{blum2021fishyscapes} & 2021 & 1,030 frames & $\times$ & onboard & SS, anomaly detection & real \\
        RadarScenes~\cite{schumann2021radarscenes} & 2021 & 40K radar frames & $\checkmark$ & onboard & OD, classification & real \\
        ROAD~\cite{singh2022road} & 2021 & 122K frames & $\checkmark$ & onboard & OD, SS & real \\
        All-in-One Drive~\cite{weng2023all} & 2021 & 100K & $\checkmark$ & onboard & (3D) OD, (3D) SS, trajectory prediction & real \\
        PandaSet~\cite{xiao2021pandaset} & 2021 & 8,240 frames & $\checkmark$ & onboard & (3D) OD, SS, OT & real \\
        SODA10M~\cite{han2021soda10m} & 2021 & 20K labeled images & $\checkmark$ & onboard & OD & real \\
        PixSet~\cite{deziel2021pixset} & 2021 & 29K point cloud frames & $\times$ & onboard & (3D) OD & real \\
        RoadObstacle21~\cite{chan2021segmentmeifyoucan} & 2021 & 327 scenes & $\times$ & onboard & anomaly segmentation & synthetic \\
        VIL-100~\cite{zhang2021vil} & 2021 & 10K frames & $\times$ & onboard & lane detection & real \\
        OpenMPD~\cite{zhang2022openmpd} & 2021 & 15K frames & $\times$ & onboard & (3D) OD, 3D OT, semantic segmentation & real \\
        WADS~\cite{kurup2021dsor} & 2021 & 1K point cloud frames & $\checkmark$ & onboard & SS & real \\
        CCTSDB 2021~\cite{zhang2022cctsdb} & 2021 & 16,356 frames & $\times$ & onboard & traffic sign detection & real \\
        SemanticUSL~\cite{jiang2021lidarnet} & 2021 & 1.2K frames & $\times$ & onboard & domain adaptation 3D SS & real \\
        CARLANE~\cite{gebele2022carlane} & 2022 & 118K frames & $\checkmark$ & onboard & lane detection & synthetic \\
        CrashD~\cite{lehner20223d} & 2022 & 15,340 scenes & $\times$ & onboard & 3D OD & synthetic \\
        CODD~\cite{arnold2021fast} & 2022 & 108 sequences & $\checkmark$ & V2X & multi-agent SLAM & synthetic \\
        CarlaScenes~\cite{kloukiniotis2022carlascenes} & 2022 & 7 sequences & $\checkmark$ & onboard & (3D) SS, SLAM, depth estimation & synthetic \\
        OPV2V~\cite{xu2022opv2v} & 2022 & 11,464 frames & $\times$ & V2X & onboard-to-onboard perception & synthetic \\
        CARLA-WildLife~\cite{maag2022two} & 2022 & 26 videos & $\checkmark$ & onboard & out-of-distribution tracking & synthetic \\
        SOS~\cite{maag2022two} & 2022 & 20 videos & $\checkmark$ & onboard & out-of-distribution tracking & real \\
        RoadSaW~\cite{cordes2022roadsaw} & 2022 & 720K frames & $\checkmark$ & onboard & Road surface and wetness estimation & real \\
        I see you~\cite{quispe2022see} & 2022 & 170 sequences, 340 trajectories & $\checkmark$ & V2X & OD & real \\
        ASAP~\cite{wang2023we} & 2022 & 1.2M images & $\checkmark$ & onboard & online 3D OD & real \\
        Amodal Cityscapes~\cite{breitenstein2022amodal} & 2022 & 5K frames & $\times$ & onboard & amodal SS & real \\
        SynWoodScape~\cite{sekkat2022synwoodscape} & 2022 & 80K frames & $\times$ & onboard & (3D) OD, segmentation & synthetic \\
        TJ4RadSet~\cite{zheng2022tj4dradset} & 2022 & 7,757 frames & $\checkmark$ & onboard & OD, OT & real \\
        CODA~\cite{li2022coda} & 2022 & 1,500 frames & $\times$ & onboard & corner case detection & real \\
        LiDAR Snowfall~\cite{hahner2022lidar} & 2022 & 7,385 point cloud frames & $\checkmark$ & onboard & 3D OD & synthetic \\
        MUAD~\cite{franchi2022muad} & 2022 & 10.4K frames & $\checkmark$ & onboard & OD. SS, depth estimation & synthetic \\ 
        AUTOCASTSIM~\cite{cui2022coopernaut} & 2022 & 52K frames & $\checkmark$ & V2X & (3D) OD, OT, SS & real \\
        CARTI~\cite{bai2022pillargrid} & 2022 & 11K frames & $\checkmark$ & V2X & cooperative perception & synthetic \\
        K-Lane~\cite{k-lane} & 2022 & 15,382 frames & $\times$ & onboard & lane detection & real \\
        Ithaca365~\cite{diaz2022ithaca365} & 2022 & 7K frames & $\times$ & onboard & 3D OD, SS, depth estimation & real \\
        GLARE~\cite{gray2023glare} & 2022 & 2,157 frames & $\times$ & onboard & traffic sign detection & real \\
        SUPS~\cite{hou2022sups} & 2023 & 5K frames & $\checkmark$ & onboard & SS, depth estimation, SLAM & synthetic \\
        Boreas~\cite{burnett2023boreas} & 2023 & 7,111 frames & $\checkmark$ & onboard & (3D) OD, localization & real \\
        Robo3D~\cite{kong2023robo3d} & 2023 & 476K frames & $\checkmark$ & onboard & (3D) OD, 3D SS & real \\
        ZOD~\cite{alibeigi2023zenseact} & 2023 & 100K frames & $\checkmark$ & onboard & (3D) OD, segmentation & real \\
        K-Radar~\cite{k-radar} & 2023 & 35K radar frames & $\times$ & onboard & 3D OD, OT & real \\
        aiMotive~\cite{matuszka2022aimotive} & 2023 & 26,583 frames & $\checkmark$ & onboard & 3D OD, MOT & real \\
        UrbanLaneGraph~\cite{buchner2023learning} & 2023 & around 5,220 km lane spans & $\checkmark$ & drone & lane graph estimation & real \\
        WEDGE~\cite{marathe2023wedge} & 2023 & 3,360 frames & $\times$ & others & OD, classification & synthetic \\
        OpenLane-V2~\cite{wang2023openlane} & 2023 & 466K images & $\checkmark$ & onboard & lane detection, scene understanding & real \\
        V2X-Seq (perception)~\cite{yu2023v2x} & 2023 & 15K frames & $\checkmark$ & V2X & cooperative perception & real \\
        SSCBENCH~\cite{li2023sscbench} & 2023 & 66,913 frames & $\times$ & onboard & semantic scene completion & real \\
        RoadSC~\cite{cordes2023camera} & 2023 & 90,759 images & $\times$ & onboard & road snow coverage estimation & real \\ 
        V2X-Real~\cite{xiang2024v2x} & 2024 & 171K images, 33K LiDAR point clouds & - & V2X & 3D OD & real \\
        RCooper~\cite{hao2024rcooper} & 2024 & 50K images, 30K LiDAR point clouds & $\checkmark$ & V2X & 3D OD, OT & real \\
        FLIR~\cite{flir} & - & 26,442 thermal frames & $\checkmark$ & onboard & thermal image OD & real \\
        \bottomrule[1pt]
    \end{tabular}}
    \label{all_dataset2}
\end{table*}

\newpage
{
    \bibliographystyle{ieeetr}
    \bibliography{reference}
}

\end{document}